\documentclass[pdflatex,sn-mathphys-ay,iicol]{sn-jnl}

\usepackage{graphicx}%
\usepackage{multirow}%
\usepackage{amsmath,amssymb,amsfonts}%
\usepackage{amsthm}%
\usepackage{mathrsfs}%
\usepackage[title]{appendix}%
\usepackage{xcolor}%
\usepackage{textcomp}%
\usepackage{manyfoot}%
\usepackage{booktabs}%
\usepackage{algorithm}%
\usepackage{algorithmicx}%
\usepackage{algpseudocode}%
\usepackage{listings}%

\usepackage{titlesec}
\usepackage{geometry}
\usepackage{natbib}
\makeatletter
\renewcommand\bibfont{\reset@font\fontfamily{\rmdefault}\fontsize{9bp}{11bp}\selectfont}
\makeatother

\theoremstyle{thmstyleone}%
\theoremstyle{thmstyletwo}%

\theoremstyle{thmstylethree}%

\begin{document}

\title[Open World Detection Survey]{Towards Open World Detection: A Survey}


\author[1]{\fnm{Andrei-Ștefan} \sur{Bulzan}}

\author*[1]{\fnm{Cosmin} \sur{Cernăzanu-Glăvan}}\email{cosmin.cernazanu@cs.upt.ro}

\affil[1]{\orgdiv{Department of Computer and Information Technology}, \orgname{Universitatea Politehnică Timișoara}, \orgaddress{\street{Timișoara}, \country{Romania}}}


\abstract{For decades, Computer Vision has aimed at enabling machines to perceive the external world. Initial limitations led to the development of highly specialized niches. As success in each task accrued and research progressed, increasingly complex perception tasks emerged. This survey charts the convergence of these tasks and, in doing so, introduces Open World Detection (OWD), an umbrella term we propose to unify class-agnostic and generally applicable detection models in the vision domain. We start from the history of foundational vision subdomains and cover key concepts, methodologies and datasets making up today's state-of-the-art landscape. This traverses topics starting from early saliency detection, foreground/background separation, out of distribution detection and leading up to open world object detection, zero-shot detection and Vision Large Language Models (VLLMs). We explore the overlap between these subdomains, their increasing convergence, and their potential to unify into a singular domain in the future, perception.}


\maketitle

\section{Introduction}\label{sec1}

In its pursuit of enabling machines to perceive and interpret the world, Computer Vision has seen considerable progress. Initially, the field tackled this complex challenge by dividing it into smaller, more manageable problems. Early research concentrated on specialized tasks like edge detection, image classification, and object recognition, resulting in the creation of task-specific algorithms. Although successful, these efforts highlighted the limitations of handling vision problems in isolation.

\begin{figure}[h]
    \centering
    \includegraphics[width=\linewidth]{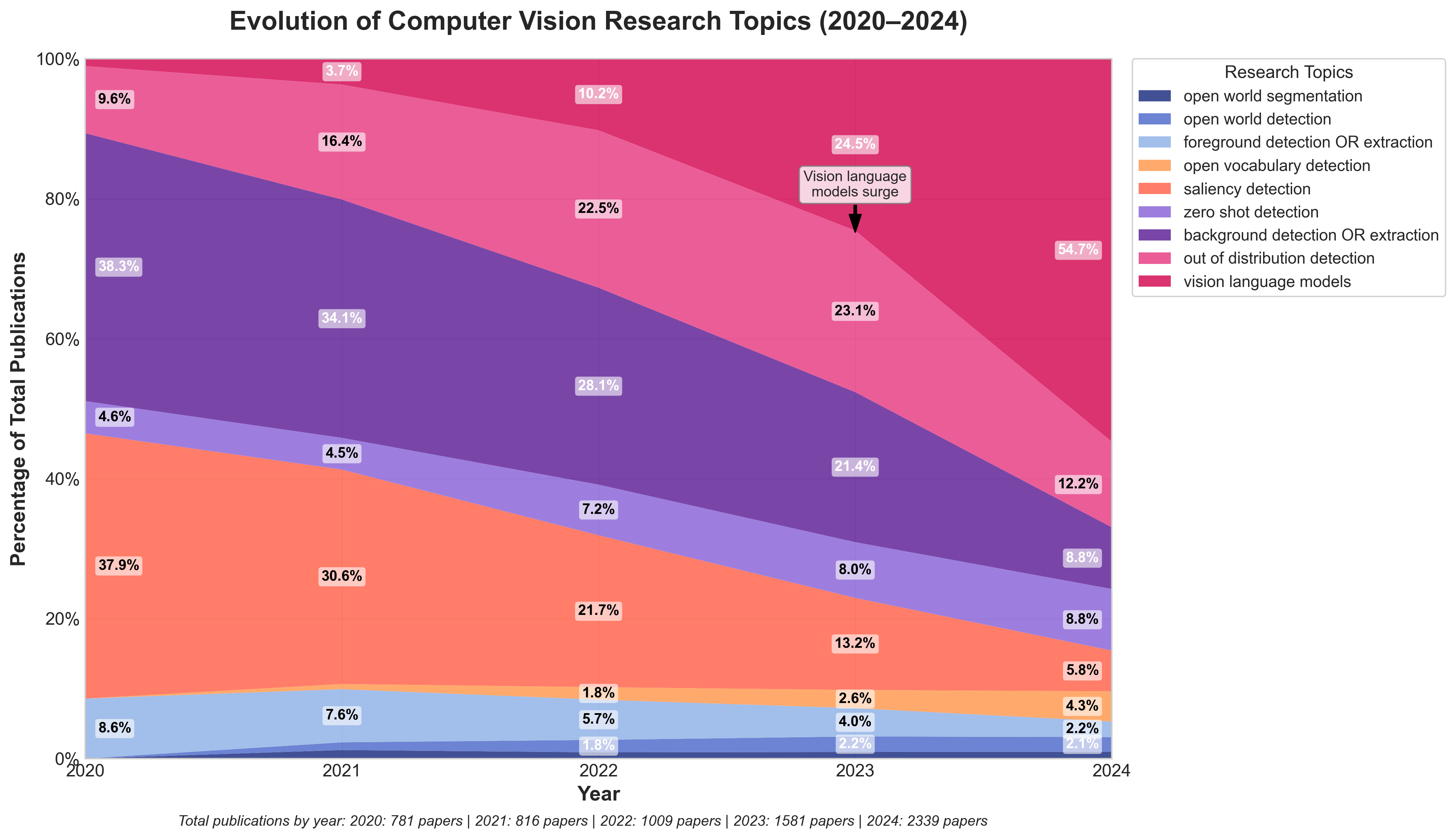}
    \caption{Recent Trends 2020-2024 of Published Articles based on words contained in title}
    \label{fig:trends3}
\end{figure}

The onset of more sophisticated machine learning techniques and the exponential growth in computational power have driven the turn towards more holistic approaches in computer vision. Recent years have seen the emergence of models capable of performing multiple vision tasks simultaneously, fueled by the integration of vast datasets and the development of large-scale multimodal foundational models - Fig.~\ref{fig:trends3} displays some of the recent trends in the field.

\begin{figure}[h]
    \centering
    \includegraphics[width=\linewidth]{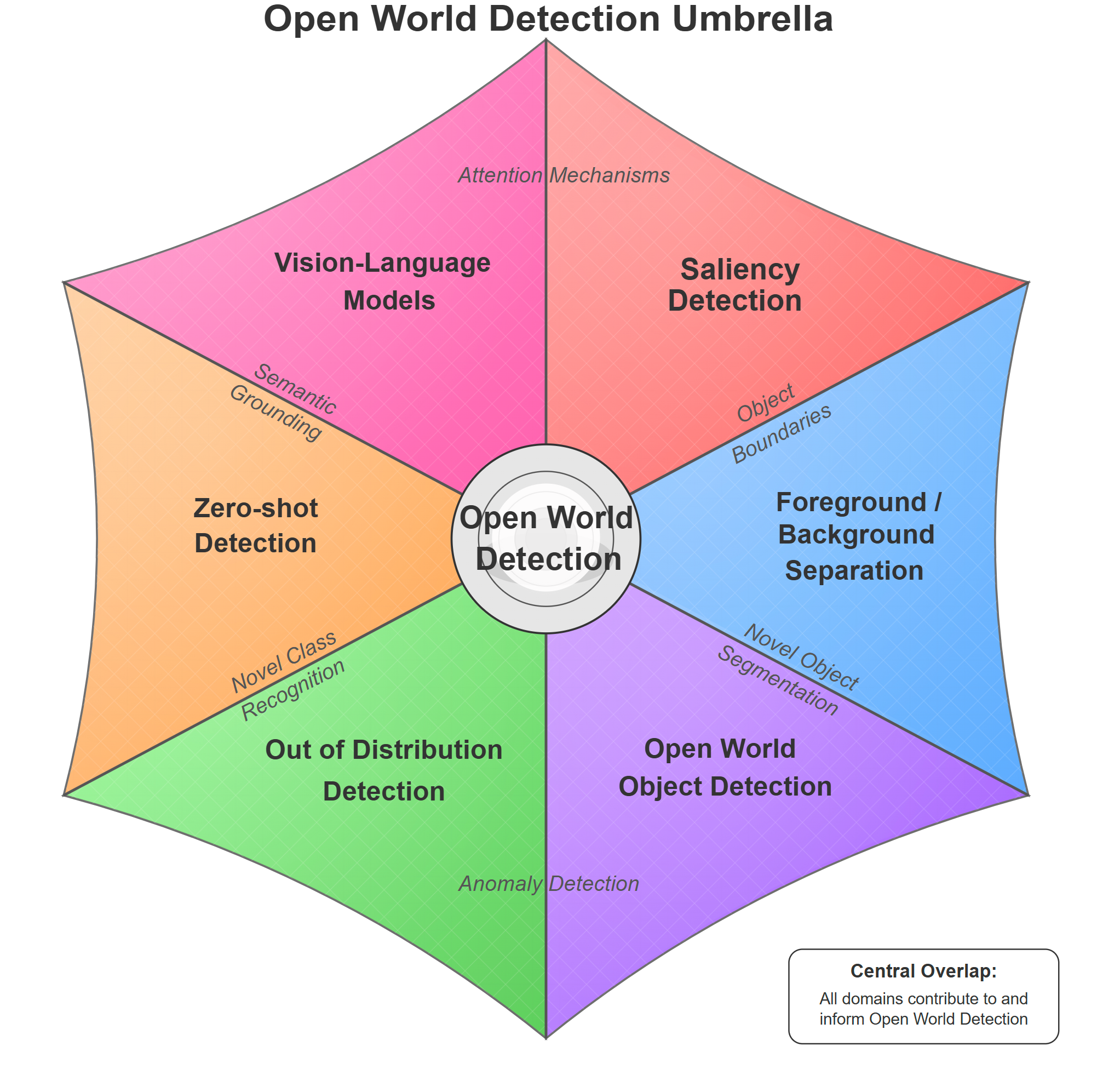}
    \caption{OWD Umbrella}
    \label{fig:owd_umbrella}
\end{figure}

Open World Object Detection (OWOD), term coined by \cite{joseph2021towards}, extends classical object detection to enable perception of all objects visible in a given scene. This is irrespective of their class being explicitly presented to the model at training time, essentially creating models capable of class-agnostic, open world detection. We strip away the 'Object' restriction and refer to Open World Detection (OWD) as an umbrella term for solutions generalizing detection across a wide range of visual tasks. This conceptualization of OWD with its various contributing domains, is visually depicted in Fig.~\ref{fig:owd_umbrella}. Unlike traditional models that operate within closed and well-defined categories, OWD systems strive to recognize and adapt to the endless variability of real-world environments. This approach is underpinned by advancements in key niche vision areas we aim to present here.

In setting out to provide a comprehensive overview of the current state of what falls under OWD, we explore the foundational concepts, methodologies, and applications that define the field. We present the integration and overlap of subdomains such as open world object detection, open vocabulary detection, saliency detection, foreground/background separation, and zero-shot detection. Additionally, we examine the role of Vision-Language Models (VLMs) and Large Multimodal Models (LMMs) in enhancing the capabilities of OWD systems and discuss relevant datasets that support the development and evaluation of these models.

By synthesizing insights from these various subdomains, we highlight the increasing convergence of the field of computer vision. This convergence suggests a future where the boundaries between different vision tasks blur, leading to unified models that can seamlessly address a wide spectrum of visual challenges. Our goal is to provide researchers and practitioners with a thorough review of where we stand in terms of OWD and its potential to fundamentally transform how machines perceive and navigate the complexities of the world.

\noindent To systematically explore this convergence and chart the trajectory of OWD, the rest of this work is structured as follows:
Chapter 2 commences by analyzing the foundational subdomains that have paved the way for OWD. This chapter journeys from an examination of saliency detection and foreground/background separation, progresses through out-of-distribution and zero-shot detection, and culminates in Open World Object Detection and the pivotal role of Vision-Language Models.
Subsequently, Chapter 3 transitions to a comprehensive review of datasets and benchmarks crucial for training and evaluating OWD systems, providing a landscape of the data driving advancements in the field.
Chapter 4 then maps the current state-of-the-art, delineating the major eras of visual detection and highlighting the paradigm shifts that have led to contemporary methodologies, offering a structured understanding of the field’s evolution and current state.
Looking forward, Chapter 5 is dedicated to exploring promising avenues for advancement and the unfolding potential of this domain, casting light on the next horizons of OWD research.
Finally, Chapter 6 synthesizes the key insights gleaned throughout the survey and reiterates the transformative potential of OWD, providing a closing perspective.

\section{Subdomains}\label{sec2}

The term "Open World Object Detection" was initially proposed to address a specific subset of detection tasks. However, we extend this concept to encompass a broader range of tasks that collectively contribute to the detection of all objects of interest in a scene. These tasks, although independent, can be synergistically utilized to achieve the overarching goal of OWD. As an example, saliency detection identifies objects that attract attention, while foreground/background separation distinguishes objects from the background. Each of these tasks plays a role in the holistic detection of objects in an open-world context.
Note: When using the 'object' term here, we refer to the smallest visually perceivable entity present in a given scene that holds a minimum of interest to an observer.

In this chapter we present what we deem to be the vision subdomains that have contributed the most in the past decades to an open ended perception of the world. Some of these domains are building blocks of what makes today's state of the art, with little chance of alone conquering the open world perception task (e.g. perfectly separating the foreground from the background would still have the shortcoming of being unable to understand the scene it had just separated or the objects within). Others are in and of themselves potential paths to fully solving open world detection. However, the latter ones in doing so benefit from the building blocks that lay at their foundation.

Our selection of these domains is predicated on their immediate significance to the OWD paradigm. We present here each domain's progression throughout the years and conclude for each with a brief discussion on the domain's integral relationship to the overarching goals of OWD.


\subsection{Saliency Detection}\label{subsec2_1}

\begin{figure}[h]
    \centering
    \includegraphics[width=\linewidth]{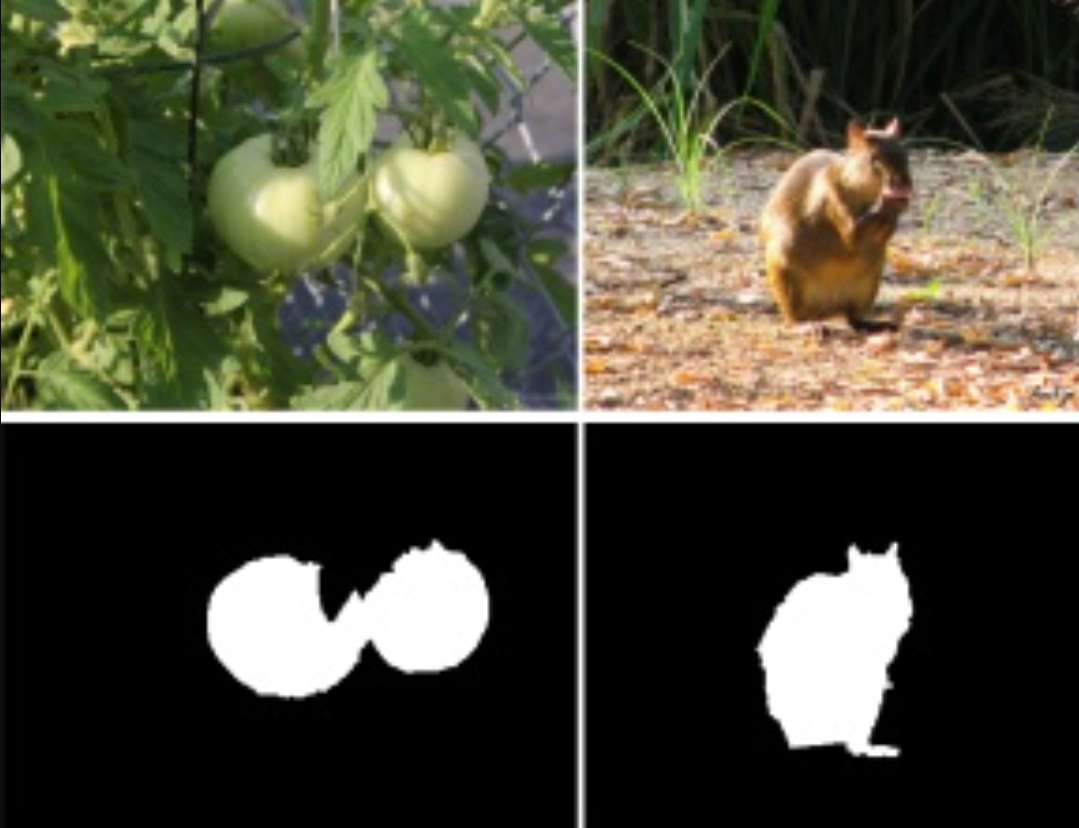}
    \caption{Saliency Detection on ECSSD \cite{shi2015hierarchical}. Top - input image, Bottom - ground truth mask}
    \label{fig:saliency}
\end{figure}

Saliency detection, in computer vision, is the process of identifying the most visually significant or attention-grabbing regions within an image. It aims to highlight areas that are likely to attract human attention, often based on factors like color, contrast, and uniqueness relative to the surrounding context. Fig.~\ref{fig:saliency} exemplifies saliency detection, with each image from the top having its saliency map counterpart displayed below it. In the context of OWD, saliency is not merely a sub-domain but a foundational pillar, an echo of the inherent human ability to prioritize the relevant and filter out the noise.
The domain has seen a significant advance over the years, from early heuristic approaches to modern deep learning methods, visual saliency has had an increasing success in drawing the boundaries between what we might view as relevant objects in a scene.

\subsubsection{Early Heuristic Models (1980s-2000s)}\label{subsubsec2_1_1}
The origins of computational saliency detection can be traced back to neurobiological and psychological studies, most notably Anne Treisman's Feature Integration Theory \cite{TREISMAN198097}, which described how humans process visual information by integrating different features — color, intensity, orientation — in parallel, and integrate them to form coherent object percepts. Building on these ideas, early computational models of saliency detection aimed to emulate this human visual attention mechanism. A notable early contribution was the Saliency Network model by \cite{730558}, which utilized a saliency map to emphasize significant regions based on these low-level features. It essentially mimicked the early visual cortex, calculating feature contrasts across different scales and combining them into a single map.

These early methods were largely heuristic, relying on simple image features to generate saliency maps. Techniques such as edge detection \cite{Rosin1995EdgesSM} and spectral residual analysis \cite{hou2007saliency} became popular, leveraging image properties that deviate from the norm to highlight salient regions.

\subsubsection{Incorporation of Context and Similarity (2000s-2010s)}\label{subsubsec2_1_2}
With the growth of large-scale image datasets and increased computational power, researchers began to incorporate context and inter-image similarities to improve saliency detection. \cite{marchesotti2009framework} introduced a method that utilized annotated databases to enhance saliency predictions by retrieving similar images, thus leveraging external knowledge. This approach marked a shift towards context-aware saliency detection, emphasizing the importance of considering the scene as a whole rather than isolated objects.

The introduction of models that capture contextual information, such as the context-aware saliency approach by \cite{goferman2012context}, aligned well with the emerging needs of OWD, where understanding the environment and its context is crucial for recognizing objects in dynamic and complex scenes.

\subsubsection{The Rise of Deep Learning (2010s-2020s)}\label{subsubsec2_1_3}
The arrival of deep learning brought about a paradigm shift in saliency detection, enabling significant advancements in both accuracy and applicability. Convolutional Neural Networks (CNNs) \cite{lecun1998gradient} became the backbone of modern saliency detection methods, capable of learning hierarchical representations from large datasets. Deep learning models demonstrated the ability to capture both fine-grained details and high-level contextual information, leading to more precise and robust saliency maps \cite{cornia2016deep}.

Among the notable advancements in this era was BASNet \cite{qin2019basnet}, which introduced a boundary-aware approach using a predict-refine architecture. This model focused not only on identifying salient regions but also on accurately predicting object boundaries, a crucial aspect for high-quality segmentation. Saliency models during this deep learning era shared many similarities with other parallel vision subdomains like object detection: deep convolutional networks, encoder/decoder architectures, two-stage paradigm to name a few.


Similarly, the Multi-Scale Interactive Network (MINet) \cite{pang2020multi} tackled the challenge of variable object scales by integrating multi-level and multi-scale features, enhancing the model's ability to detect salient objects of various sizes effectively.

\subsubsection{Integration of Multi-Modal Data (2020s-Present)}\label{subsubsec2_1_4}
The use of multi-modal data, particularly depth information, has further enhanced saliency detection capabilities. RGB-D saliency detection models \cite{zhou2021specificity} combine color (RGB) and depth (D) data, providing a richer representation of scenes. This integration allows models to utilize depth cues to better differentiate between foreground and background, significantly improving performance in challenging scenarios with complex backgrounds or low contrast. The comprehensive survey by \cite{zhou2021rgb} highlighted the progress and ongoing challenges in RGB-D saliency detection, suggesting that multi-modal approaches are critical for advancing the domain.

\subsubsection{Towards Open World Saliency (2020s-Present)}\label{subsubsec2_1_5}
The field is now moving towards a more inclusive understanding of saliency. Recent developments have focused on identifying all potentially salient objects within a scene, rather than just the most salient one. Techniques like PiCANet \cite{liu2020picanet}, which use pixel-wise contextual attention mechanisms, and approaches incorporating pyramid attention structures \cite{wang2019salient}, have shown promising results in highlighting multiple objects at varying levels of saliency, supporting thus more comprehensive scene understanding.

Furthermore, advancements in Generalised Co-Salient Object Detection (GCoSOD) \cite{liu2022generalised} are pushing the boundaries of saliency detection by allowing for the detection of common salient objects across images with varying degrees of noise and uncertainty. This progress is indicative of the field's evolution towards handling the diversity and complexity of real-world environments, the core challenge in OWD.

\subsubsection{Relationship to OWD}\label{subsubsec2_1_6}
In recent years, the distinctions between saliency detection and other visual detection fields have begun to blur, driven by the emergence of large multimodal models and integrated vision-language frameworks. As these comprehensive models gain prominence, saliency detection is increasingly viewed as a component of broader object detection tasks, where the need to isolate and highlight salient regions is naturally integrated into the overall perception model \cite{rs12091435}.
This trend suggests a future where saliency detection may not exist as a standalone task but rather as a feature embedded within more holistic perception models capable of addressing a wide range of open-world scenarios.

\subsection{Foreground/Background Separation}\label{subsec2_2}
\begin{figure}[h]
    \centering
    \includegraphics[width=\linewidth]{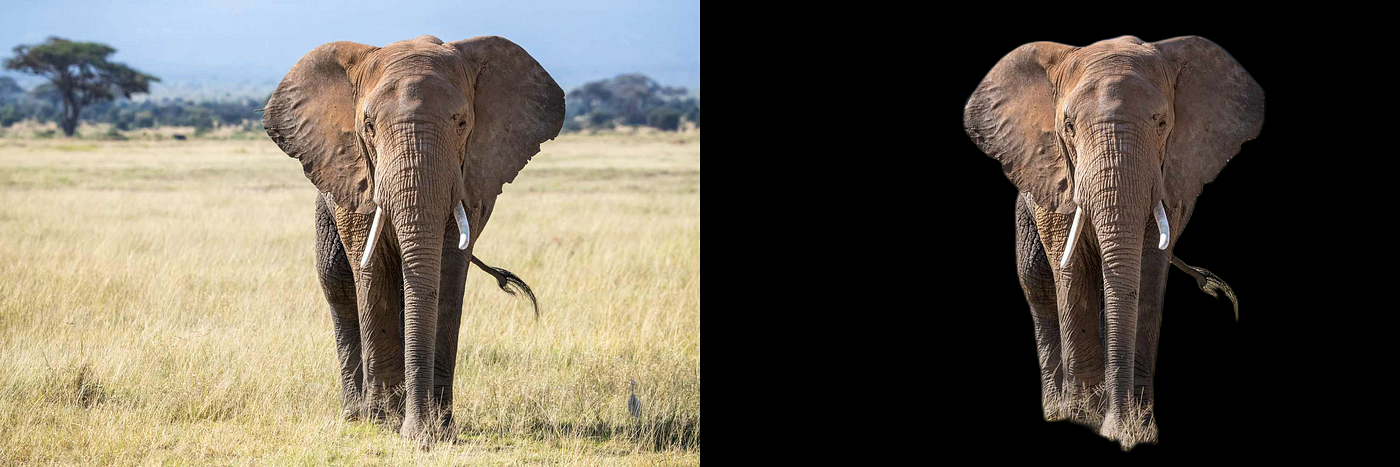}
    \caption{Forward - Background separation: Left - Original Image, Right: Resulting Separation}
    \label{fig:foreground-background}
\end{figure}

Foreground/background separation, also known as background subtraction or foreground detection, aims to segment an image or video sequence into two distinct regions: the foreground, which typically contains objects of interest, and the background (observed in black in Fig.~\ref{fig:foreground-background}), representing the relatively static or less important parts of the scene. This technique is foundational in many computer vision applications, such as surveillance, object tracking, and human-computer interaction. It also aligns with the goals of OWD, where detecting all objects within a scene, regardless of predefined categories, is of the essence.

Early methods in foreground/background separation operated under the assumption of static backgrounds, suitable for controlled environments. However, real-world scenarios, with varying lighting, camera motion, and background complexity, demanded more adaptive and robust solutions.

\subsubsection{Early Approaches (1980s - early 2000s)}\label{subsubsec2_2_1}
Initial methods for background subtraction were relatively simple yet effective for their time. One of the earliest approaches was the Running Gaussian Average, proposed by \cite{wren1997pfinder}. This method modeled each pixel as a Gaussian distribution, updating its mean and variance over time to accommodate slow changes in the scene. This computationally efficient method required minimal memory but struggled with complex or rapidly changing backgrounds.

\cite{koller1994towards} emphasized robustness in traffic automatic scene analysis, highlighting the need for selective updating of the background model. The proposed update mechanism prevented the background model from being corrupted by foreground objects.

Another straightforward yet effective method was the Temporal Median Filter, discussed by \cite{lo2001automatic} and further by \cite{cucchiara2003detecting}. This method maintained a buffer of recent frames and used the median pixel value to represent the background. While robust to noise and capable of handling transient changes, the temporal median approach required significant memory to store past frames and lacked adaptability for fast-changing environments.

In parallel to temporal analysis methods, techniques based on image intensity thresholding offered a different approach to foreground/background separation. Otsu's method \cite{4310076}, although originally designed for image binarization, could be adapted for foreground detection by assuming that foreground and background pixels exhibit distinct intensity distributions. By automatically calculating an optimal threshold that minimizes intra-class variance and maximizes inter-class variance, Otsu's method effectively segments the image into two regions. This technique provided a computationally inexpensive way to achieve background subtraction, particularly in scenarios where the intensity characteristics of foreground and background were sufficiently separable.

The Mixture of Gaussians (MoG) model came to help with dynamic backgrounds. Introduced by \cite{stauffer1999adaptive}, it addressed the limitations of earlier methods by modeling each pixel as a mixture of multiple Gaussian distributions. This approach could handle multi-modal background distributions, such as waving trees or water surfaces, by dynamically adapting the number and parameters of Gaussian components based on pixel value variations. MoG's flexibility made it highly effective in environments where background elements frequently changed.

The introduction of non-parametric methods like Kernel Density Estimation (KDE) by \cite{elgammal2000non} provided a more general approach to background separation by using kernels to approximate the probability distribution of pixel values, making no assumptions about the distribution's form. Although more computationally intensive, KDE offered superior adaptability to various background conditions.

\cite{oliver2000bayesian} introduced the concept of Eigenbackgrounds, utilizing Principal Component Analysis (PCA) to model the background. By projecting background variations into a lower-dimensional eigenspace, this method effectively captured the primary patterns of static scenes, isolating foreground objects even with significant background variability. Eigenbackgrounds leveraged spatial correlation across the image, offering a robust solution for scenarios with structured background changes.

\subsubsection{Robustness and Efficiency Focus (2000 - Present)}\label{subsubsec2_2_2}
Foreground/background separation techniques have evolved substantially since 2000, with early innovations establishing foundations that later work would build upon. \cite{li2004statistical} proposed an influential Bayesian framework that incorporated spectral, spatial, and temporal features to model complex backgrounds. This approach used principal features at each pixel to make robust foreground/background classification decisions, adapting well to both gradual and sudden changes - a versatility that would become a necessity for more advanced models. Following this, \cite{kim2005real} introduced the codebook model, providing an efficient way to handle structural background variations over long video sequences. By quantizing background values into codebooks, this model could handle scenes with periodic motions, such as moving trees or changing illumination, while maintaining low memory usage.

Building on these early advances, subsequent research has further enhanced robustness, adaptability, and computational efficiency. Sobral and Vacavant \cite{sobral2014comprehensive} conducted a comprehensive evaluation of 29 background subtraction algorithms using the Background Models Challenge (BMC) dataset \cite{vacavant2013benchmark}. Their analysis highlighted the need for algorithms that can balance robustness with practical performance, particularly in terms of processing speed and memory requirements, which are critical for real-time applications.

One trend in more recent developments is the integration of saliency detection concepts into background modeling. \cite{zhu2014saliency} introduced a robust background measure, known as boundary connectivity, to optimize saliency detection. By characterizing the spatial layout of image regions relative to image boundaries, this approach provided more robust foreground segmentation, linking background modeling with saliency cues.

The importance of feature selection in enhancing background modeling was emphasized by \cite{bouwmans2018role}, who reviewed the role of different features in handling challenges such as illumination variation, camera motion, and occlusion. Their work showed that combining multiple features, including spectral, spatial, and temporal properties, can significantly improve the robustness and accuracy of foreground detection.

Recent methods have also leveraged deep learning to enhance foreground detection. \cite{wang2018foreground} proposed a dual multi-scale 3D fully-convolutional neural network that captures both spatial and temporal features. This approach proposed learning hierarchical representations that adapt well to both gradual and sudden changes in the scene, improving results in complex background modeling scenarios - a direct evolution from the early adaptive models that has significantly advanced the field's capabilities.

\subsubsection{Relationship to OWD}\label{subsubsec2_2_3}
Foreground/background separation techniques are closely related to saliency detection, as both aim to highlight regions of interest within an image. The integration of saliency cues into background modeling helps improve the detection of salient objects, thereby enhancing the robustness of OWD systems. By effectively separating foreground from background, these techniques allow OWD systems to separate all foreground objects from their surrounding background noise, irrespective of their potentially novel or unknown characteristics.

\subsection{Out-of-Distribution Detection}\label{subsec2_3}

\begin{figure}[h]
    \centering
    \includegraphics[width=\linewidth]{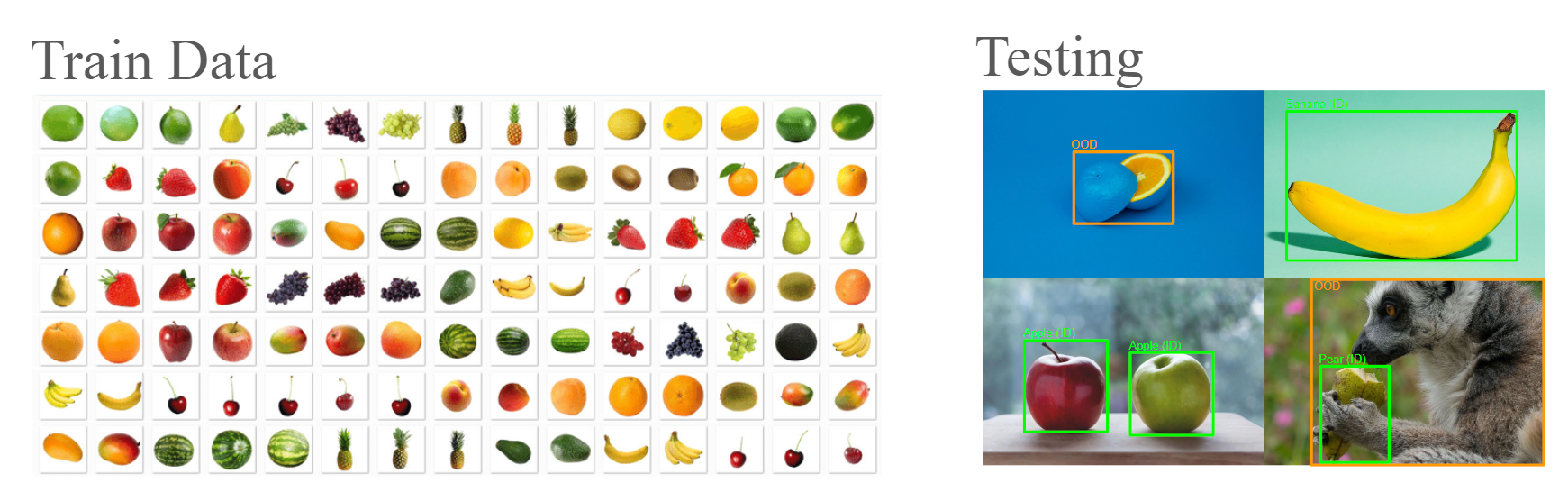}
    \caption{Out of Distribution (OOD) vs In Distribution (ID) for Object Detection}
    \label{fig:ood}
\end{figure}

Out-of-distribution (OOD) detection has fundamentally set out to equip a system with the ability to recognize and handle data that lies beyond the confines of its training experience. It focuses on distinguishing between familiar patterns and novel, unforeseen inputs that the model has never encountered before. As it stands for the domain of computer vision, OOD detection enables a model to identify when it is presented with visual data outside its training distribution, thereby preventing the model from making overconfident and potentially dangerous predictions in unfamiliar domains. For a clearer picture of what constitutes "familiar" (In-Distribution) versus "unfamiliar" (Out-of-Distribution) data, consider Fig.~\ref{fig:ood}, which provides visual examples in the context of object detection.

\subsubsection{Early Pre-Deep Learning Stages (Early 2000s)}\label{subsubsec2_3_1}
The conceptual seeds of OOD detection were planted in early works on anomaly and novelty detection. \cite{bishop1994novelty}'s seminal research on statistical methods for novelty detection and \cite{scholkopf2001estimating}'s development of a generalized support vector method provided a foundation by focusing on identifying deviations from a learned distribution. However, these approaches were not yet focused on the specific challenges that would later define OOD detection, as they were developed in a time when machine learning models and their applications were far less complex.

\subsubsection{Emergence with Deep Learning (Mid-2010s - Early 2020s)}\label{subsubsec2_3_2}

The field of OOD detection, as we recognize it today, began to crystallize with the advent of deep learning. The work by \cite{hendrycks2016baseline} demonstrated that even simple techniques, such as using the maximum softmax probability, could serve as a baseline for distinguishing OOD examples from in-distribution (ID) ones. This insight prompted a surge of research, as it became clear that OOD detection was not just a desirable feature but a necessity for deploying deep neural networks in real-world scenarios.

The reliance on softmax scores, however, presented limitations. Models often exhibited overconfidence in their predictions, even when faced with OOD inputs. To address this, \cite{liu2020energy} introduced an energy-based approach to OOD detection. By using an energy score aligned with the probability density of inputs, their method reduced the susceptibility to overconfidence and improved the differentiation between ID and OOD samples. This energy-based framework allowed for greater flexibility, serving as both a scoring function for pre-trained classifiers and a trainable cost function, thereby advancing the robustness of OOD detection.

Another significant contribution to the field came from \cite{liang2017enhancing}, who proposed ODIN (Out-of-DIstribution detector for Neural networks), which leverages temperature scaling and input pre-processing to improve OOD detection performance. These methods not only adjusted the model's confidence but also provided a clearer separation between ID and OOD inputs, furthering the understanding of how neural networks could better manage uncertainty.

\subsubsection{Recent Advancements and Expanding Scope (2020s-Present)}\label{subsubsec2_3_3}
As OOD detection evolved, it became evident that the problem extended beyond traditional classification boundaries. Researchers began exploring the limits of OOD detection in various modalities and settings. For instance, \cite{fort2021exploring} demonstrated the efficacy of large-scale pre-trained transformers \cite{vaswani2017attention} in enhancing OOD detection across different data modalities, underscoring the adaptability of these models to new tasks.

Furthermore, distance-based methods have gained traction, as illustrated by \cite{sun2022out}, who explored non-parametric nearest-neighbor approaches to OOD detection. Their findings highlighted the importance of flexibility and generality in detecting OOD samples, free from the constraints of strong distributional assumptions imposed by parametric methods. This approach demonstrated superior performance, reducing false positives and enhancing the robustness of OOD detection.

The comprehensive survey by \cite{yang2024generalized} has brought clarity to the field, proposing a unified framework for generalized OOD detection. This framework elegantly ties together related concepts such as anomaly detection (AD), novelty detection (ND), open set recognition (OSR), and outlier detection (OD). By framing these problems as sub-tasks under the broader OOD umbrella, this work has laid the groundwork for a more coherent understanding of how various detection challenges interrelate, and how insights from one area can inspire advances in another.

\subsubsection{Relationship to OWD}\label{subsubsec2_3_4}
Out-of-distribution detection forms a critical foundation for OWD systems. Where traditional closed-world models fail by confidently misclassifying novel objects into known categories, OOD detection provides the essential capability to recognize when an input falls outside the training distribution. This "unknown awareness" is precisely what enables OWD systems to identify novel objects that require either human intervention or incremental learning. The techniques developed for OOD detection—from energy-based approaches to nearest-neighbor methods—directly inform how OWD systems separate unknowns from knowns, creating a crucial bridge between closed-set and truly open-world perception. Furthermore, as \cite{yang2024generalized} demonstrated in their unified framework, OOD detection not only serves OWD directly but also creates important connections to related fields like anomaly detection and open set recognition, fostering a more integrated approach to the broader challenge of open-ended perception.

\subsection{Zero-Shot Detection}\label{subsec2_4}

While Out-of-Distribution detection focuses on identifying whether an input falls outside a model's training distribution (flagging 'unknowns' without further classification), Zero-Shot Detection (ZSD) takes a fundamentally different approach: it aims to both detect and classify objects from entirely unseen categories by leveraging semantic knowledge. Rather than merely recognizing something as unknown, ZSD seeks to bridge the gap to novel classes through transferable semantic relationships, typically encoded in language or attributes.

\begin{figure}[h]
    \centering
    \includegraphics[width=\linewidth]{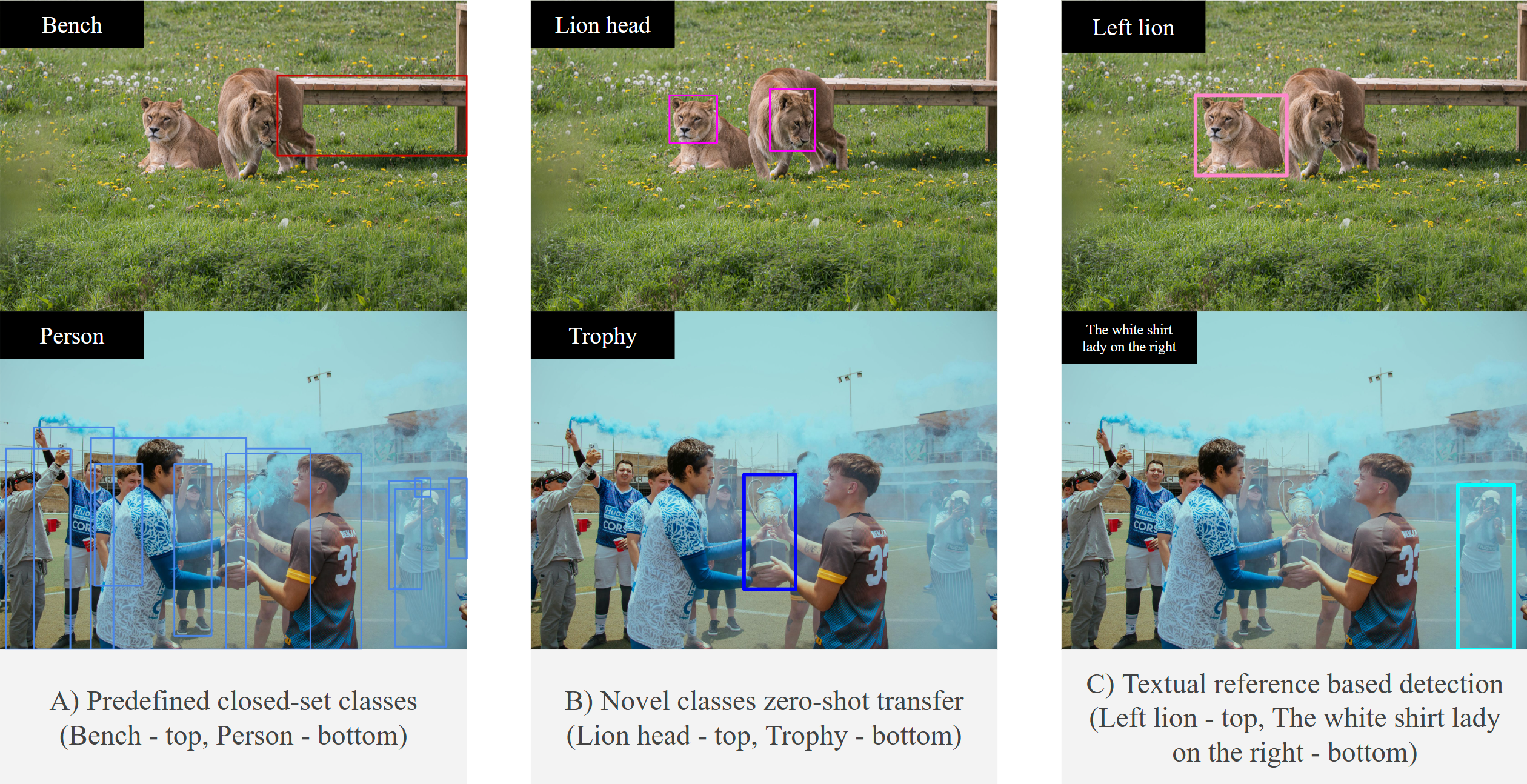}
    \caption{Idea behind Grounding Dino~\cite{liu2025grounding}- Zero Shot Detection. Text-based zero-shot prompting permits novel visual detections}
    \label{fig:zero-shot}
\end{figure}

The origins of Zero-Shot Detection are deeply rooted in the broader field of zero-shot learning, which seeks to overcome one of the most profound challenges in machine learning: the ability to recognize and understand objects or concepts without explicit prior exposure. This quest began well before the recent surge in interest, driven by the recognition that real-world applications often require models to identify and adapt to new, unseen categories, akin to a child recognizing an object based on a description rather than direct experience. Such description-based inference can be seen in Fig.~\ref{fig:zero-shot}.

\subsubsection{Early Foundations (Pre-2017)}\label{subsubsec2_4_1}

Zero-shot learning initially focused on classification tasks, where models leveraged semantic representations, such as attributes or word embeddings, to bridge the gap between seen and unseen classes. As highlighted by \cite{xian2018zero} in their survey, foundational methods such as Direct Attribute Prediction (DAP) and Indirect Attribute Prediction (IAP) demonstrated how shared attributes could classify unseen objects. However, these early methods were limited in scope, typically handling single, dominant object categories within controlled environments.

The leap from classification to detection introduced the dual challenges of recognition and localization. Early methods like Attribute Label Embedding (ALE) \cite{akata2015label} and Deep Visual Semantic Embedding (DEVISE) \cite{frome2013devise} explored the projection of visual features into semantic spaces. These approaches laid the groundwork for zero-shot detection by showing how visual-semantic embeddings could generalize beyond the classes encountered during training. However, the task of detecting objects amidst the clutter and complexity of real-world scenes required more sophisticated methodologies.

\subsubsection{The Rise of Explicit Zero-Shot Detection (2017-2021)}\label{subsubsec2_4_2}

While the seeds of zero-shot detection were sown early, it was not until the later part of the 2010s that the field began to flourish. The work of \cite{bansal2018zero}, which framed zero-shot detection explicitly, was instrumental in this evolution. They introduced techniques to adapt visual-semantic embeddings for detection tasks, emphasizing the importance of handling background variability and introducing methods like background-aware training to improve robustness. This important framing was followed by \cite{rahman2018zero}'s \textit{Zero-Shot Object Detection: Learning to Simultaneously Recognize and Localize Novel Concepts}, which proposed an end-to-end deep network that jointly models visual and semantic information, using novel loss functions to achieve synergy between max-margin class separation and semantic space clustering.

The field saw the introduction of methods that utilized textual descriptions to improve detection accuracy. \cite{li2019zero} presented a framework that incorporated natural language descriptions, enhancing the ability to detect and recognize novel objects by aligning visual and textual modalities at a granular level. This approach represented a shift towards more nuanced models capable of understanding and using detailed linguistic information to inform visual tasks.

The incorporation of attributes to aid in detection was also explored in \cite{bansal2018zero}, emphasizing the role of attribute-based semantic embeddings.

\subsubsection{The Unseen and Contrastive Learning (2020-2022)}\label{subsubsec2_4_3}
A notable development in this period was the shift from mapping visual features to semantic spaces to synthesizing visual features for unseen classes. Hayat et al.'s work, \textit{Synthesizing the Unseen for Zero-shot Object Detection} \cite{hayat2020synthesizing}, proposed generating synthetic features for unseen classes using class semantics, thus training detection models that better represent both seen and unseen objects. \cite{zhu2020don}'s \textit{Don't Even Look Once} (DELO) further refined this approach by introducing algorithms that synthesized visual features for unseen objects, which were then used to augment training. These methods addressed the inherent skew towards seen classes by providing the model with a richer and more diverse training set that included synthesized instances of unseen objects.
The introduction of contrastive learning mechanisms \cite{chen2020simple} also marked a significant advancement in zero-shot detection. The Semantics-Guided Contrastive Network, or ContrastZSD, developed by \cite{yan2022semantics}, employed contrastive learning to enhance the separation of visual features for different classes. By leveraging both region-category and region-region contrastive tasks, this approach reduced bias towards seen classes and optimized the alignment of visual and semantic spaces. These innovations underscored the importance of not merely recognizing unseen objects but doing so in a way that maintains the structure and integrity of the learned feature space.

\subsubsection{Current Trends and Future Directions: A New Paradigm}\label{subsubsec2_4_4}

The integration of large-scale multimodal pre-training, as seen in recent works like \textit{Grounding DINO} \cite{liu2025grounding}, highlights the importance of grounding visual learning within a broader context—much like how humans draw upon language and experience to make sense of their surroundings, or how we have seen context infusion advancing the domain of saliency detection. The pursuit of ZSD is, at its core, an attempt to transcend the limitations of conventional training paradigms, aiming to equip machines with a more human-like capacity to generalize from the known to the unknown. It is this vision of adaptable, intelligent perception that drives the ongoing innovation within the field, challenging us to keep moving the goal posts of what vision models strive to achieve.

\subsubsection{Relationship to OWD}\label{subsubsec2_4_5}
The advancements made in the past decade in ZSD underscore an important evolution: from simply classifying unseen objects to effectively detecting and understanding them in their complex real-world environments. This trajectory reflects not just a technical progression - where new models, architectures, methodologies advance SOTA - but a conceptual one, where the boundaries between different modalities of understanding—visual, linguistic, contextual—are increasingly blurred. As research continues to develop and intertwine these capabilities, the goal of machines that can see, understand, and adapt with fluidity similar to human perception draws closer to reality.


This leap in generalization is powerfully demonstrated by foundational models like DINOv2 \cite{oquab2023dinov2}. Without being fine-tuned for specific downstream applications, its largest variant achieves state-of-the-art zero-shot results across a remarkable range of vision tasks, attaining, for instance, 86.5\% top-1 accuracy on ImageNet-1k classification, 65.1 mIoU on PASCAL VOC segmentation, and a 0.9 R-MSE on NYUv2 depth estimation. Crucially, Oquab et al. show a consistent trend: as model scale increases, so does its zero-shot performance across all tested domains. In a similar display of emergent ability, the Segment Anything Model (SAM) \cite{kirillov2023segment} can produce highly detailed and accurate edge maps on the BSDS500 benchmark \cite{MartinFTM01}, despite having never been trained on edge prediction, illustrating the latent capabilities of modern vision models in open-world contexts.


\begin{figure}[h]
    \centering
    \includegraphics[width=\linewidth]{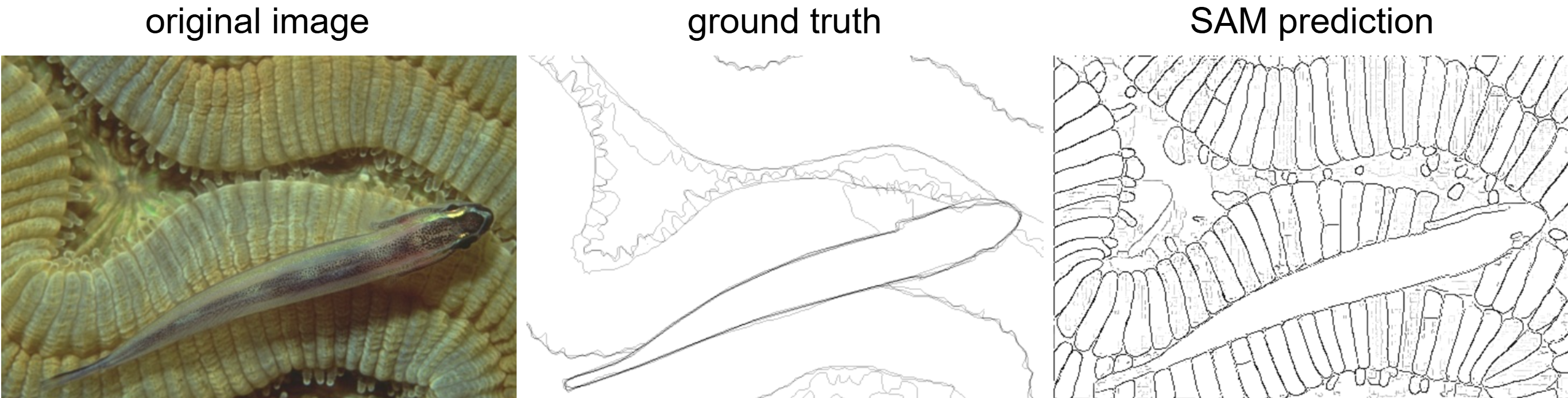}
    \caption{Zero-shot edge prediction on BSDS500 data \cite{MartinFTM01}. SAM \cite{kirillov2023segment}  had no access to BSDS images or annotations during training and was never trained to predict edge maps}
    \label{fig:ZeroShotEdgesSam}
\end{figure}

\subsection{Object Detection}\label{subsec2_5}
Traditional object detection, as comprehensively reviewed by \cite{zou2023object} and \cite{zhao2019object}, has undergone a significant evolution over the past two decades.

\subsubsection{Early Object Detection (Pre-Deep Learning - Early 2010s)}\label{subsubsec2_5_1}
The field initially relied on handcrafted features, with early approaches exemplified by the Viola-Jones detector \cite{viola2001rapid} and the Histogram of Oriented Gradients (HOG) detector \cite{dalal2005histograms}. These methods laid the groundwork for object detection by introducing effective feature extraction techniques for specific tasks.

The Deformable Parts Model (DPM) \cite{felzenszwalb2009object} then incorporated part-based representations and introduced flexibility in handling object deformation. However, the real paradigm shift came with the advent of deep learning, starting with the introduction of R-CNN \cite{girshick2014rich}, which allowed models to learn robust and discriminative features directly from raw image data.

\subsubsection{Deep Learning in Closed-Set Detection (Mid 2010s - Late 2010s)}\label{subsubsec2_5_2}
Building on this breakthrough, subsequent advancements such as Spatial Pyramid Pooling Networks (SPPNet) \cite{he2015spatial} and Fast R-CNN \cite{girshick2015fast} improved both the efficiency and accuracy of object detection frameworks. SPPNet introduced a pooling strategy that eliminated the need for fixed-size inputs, thereby improving computational efficiency and robustness to scale variations. Fast R-CNN streamlined training and inference by sharing convolutional features across region proposals, leading to marked improvements in both detection accuracy and processing speed.

Region Proposals: A Glimpse into the Open World

It is in this era that Faster R-CNN \cite{ren2016faster} introduced the Region Proposal Network (RPN), significantly speeding up the detection process by generating region proposals directly within the neural network. This region proposal network in itself can be seen as a precursor to OWD, for its goal was that of proposing the regions, similarly to saliency detection, that are most likely to contain objects of interest. The class-agnostic nature of RPNs aligns with the requirements of OWOD, where the model must detect objects regardless of whether they belong to known or unknown classes.

Despite these advancements, most traditional object detection methods continue to operate within a \textit{closed-set paradigm}, where the models are limited to recognizing only the predefined set of classes seen during training. This limitation hinders their ability to adapt to real-world scenarios where unknown objects frequently appear. To address this challenge, domains such as Open Vocabulary Detection (OVD) and Open World Object Detection (OWOD) have emerged, aiming to broaden the detection capabilities of models and enable them to handle the vast diversity of objects encountered in dynamic environments.

\subsubsection{From Open Vocabulary Detection to Open World Detection (Late 2010s - Early 2020s)}\label{subsubsec2_5_3}

Open Vocabulary Detection (OVD) can be seen as a precursor to OWOD, focusing on expanding detection capabilities beyond predefined categories. OVD models are trained using a limited set of labeled categories but are expected to generalize to recognize and detect a broader set of objects. For example, \cite{zhu2024survey} provide a comprehensive survey on OVD, outlining methodologies such as visual-semantic space mapping, novel visual feature synthesis, and region-aware training. These approaches enable models to utilize weak supervision signals and large-scale image-text pairs, enhancing their ability to recognize unseen objects.

The concept of open set detection closely aligns with the objectives of OVD and OWOD. \cite{dhamija2020overlooked} formalized the problem of open set object detection, highlighting the limitations of current detectors in handling unknown objects. Their study revealed that many state-of-the-art detectors incorrectly classify unknown objects as one of the known categories with high confidence, demonstrating the need for open set protocols and evaluation metrics.

Recent works such as \textit{PromptDet} \cite{feng2022promptdet} have further advanced OVD by integrating a text encoder from pre-trained vision-language models to classify class-agnostic object proposals. This integration helps align the visual latent space of object proposals with the textual embedding space. Similarly, \cite{minderer2022simple} utilize a Vision Transformer architecture with large-scale pre-training to achieve high performance in zero-shot and one-shot object detection scenarios. These developments demonstrate the potential of leveraging weakly supervised data to achieve open vocabulary capabilities, which in turn lays the groundwork for OWOD.

\subsubsection{Formalization of Open World Object Detection (Early 2020s - Present)}\label{subsubsec2_5_4}

\begin{figure}[h]
    \centering
    \includegraphics[width=\linewidth]{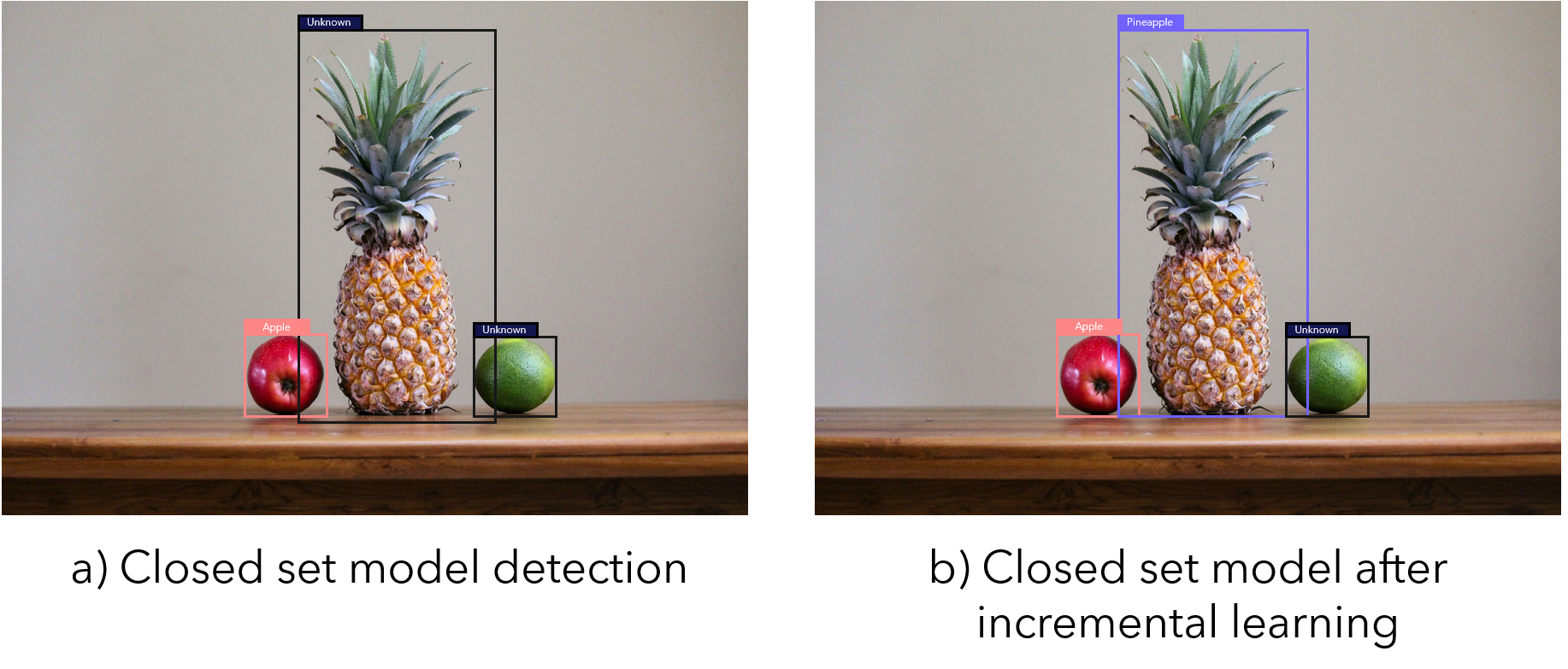}
    \caption{Open World Object Detection
    a) before incremental learning
    b) after pineapple class knowledge is added through incremental learning}
    \label{fig:owod}
\end{figure}

The formalization of OWOD as a dedicated problem setting was marked by the seminal work "Towards Open World Object Detection" \cite{joseph2021towards}, which defined the objectives and challenges unique to OWOD. This work introduced a comprehensive evaluation protocol and proposed the ORE (Open World Recognition Engine) framework, leveraging contrastive clustering, an unknown-aware proposal network, and an energy-based unknown identifier. These components allow OWOD systems to dynamically manage known and unknown object categories - Fig.~\ref{fig:owod}, making significant strides toward realizing a more adaptive object detection framework.

Following this groundwork, several studies have contributed to advancing OWOD. \cite{zhao2023revisiting} revisited the OWOD problem, refining methods for handling background clutter and improving the robustness of unknown object detection. \cite{ma2023cat} proposed the \textit{CAT} model, a cascade detection transformer that employs localization and identification mechanisms to progressively refine object proposals, effectively distinguishing unknowns from known categories. These works emphasize the importance of accurately identifying unknown objects and minimizing misclassification, which are critical challenges in OWOD.

\subsubsection{Open World Segmentation: Extending OWOD to Pixel-Level Understanding (Early 2020s - Present)}\label{subsubsec2_5_5}

\begin{figure}[h]
    \centering
    \includegraphics[width=\linewidth]{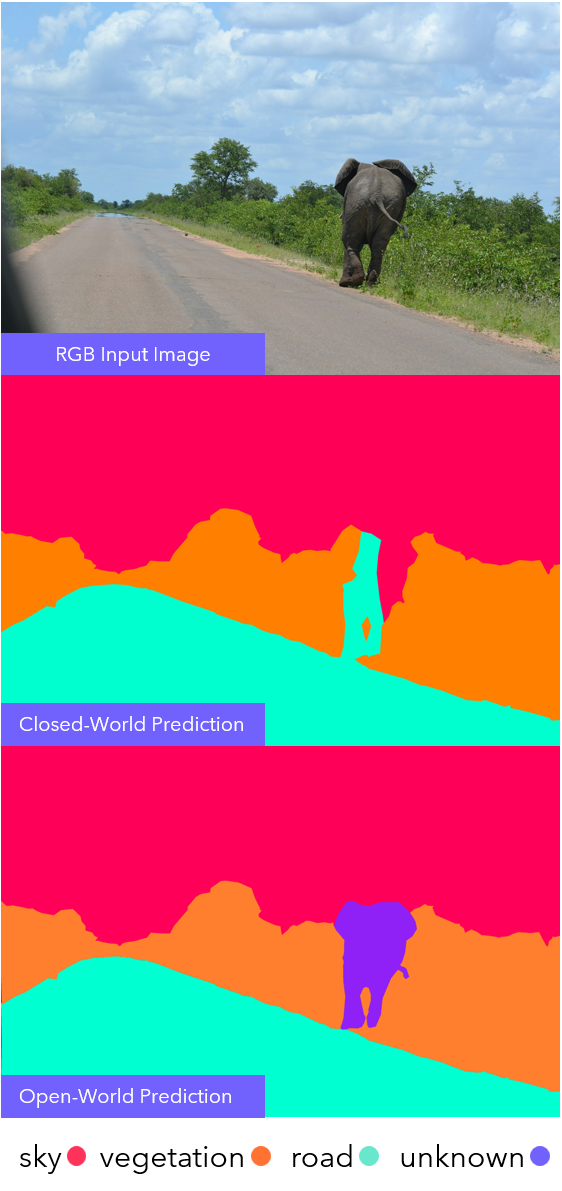}
    \caption{Open World Segmentation. Adding the concept of unknown reduces known classes noise}
    \label{fig:pixlevel}
\end{figure}

While OWOD focuses on detecting objects as bounding boxes, Open World Segmentation (OWS) extends this concept to the pixel level (Fig.~\ref{fig:pixlevel}, aiming to segment all objects within a scene, regardless of their known or unknown status. This finer-grained understanding of the visual world complements OWOD by providing detailed object boundaries and shapes \cite{sodano2024open}.

OWS inherits the challenges of OWOD, such as distinguishing unknowns from the background and known classes, while also addressing the complexities of segmentation, including handling object boundaries and diverse shapes. And, similarly to how other precursor domains contributed to OWD through sharing technique advancements, so does OWD, perhaps to an ever greater extent, contribute to OWS. Early efforts in OWS focused on adapting existing semantic and instance segmentation methods to open-world scenarios. For instance, \cite{wang2021unidentified} introduced UVO (Unidentified Video Objects), a benchmark for open-world class-agnostic object segmentation in videos, highlighting the need for models capable of segmenting novel objects in dynamic environments.

Approaches like Deep Metric Learning for Open World Semantic Segmentation \cite{cen2021deep} leverage contrastive clustering to enable open-set semantic segmentation, effectively detecting both in-distribution and OOD objects. The integration of Vision-Language Models (VLMs) has also shown promise in OWS. \cite{liu2022open} proposed ViL-Seg, an open-world semantic segmentation pipeline that utilizes image-caption data to learn to segment objects of various categories without dense annotations, demonstrating the potential of leveraging weakly supervised data for OWS.

Furthermore, techniques like exploiting pseudo ground truth from learned pairwise affinity \cite{wang2022open} have advanced open-world instance segmentation, enabling the grouping of pixels into object instances without pre-determined taxonomy.  Recently, the Segment Anything Model (SAM) \cite{kirillov2023segment} and its successor, SAM 2 \cite{ravi2024sam}, have demonstrated impressive zero-shot segmentation capabilities by leveraging a massive dataset of masks and a promptable transformer architecture.

\subsubsection{Relationship to OWD}\label{subsubsec2_5_6}

Addressing the core challenges of OWOD involves developing strategies for:

Distinguishing unknowns from the background: Accurately identifying potential unknown objects amidst background clutter requires sophisticated proposal generation and unknown identification mechanisms, as highlighted by the foundational and follow-up studies \cite{joseph2021towards, zhao2023revisiting}.

Discriminating unknowns from known classes: Preventing unknown objects from being misclassified into known categories is crucial for maintaining the integrity of the model's knowledge base. Approaches like contrastive clustering and energy-based identification have been pivotal in this regard \cite{joseph2021towards, ma2023cat}.

Incrementally learning new classes: OWOD systems must seamlessly integrate new knowledge about previously unknown classes without forgetting what has already been learned. This challenge necessitates the development of efficient and robust incremental learning techniques, a topic explored in various works \cite{joseph2021towards, mullappilly2024semi, shaheen2023framework}.

Addressing the inherent challenges of Open World Object Detection—namely, distinguishing unknowns from background clutter, discriminating them from known classes, and incrementally learning new categories—is paramount for advancing truly open perception.  OWOD provides a critical framework and initial methodologies for tackling these complexities, directly contributing to the broader goals of OWD.  While OWOD marks substantial progress, the integration of Vision-Language Models, as explored in the subsequent section, offers a promising trajectory for even more robust and versatile open world detection systems.

\subsection{Vision-Language Models (VLMs)}\label{subsec2_6}

Vision-Language Models (VLMs) stand at the intersection of computer vision and natural language processing, providing unified frameworks that handle and interpret both visual and textual inputs. While Large Language Models (LLMs) excel in text-based tasks, the domain of computer vision continues to demand solutions capable of detecting, localizing, and understanding novel visual entities. VLMs bridge this gap by mapping images (or video frames) and text into a shared space, greatly benefitting any approach in search of open-ended visual perception. Below, we outline the evolution of VLMs, highlight new research trends (both open and commercial), and discuss how these models support a more comprehensive and flexible approach to visual understanding.

\subsubsection{Early Multimodal Approaches (Pre-2020)}\label{subsubsec2_6_1}
Early work combining image and text modalities focused primarily on image captioning \cite{vinyals2015show} or visual question answering (VQA) \cite{kiros2014unifying}, using CNNs for image features and RNNs \cite{rumelhart1986learning} or LSTMs \cite{hochreiter1997long} for text. These models, while illustrating proof-of-concept multimodal abilities, were constrained by the limited scale of image-text corpora.
\subsubsection{The Rise of Large-Scale Pre-training (Mid-2010s - Early 2020s)}\label{subsubsec2_6_2}
A turning point arose with the introduction of large-scale image-text pre-training and contrastive objectives, most prominently in CLIP \cite{radford2021learning} and ALIGN \cite{jia2021scaling}. By embedding both images and textual descriptions into a shared high-dimensional space, these models demonstrated robust zero-shot recognition capabilities. Subsequent works like FILIP \cite{yao2021filip} refined the alignment process with finer region-word correspondence, improving performance on tasks requiring more granular visual understanding.

\subsubsection{Self-Supervised Transformers and Emergent Properties (Early 2020s)}\label{subsubsec2_6_3}
Another significant advance stems from applying self-supervised learning (SSL) to Vision Transformers (ViTs). DINO \cite{caron2021emerging} demonstrated how self-distillation techniques can lead ViTs to learn salient object boundaries, often without explicitly being taught segmentation. DINOv2 \cite{oquab2023dinov2} scaled this approach with improved data curation and larger models, showing that SSL yields highly transferable features for downstream tasks.

These self-supervised ViTs often serve as backbones in broader vision-language pipelines. Their emergent object-centric representations align naturally with textual cues, particularly when combined with contrastive or caption-based objectives in the latter stages of training.

\subsubsection{Architectural Advancements and Ecosystem Overview (Early 2020s - Present)}\label{subsubsec2_6_4}
With the growing demand for richer multimodal modeling, VLM architectures have diversified significantly:

    Encoder-Free Paradigms - as exemplified by EVE \cite{diao2024unveiling}, an encoder-free approach processes raw image patches directly in a decoder, doing away with a specialized vision encoder. Despite the architectural simplicity, these models reach competitive performance via LLM-centric “pre-aligning” and visual feature supervision stages.

    Region-Text Alignment for Detection - methods like YOLO-World \cite{cheng2024yolo} meld region-level detection ideas with text encoders (often from CLIP or similar). By optimizing for region-text contrastive losses, these systems enhance open-vocabulary detection, allowing real-time or near real-time inference of previously unseen object categories.

    Generative Vision Encoders - Florence-VL \cite{chen2024florence} expands beyond contrastive training by integrating a “generative vision foundation model,” capturing different semantic and structural aspects of images. A novel \emph{depth-breadth fusion} (DBFusion) combines features from multiple layers (“depth”) and prompts (“breadth”), yielding richer object-level details.

    Scaling Up Vision Backbone Size - InternVL \cite{chen2024internvl,chen2024far} and Qwen-VL \cite{bai2023qwen} scale their vision modules to billions of parameters, aligning them with similarly large LLMs. Meanwhile, PaliGemma \cite{beyer2024paligemma,steiner2024paligemma} underscores the importance of a carefully designed vision encoder paired with a robust text model, open-sourcing the code, model weights, and training procedures.

Such diverse architectural designs reflect the broader ecosystem: each model family may specialize in specific tasks (e.g., zero-shot detection, video understanding, dense segmentation) or general-purpose capabilities (e.g., image captioning, chat-like multimodal interaction).

As with LLMs, the VLM space comprises both open and commercial closed solutions. Closed platforms (Anthropic Claude, Google Gemini, OpenAI GPT-family, etc.) are recognized for high performance but release minimal details on architecture, hyperparameters, and datasets. In contrast, open models (e.g., Meta Llama \cite{touvron2023llama}, InternVL, PaliGemma, Qwen-VL) generally disclose more about their technical internals and training data and are thus more amenable to reproducibility and fine-grained research. A notable example is how Google, while developing proprietary models like Gemini, also supports open science with publications such as PaliGemma.

This dual landscape exposes a tension between commercial competitive edges—where advanced architectures and large-scale data handling methods may remain proprietary—and the academic ideal of open research.

A core advantage of VLMs is their natural fit for open-ended visual understanding. Because they encode text and images into a partially shared embedding space, they can handle queries and tasks specified in language (e.g., “Find anything unusual or unknown”), much like large language models handle text-based knowledge queries. This attribute aligns closely with the mission of open-world or open-vocabulary detection, wherein the model must detect objects not necessarily encountered during training.

Recent works highlight how VLMs expand the boundaries of what vision systems can detect or recognize:
    Unseen Object Generalization: By pairing region proposals with textual embeddings of unseen categories, YOLO-World \cite{cheng2024yolo} and PromptDet \cite{feng2022promptdet} enable zero-shot or few-shot detection of classes absent in training.
    Cross-Task Transfer: VLMs frequently excel at tasks beyond bounding-box detection (e.g., segmentation, retrieval, image captioning) due to shared multimodal embeddings. PaliGemma 2 \cite{steiner2024paligemma}, for instance, demonstrates state-of-the-art performance on specialized tasks like molecular structure or music score recognition, illustrating how a broad vision-language foundation can transfer to niche scenarios.

As a result, VLMs provide a powerful means to infuse open-ended perception with the semantic richness of language, particularly for detecting and describing novel objects under minimal additional supervision.

\subsubsection{Instruction Tuning and Multimodal Reasoning (Early 2020s - Present)}\label{subsubsec2_6_5}
\begin{figure}[h]
\centering
\includegraphics[width=\linewidth]{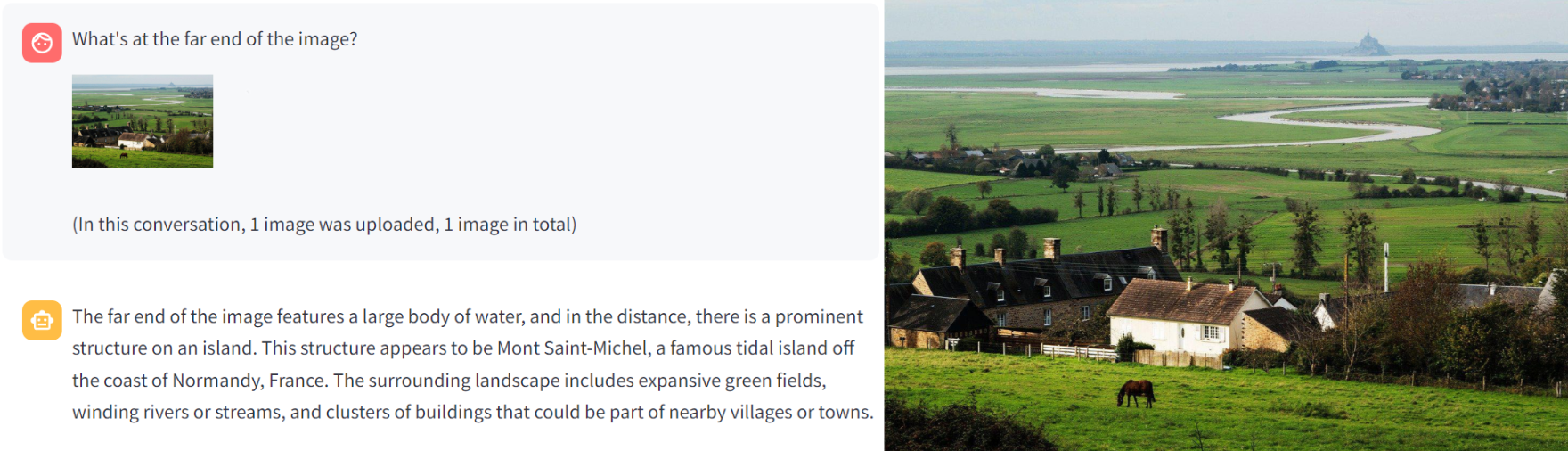}
\caption{Vision Language Model - InternVL2 \cite{chen2024internvl, sodano2024open}}
\label{fig:vlms}
\end{figure}

An emergent development is the alignment of VLMs with instruction tuning protocols, often inspired by language-only GPT techniques. Models such as LLaVA \cite{liu2024visual}, InternVL2 \cite{chen2024expanding}, and Qwen-VL-Chat \cite{bai2023qwen} incorporate multi-turn dialogue training, chain-of-thought (CoT) reasoning \cite{wei2022chain}, and instruction-following data. This transforms VLMs into interactive agents that can discuss images (Fig.\ref{fig:vlms}), ground or localize objects, answer fine-grained queries, and even read textual information in images. When combined with open-ended detection objectives, instruction-tuned VLMs can be prompted to \emph{discover} novel objects or emphasize specific attributes of interest. Multimodal models that arise out of such fusion between Visual Models and Large Language Models often get referred to as Visual Large Language Models (VLLMs).

Such instruction-tuned VLMs mark a shift from static “one-shot” inference to dynamic, iterative exploration of scenes in response to user queries—essential for real-world deployments where continuous interaction and complex reasoning are required.

Towards a More Open World Through VLMs

Finally, as research on VLMs progresses, challenges remain in areas of bias, domain adaptation, interpretability, and computational cost—especially as models scale into billions of parameters \cite{bai2023qwen}. Nonetheless, the synergy between large-scale self-supervised vision backbones and powerful language encoders fosters unprecedented versatility. From purely open detection of new objects to grounded reasoning in novel domains (medical imaging, robotics, remote sensing), VLMs are poised to serve as a central axis of progress. By tightly integrating textual semantics with high-level visual features, VLMs effectively promote a more open, context-aware mode of visual perception—laying a robust foundation for the next stage in open-world detection research.

\vspace{0.3em}
\noindent
Unified Interactive Segmentation and Embedding Interfaces.
Beyond the broad class of VLMs discussed above, several recent studies propose specialized frameworks that emphasize unified or interactive segmentation and embedding approaches, further strengthening open-ended understanding:

    SEEM \cite{zou2024segment} presents a promptable and interactive model designed for “segmenting everything everywhere all at once,” introducing a novel decoder that allows for diverse prompts (e.g., points, boxes, scribbles) and text queries to be integrated in a single universal segmentation interface.
    
    X-Decoder \cite{zou2023generalized} proposes a generalized decoding model that seamlessly predicts pixel-level segmentation and language tokens, thus supporting multiple segmentation tasks (open-vocabulary, referring segmentation, etc.) within a common semantic space.
    
    FIND \cite{zou2023interfacing} introduces a lightweight transformer interface for foundation models’ embeddings, focusing on interleaving modalities and tasks (segmentation, retrieval, grounding) without tuning any foundation model weights. This interface aligns visual and language embeddings, enabling robust cross-modal reasoning.

\subsubsection{Relationship to OWD}

Collectively, these demonstrate the growing importance of flexible architectures that unify segmentation, detection, and language-based tasks within a single pipeline. They also illustrate a trend toward promptable or interactive designs, where minimal additional supervision—or even user-supplied scribbles—can guide the model toward focusing on specific objects or attributes in an open-world setting.



\section{Datasets}\label{sec3}
While conventional detection tasks typically rely on a fixed set of categories with clearly defined training and testing splits, OWD calls for data that captures unknown or out-of-distribution entities in diverse, realistic settings. Moreover, the advent of large-scale Vision-Language Models (VLMs) has ushered in a new paradigm for learning open-ended representations from colossal image-text corpora, where the new comprehensive visual capabilities of models get evaluated on standalone benchmarks aimed to test across multiple dimensions. In this section, we review both established datasets frequently used for \emph{training} and \emph{evaluation}, and those designed for \emph{pre-training} and \emph{benchmarking} under open-world scenarios.

\subsection{Task-Specific Datasets: Training and Evaluation}\label{subsec3_1}
In the traditional paradigm of computer vision, datasets are often curated for specific tasks, with clear distinctions between training and evaluation sets. These datasets have been instrumental in the progress of object detection, particularly in closed-set scenarios.

\subsubsection{Training Datasets}\label{subsubsec3_1_1}
These datasets are designed for supervised training of detection models, focusing on learning to identify and localize a predefined set of object categories.

For a considerable period, MS-COCO~\cite{lin2014microsoft} has been a cornerstone dataset for object detection. It encompasses 1.5 million object annotations across 80 categories, representing diverse real-world scenes. Although inherently a closed-set dataset, COCO serves as a foundation for numerous OWD frameworks, often adapted or extended to incorporate unknown objects.
\\[3pt]
Objects365~\cite{shao2019objects365} expands upon this with an even larger scope, featuring 365 annotated categories and over 10 million bounding boxes within more than 600k training images. Introduced to enhance localization-sensitive tasks, Objects365 pre-training has demonstrated significant improvements in downstream COCO detection performance and facilitates transfer learning to other perception tasks.

\subsubsection{Evaluation Datasets}\label{subsubsec3_1_2}
The evaluation of visual models, especially in the earlier paradigm, was heavily centered on assessing core visual recognition capabilities. These benchmarks primarily targeted tasks like image classification, object detection, and segmentation.

\begin{figure}[h]
    \centering
    \includegraphics[width=\linewidth]{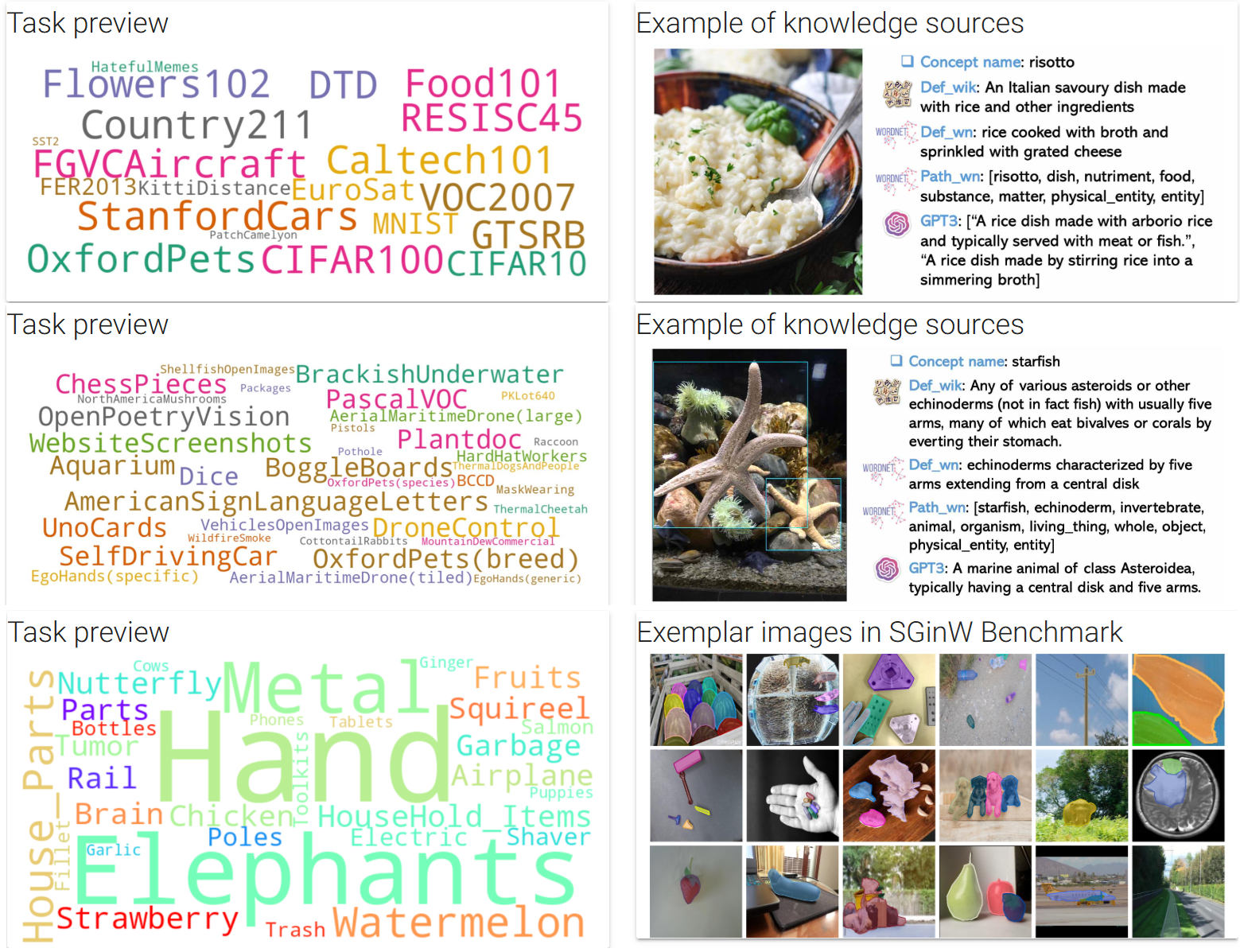}
    \caption{Computer Vision in the Wild \cite{li2022elevater,zou2023generalized}}
    \label{fig:cvitw}
\end{figure}

Traditional benchmarks like PASCAL VOC~\cite{Everingham15}, MS-COCO~\cite{lin2014microsoft}, and LVIS~\cite{gupta2019lvis} are still extensively used for evaluating detection performance on known classes, often adapted for open-world protocols. COCO2014 and COCO2017 feature 80 categories, while LVIS extends to over 1200 classes, providing a wider range of object categories for evaluation.
\\[3pt]
ODinW~\cite{li2022elevater} (Object Detection in the Wild - Fig.~\ref{fig:cvitw}) is a composite benchmark, grouping 13 specialized detection sets (e.g., \textit{AerialDrone, Aquarium, Raccoon, Thermal} \cite{dwyer2024roboflow}) to evaluate performance across diverse and unusual domains and object categories. It has been utilized to assess zero/few-shot transfer capabilities, highlighting its relevance beyond standard closed-set evaluation, and laying at the edge of the old train/evaluate paradigm, a stepping stone to the large pretrain corpora paradigm.

\paragraph{Out-of-Distribution Detection Benchmarks}
Out-of-distribution (OOD) detection benchmarks are crucial for evaluating a model's ability to identify inputs that differ from its training data. In OOD detection, a dataset is chosen as in-distribution (ID), and others are designated for out-of-distribution testing. Popular ID datasets from the OpenOOD benchmark \cite{yang2022openood} include CIFAR-10 and CIFAR-100~\cite{krizhevsky2009learning}, each with standard splits and 10 or 100 classes, respectively. OOD sets are categorized as:

 Near-OOD - semantically similar but unseen classes, such as CIFAR-100 or TinyImageNet~\cite{le2015tiny}.
 
 Far-OOD - datasets with substantial content or domain differences from ID sets, like MNIST~\cite{lecun1998mnist}, SVHN~\cite{netzer2011reading}, Texture~\cite{cimpoi2014describing}, or Places365~\cite{zhou2017places}.

Frameworks like the OpenOOD library~\cite{yang2022openood} standardize these splits, evaluating detectors and classifiers using metrics like AUROC or FPR at 95\% TPR. Integrating OOD detection is vital for OWD solutions to flag unknown objects outside the training set.

\paragraph{Fine-Grained and Multimodal Benchmarks}
Multimodal datasets like Flickr30K~\cite{plummer2015flickr30k} and COCO Caption~\cite{chen2015microsoft} are employed for image-text retrieval and phrase grounding evaluation, particularly in VLM-based detection. These datasets assess the ability to detect novel objects mentioned in textual prompts through recall of localizable phrases.

\paragraph{Semantic Segmentation and Pixel-Level Evaluation}
Open-world segmentation evaluation utilizes datasets that extend detection to pixel-level annotations. Common sets include:

    PASCAL VOC 2012~\cite{pascal-voc-2012} for semantic segmentation (20 classes), Cityscapes~\cite{Cordts2016Cityscapes} (19 classes), focusing on urban driving scenes, and ADE20K~\cite{zhou2019semantic,zhou2017scene} with 150 object/stuff categories for wide scene parsing coverage.
    
    LVIS instance segmentation subset~\cite{gupta2019lvis}, featuring large-vocabulary segmentation with many rare categories.
OWD methods adapt these for discovering \emph{unknown} objects or handling partial annotation protocols, such as the Open World Evaluation Protocol \cite{joseph2021towards}. Emerging benchmarks like UVO~\cite{wang2021unidentified} explicitly label class-agnostic objects, including those from unknown categories.

\subsection{Large-Scale Corpora and Open-Ended Benchmarks}\label{subsec3_2}
The paradigm has shifted with the advent of Vision-Language Models (VLMs), moving towards pre-training on massive datasets and evaluating on open-ended benchmarks that assess broader multimodal understanding and generalization.

\subsubsection{Pre-training Corpora}\label{subsubsec3_2_1}
Vision-Language pre-training relies on large-scale image-text corpora to learn multimodal representations capable of zero-shot and open-world detection. These corpora provide the scale and diversity necessary for training robust VLMs.

\paragraph{Vision-Language Pre-training Corpora}

\noindent SBU Caption~\cite{ordonez2011im2text} and COCO Caption~\cite{chen2015microsoft} offered 1M and 1.5M image-text pairs, respectively, initially for captioning and retrieval tasks.

\par\noindent Visual Genome~\cite{krishna2017visual} (5.4M region descriptions) extends COCO with dense region descriptions, enhancing object-level concept learning.

\par\noindent Web-crawled datasets like Conceptual Captions~\cite{sharma2018conceptual} (3.3M pairs), Conceptual 12M~\cite{changpinyo2021cc12m}, and Red Caps~\cite{desai2021redcaps} (12M pairs) offer diverse categories and language usage.

\par\noindent LAION~\cite{schuhmann2021laion,schuhmann2022laion} (400M--5B pairs) stands out as one of the largest public image-text datasets, spanning over 100 languages, crucial for building open-domain visual encoders.

\par\noindent Proprietary datasets associated with models like CLIP~\cite{radford2021learning}, ALIGN~\cite{jia2021scaling}, and FILIP~\cite{yao2021filip} (with partially disclosed compositions) have produced powerful zero-shot detection backbones when combined with large language models.

\paragraph{Grounding Data for VLMs}
Models like Grounded Language-Image Pre-training (GLIP)~\cite{li2022grounded} utilize aggregated datasets for grounding language and vision:

    FourODs \cite{li2022grounded}: A combination of Objects365, OpenImages \cite{openimages}, Visual Genome, and ImageNetBoxes \cite{krizhevsky2012imagenet} (2.66M images total), for broad object detection pre-training.
    
    Cap4M/Cap24M: Large-scale image-text pairs (4M or 24M) with non-public, synthetic bounding-box annotations, generated via self-training \cite{li2022grounded}.
    
    GoldG (0.8M curated human-annotated grounding data) \cite{kebe2021a}: Aggregates Flickr30K \cite{young2014image}, Visual Genome Caption \cite{krishna2017visual}, and GQA \cite{hudson2019gqa} for high-quality phrase-region correspondences.

These datasets are fundamental for pre-training VLMs, enabling them to learn robust vision-language representations and enhancing downstream tasks like detailed object detection and localization.

\subsubsection{Open-Ended Evaluation Benchmarks}\label{subsubsec3_2_2}
The increasing complexity of open-world perception, reflected in fragmented dataset efforts, drives the community towards more unified benchmarks. These benchmarks aim to holistically evaluate models on detection, segmentation, and language understanding in open-ended scenarios.

The fragmentation of dataset efforts---COCO-based OWD splits, specialized OOD sets, large-scale VLM corpora, and cross-domain evaluations like ODinW---reflects the increasing complexity of open-world perception. In response, the community is steadily moving toward more unified benchmarks that pair bounding-box annotations, segmentation masks, and image-text metadata. Such a holistic design would allow a single model to \emph{detect} and \emph{segment} unknown objects while \emph{describing} or \emph{retrieving} them via language queries.

From our assessment of current trends, datasets for OWD must accommodate:
\begin{itemize}
    \item \emph{Large, diverse object taxonomies}, acknowledging that unknown categories appear outside the typical “closed” label set.
    \item \emph{Annotation depth}, combining bounding boxes, segmentation masks, textual grounding, and out-of-distribution labels.
    \item \emph{Incremental updates} to mimic real-world emergence of new object classes, supporting continual or lifelong learning protocols.
\end{itemize}

\paragraph{3D Open-World Detection Datasets}
While this review primarily focuses on 2D, many applications need 3D awareness. Existing 3D datasets like \emph{KITTI}~\cite{Geiger2012CVPR}, \emph{nuScenes}~\cite{caesar2020nuscenes}, \emph{SUN RGB-D}~\cite{song2015sun}, and \emph{ScanNet}~\cite{dai2017scannet} provide 3D object annotations, but formal open-world labeling protocols are underexplored. Future benchmarks could partition 3D categories into known and unknown sets and measure performance on novel categories.

\paragraph{Incremental and Real-Time Protocols}
Open-world detection's hallmark is incremental novel class discovery. While COCO-based splits~\cite{joseph2021towards} and ODinW~\cite{li2022elevater} partially address this, specialized protocols are needed for real-time scenarios. Approaches like BSDP~\cite{chen2024bsdp} and video-based unknown object detection~\cite{du2022unknown} show promise. Benchmarks mirroring real deployment, like time-sequenced frames from surveillance or robotics, are an open challenge to enable continuous, on-the-fly learning without forgetting.

\begin{table*}[htbp]
\centering
\footnotesize
\setlength{\tabcolsep}{3pt}  
\caption{Summary of the widely-used visual recognition datasets building up to open-ended perception.}
\label{tab:vlm_evaluation}
\resizebox{0.99\textwidth}{!}{%
\begin{tabular}{l|l|c|c|c|c|c}
\hline
\textbf{Task} & \textbf{Dataset} & \textbf{Year} & \textbf{Classes} & \textbf{Training} & \textbf{Testing} & \textbf{Evaluation Metric} \\
\hline
Image & MNIST \cite{lecun1998mnist} & 1998 & 10 & 60,000 & 10,000 & Accuracy \\
Classification & Caltech-101 \cite{li2022caltech} & 2004 & 102 & 3,060 & 6,085 & MeanPerClass \\
 & Oxford 102 Flowers \cite{Nilsback08} & 2008 & 102 & 2,040 & 6,149 & MeanPerClass \\
 & CIFAR-10 \cite{krizhevsky2009learning} & 2009 & 10 & 50,000 & 10,000 & Accuracy \\
 & CIFAR-100 \cite{krizhevsky2009learning} & 2009 & 100 & 50,000 & 10,000 & Accuracy \\
 & ImageNet-1k \cite{deng2009imagenet} & 2009 & 1,000 & 1,281,167 & 50,000 & Accuracy \\
 & SUN397 \cite{xiao2010sun,xiao2016sun} & 2010 & 397 & 19,850 & 19,850 & Accuracy \\
 & SVHN \cite{netzer2011reading} & 2011 & 10 & 73,257 & 26,032 & Accuracy \\
 & STL-10 \cite{coates2011analysis} & 2011 & 10 & 1,000 & 8,000 & Accuracy \\
 & GTSRB \cite{stallkamp2012man} & 2011 & 43 & 26,640 & 12,630 & Accuracy \\
 & IIIT5k \cite{MishraBMVC12} & 2012 & 36 & 2,000 & 3,000 & Accuracy \\
 & Oxford-IIIT PETS \cite{parkhi2012cats} & 2012 & 37 & 3,680 & 3,669 & MeanPerClass \\
 & Stanford Cars \cite{KrauseStarkDengFei-Fei_3DRR2013} & 2013 & 196 & 8,144 & 8,041 & Accuracy \\
 & FGVC Aircraft \cite{maji13fine-grained} & 2013 & 100 & 6,667 & 3,333 & MeanPerClass \\
 & Facial Emotion Recognition 2013 \cite{goodfellow2013challenges} & 2013 & 8 & 32,140 & 3,574 & Accuracy \\
 & Rendered SST2 \cite{radford2021learning,socher2013recursive} & 2013 & 2 & 7,792 & 1,821 & Accuracy \\
 & Describable Textures (DTD) \cite{cimpoi14describing} & 2014 & 47 & 3,760 & 1,880 & Accuracy \\
 & Food-101 \cite{bossard2014food} & 2014 & 102 & 75,750 & 25,250 & Accuracy \\
 & Birdsnap \cite{berg2014birdsnap} & 2014 & 500 & 42,283 & 2,149 & Accuracy \\
 & RESISC45 \cite{cheng2017remote} & 2017 & 45 & 3,150 & 25,200 & Accuracy \\
 & CLEVR Counts \cite{johnson2017clevr} & 2017 & 8 & 2,000 & 500 & Accuracy \\
 & PatchCamelyon \cite{veeling2018rotation,bejnordi2017diagnostic} & 2018 & 2 & 294,912 & 32,768 & Accuracy \\
 & EuroSAT \cite{helber2019eurosat} & 2019 & 10 & 10,000 & 5,000 & Accuracy \\
 & Hateful Memes \cite{kiela2020hateful} & 2020 & 2 & 8,500 & 500 & ROCAUC \\
 & Country211 \cite{radford2021learning,thomee2016yfcc100m} & 2021 & 211 & 43,200 & 21,100 & Accuracy \\
\hline
Object & PASCAL VOC 2007 Detection \cite{pascal-voc-2012} & 2007 & 20 & 5,011 & 4,952 & 11-point mAP \\
Detection & COCO 2014 Detection \cite{lin2014microsoft} & 2014 & 80 & 83,000 & 41,000 & box mAP \\
 & COCO 2017 Detection \cite{lin2014microsoft} & 2017 & 80 & 118,000 & 5,000 & box mAP \\
 & LVIS \cite{gupta2019lvis} & 2019 & 1203 & 118,000 & 5,000 & box mAP \\
 & ODinW \cite{li2022elevater} & 2022 & 314 & 132,413 & 20,070 & box mAP \\
 & Objects365 \cite{shao2019objects365} & 2019 & 365 & 593,000 & 20,000 & box mAP \\
 & OpenImages \cite{openimages} & 2018 & 601 & 1,700,000 & 125,000 & AP (IoU-based) \\
 & \textbf{OWOD (COCO-based Splits)} \cite{joseph2021towards} & 2021 & 80 + Unknown & 118,000 & 5,000 & mAP + U-Recall \\
\hline
Semantic & PASCAL VOC 2012 Segmentation \cite{pascal-voc-2012} & 2012 & 20 & 1,464 & 1,449 & mIoU \\
Segmentation & PASCAL Context \cite{mottaghi2014role} & 2014 & 459 & 4,998 & 5,105 & mIoU \\
 & Cityscapes \cite{Cordts2016Cityscapes} & 2016 & 19 & 2,975 & 500 & mIoU \\
 & ADE20k \cite{zhou2017scene} & 2017 & 150 & 25,574 & 2,000 & mIoU \\
 & SA-1B (Segment Anything) \cite{kirillov2023segment} & 2023 & \textit{class-agnostic} & 11M images / 1.1B masks & - & - \\
\hline
Image-Text & Flickr30k \cite{plummer2015flickr30k} & 2014 & - & 31,783 & - & Recall \\
Retrieval & COCO Caption \cite{chen2015microsoft} & 2015 & - & 82,783 & 5,000 & Recall \\
\hline
\textbf{Image-Text} & SBU Captions \cite{ordonez2011im2text} & 2011 & - & 1M & n/a & - \\
\textbf{Pretraining} & YFCC100M \cite{thomee2016yfcc100m} & 2014 & - & 100M & n/a & - \\
 & Visual Genome \cite{krishna2017visual} & 2016 & - & 108K images & n/a & - \\
 & Conceptual Captions (CC3M) \cite{sharma2018conceptual} & 2018 & - & 3.3M & n/a & - \\
 & Conceptual 12M (CC12M) \cite{changpinyo2021cc12m} & 2021 & - & 12M & n/a & - \\
 & WIT \cite{srinivasan2021wit} & 2021 & - & 11.5M & n/a & - \\
 & CLIP (OpenAI) \cite{radford2021learning} & 2021 & - & 400M & n/a & - \\
 & LAION-400M \cite{schuhmann2021laion} & 2021 & - & 400M & n/a & - \\
 & RedCaps \cite{desai2021redcaps} & 2021 & - & 12M & n/a & - \\
 & ALIGN (Google) \cite{jia2021scaling} & 2021 & - & 1.8B & n/a & - \\
 & ALT200M \cite{hu2022scaling} & 2022 & - & 200M & n/a & - \\
 & COYO-700M \cite{kakaobrain2022coyo-700m} & 2022 & - & 700M & n/a & - \\
 & LAION-5B \cite{schuhmann2022laion} & 2022 & - & 5.85B & n/a & - \\
\hline
\end{tabular}
} 
\end{table*}

\begin{figure}[h]
\centering
\includegraphics[width=\linewidth]{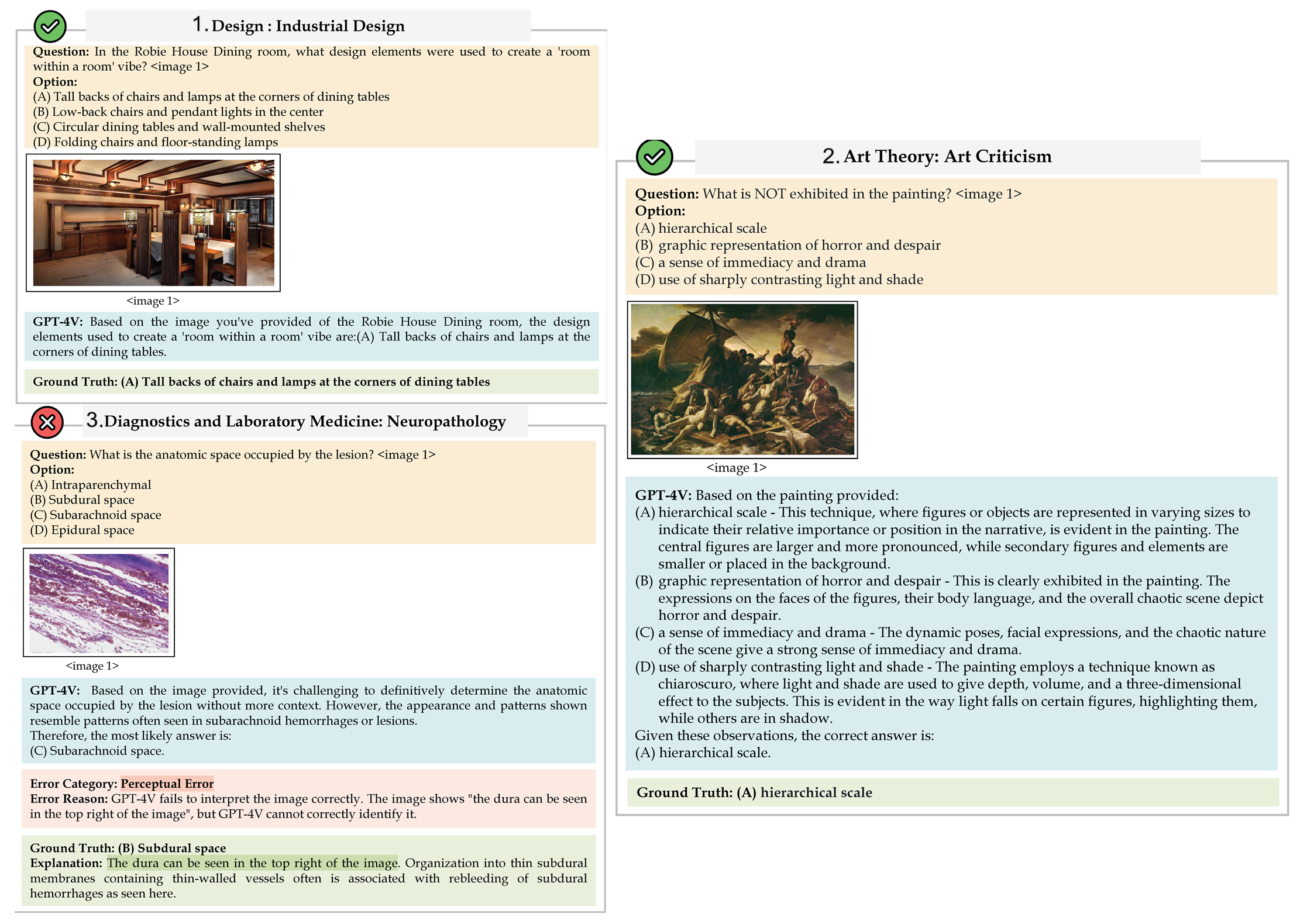}
\caption{Visual LLM Benchmark - Example Evaluation of GPT-4V from MMMU Bench. Answers for questions 1 and 2 considered sufficiently close to Ground Truth to be correct, and the one for question 3 erroneous owing to a perceptual error}
\label{fig:vllm_benchmarks}
\end{figure}

\paragraph{Advanced Multimodal Benchmarks}
Several new benchmarks focus on advanced multimodal reasoning and domain-specific knowledge, extending beyond classical tasks. RealWorldQA features 700+ high-resolution real-world images with QA requiring physical and spatial understanding \cite{xai_2025}. AI2D-RST uses 1,000 science diagrams for diagram interpretation in education \cite{hiippala2021ai2d}. MMBench, MMStar, and MMMU assess different dimensions of large models, from bilingual QA to expert knowledge \cite{liu2024mmbench, chen2024we, yue2023mmmu} - exemplified in Fig.~\ref{fig:vllm_benchmarks}. SEED-Bench emphasizes "generative comprehension" with 19k QA items across images and videos, testing temporal reasoning and cross-modal inference \cite{li2023seed}. HallusionBench and POPE specifically target hallucination detection and visual-text alignment in VLMs \cite{guan2024hallusionbench, li2023evaluating}.

\vspace{0.5em}
\noindent

\begin{figure}[htbp]
    \centering
    \includegraphics[width=\linewidth]{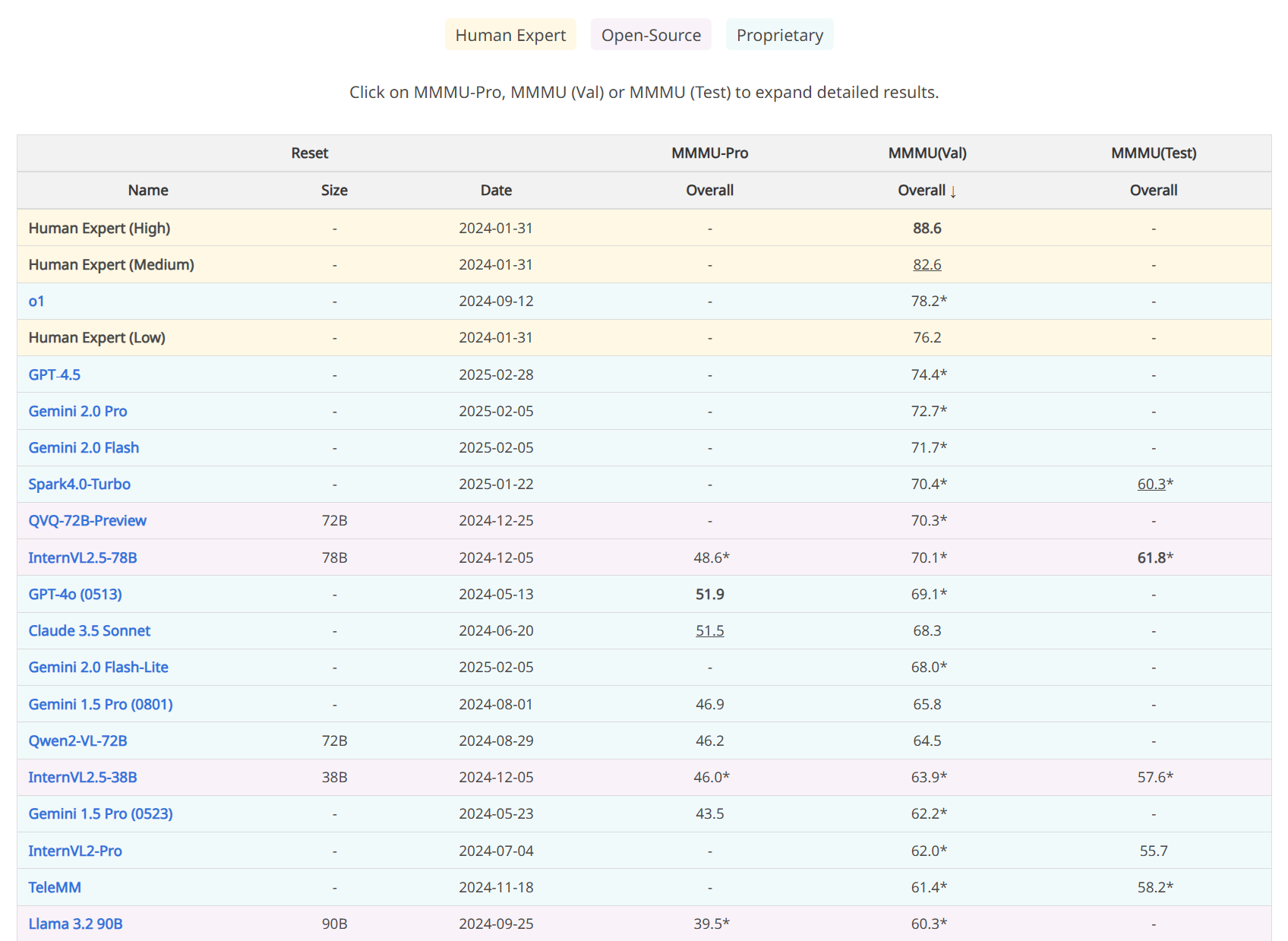}
    \caption{ MMMU top-20 results shows rapid benchmark saturation.}
    \label{fig:mmmu_top20}
\end{figure}

The rapid advancement in the field leads to benchmark saturation, one example on display in Fig.~\ref{fig:mmmu_top20}, with current systems approaching upper-bound performance on existing tasks. Continuous development of new datasets and expansions is necessary to probe more sophisticated capabilities for open-world vision-language understanding.

\subsection{Challenges in Open-World Dataset Creation}\label{subsec3_3}
\label{subsec:challenges_dataset_owd}

\noindent
Despite recent progress, building robust OWD datasets remains challenging due to several factors:

    Diversity vs.\ Annotation Load: Real-world diversity demands extensive object types and domains, but thorough annotation, especially for “unknown” items, is time-consuming.
    
    Defining Unknown Objects: Subjectivity in defining "unknown" objects among partially visible or occluded items complicates dataset creation.
    
    Domain Shifts: OWD datasets should account for domain shifts. Including domain-shift splits is crucial for evaluating model generalization.
    
    Evaluation Metrics: Beyond standard metrics, evaluating unknowns requires metrics like \emph{U-Recall}, \emph{AUROC}, or \emph{FPR}, making fair comparisons challenging.
    
    Annotating “Unknown” and Dataset Complexity: Lack of consensus in labeling unknowns complicates comparisons. Future datasets could use semi-automatic proposals for refinement.
    
    Toward Unified Evaluations: While holistic benchmarks are needed, current evaluations often separate tasks. Unified benchmarks for understanding, detecting, segmenting a scene are necessary for comprehensive open-world capability assessment.

\vspace{5pt}
\noindent
Improving OWD datasets and evaluation protocols requires community effort and semi-automated labeling approaches. Standardized benchmarks that capture real-world complexities are crucial for advancing open-world recognition.


\noindent Note on Omitted Datasets. Not all datasets mentioned in our survey appear in Tables~\ref{tab:vlm_evaluation} and~\ref{tab:advanced_vlm_benchmarks}. This deliberate choice reflects our focus on the convergence toward open-ended perception rather than comprehensive coverage of all precursor domains. While we discuss numerous specialized datasets throughout the text including them in our tables would expand this survey beyond its intended scope. We have instead prioritized datasets most relevant to the emergence of unified, open world detection capabilities. Readers interested in datasets for specific subdomains are encouraged to consult the corresponding sections where these resources are discussed in context.

\begin{table*}[htbp]
\centering
\footnotesize
\setlength{\tabcolsep}{4pt}  
\caption{Selection of newly introduced benchmarks for advanced evaluation of large Vision-Language Models. 
Each resource tests different aspects of open-ended reasoning and detection in multimodal contexts.}
\label{tab:advanced_vlm_benchmarks}
\resizebox{0.95\linewidth}{!}{
\begin{tabular}{l|c|c|p{4.7cm}|p{6.2cm}}
\hline
\textbf{Dataset} & \textbf{Year} & \textbf{\#Samples} & \textbf{Primary Focus} & \textbf{Key Features / Challenges} \\
\hline
\textbf{RealWorldQA \cite{xai_2025}} 
 & 2023 & 700+ images, each with QA & 
 Physical/spatial understanding from real-world scenes 
 & 
 - Multiple-choice QA (2--4 options)\newline
 - High-res images (1080p)\newline
 - Commonsense reasoning needed 
 \\
\hline
\textbf{AI2D-RST \cite{hiippala2021ai2d}} 
 & 2022 & 1{,}000 diagrams & 
 Diagram interpretation in primary-school science 
 & 
 - Multi-layer annotation (grouping, connections, RST discourse)\newline
 - Evaluates diagram-based reasoning, crucial for open-world detection in educational or technical diagrams 
 \\
\hline
\textbf{MMBench \cite{liu2024mmbench}}
 & 2023 &  3{,}217 multiple choice questions & 
 Bilingual (EN/ZH) multi-capability benchmark 
 & 
 - Large-scale curated questions\newline
 - CircularEval for standardized scoring\newline
 - Tests from fine-grained object recognition to multi-hop reasoning 
 \\
\hline
\textbf{MMStar \cite{chen2024we}}
 & 2024 & 1{,}500 curated\newline(vision-indispensable) samples & 
 Advanced multimodal QA with minimal data leakage 
 & 
 - Samples \emph{require} visual input (cannot be solved by text alone)\newline
 - Highlights \emph{real} multi-modal gains 
 \\
\hline
\textbf{MMMU \cite{yue2023mmmu}}
 & 2023 & 11.5k questions\newline across 6 disciplines & 
 Expert-level knowledge \& domain-specific reasoning 
 & 
 - College-level tasks in science, medicine, engineering, etc.\newline
 - Charts, tables, diagrams requiring advanced domain knowledge 
 \\
\hline
\textbf{SEED-Bench \cite{li2023seed}}
 & 2024 & 19k multiple-choice\newline QA items & 
 Generative comprehension (image + video) 
 & 
 - Covers 12 dimensions of vision understanding\newline
 - Purely objective scoring (no free-form)\newline
 - Tests deeper inference, temporal/causal reasoning 
 \\
\hline
\textbf{HallusionBench \cite{guan2024hallusionbench}}
 & 2023 & 1.13k samples & 
 Detecting and measuring hallucinations in visual QA 
 & 
 - Yes/No queries about object presence or attributes\newline
 - Evaluates fidelity of model grounding vs. “invented” details 
 \\
\hline
\textbf{POPE \cite{li2023evaluating}}
 & 2023 & 9k samples & 
 Probing visual-text alignment and hallucination 
 & 
 - Binary forced-choice tests for factual consistency\newline
 - Focus on ambiguous or challenging images to expose model overclaims 
 \\
\hline
\end{tabular}
} 
\end{table*}

\section{Convergence}\label{sec4}
In section 2 we set out to map the advances of key subdomains contributing to OWD over the years. In this section we aim at reflecting on their interdependence, organize them in major eras, and present the state of the art.

\subsection{A Foreword on Computer Vision}\label{subsec4_1}
\label{subsec:foreword_visual_detection}

In what some regard now as being the starting point of computer vision, Marvin Minsky in 1966 set out to solve artificial visual perception from images in one summer vision project at MIT. It became quickly evident that not only was the task colossal and insolvable for the time, but that even seemingly simple sub-tasks---like distinguishing a chair and a table from an image---were highly complex problems that would not be achieved until decades later. In trying to solve \emph{Vision}, a first prerequisite is having a clear definition of the problem, and of measurable goals aimed to be attained. While the project never reached its lofty ambitions, it laid the groundwork for the field of Computer Vision. Researchers began to understand the enormity of the problem and started defining well-bounded subdomains in which progress could be made and measured.

Over the ensuing decades, these smaller subdomains blossomed into specialized disciplines: edge detection gave us maps of boundaries, saliency detection spotlighted regions of elevated interest, and object detection drew bounding boxes around discrete entities. Each domain reflected a piece of the visual puzzle, helping us chip away at the far larger question: \emph{when do we say we have recognized something}? That question itself is fraught with ambiguities. For instance, do we regard a person’s hand as an independent entity, or as an inseparable extension of the body? Are five fingers five different units, or five parts of a single “object”? Even the notion of \emph{background} proves slippery. The same cluster of pixels could be “irrelevant clutter” in one context but critical to highlight in another, depending on a user’s task-oriented perspective.

\paragraph{A More Palpable Illustration.}
The specific technical goals of a vision system can dramatically reshape \emph{how} we define an object. Consider three different, but interrelated, detection subdomains:
\begin{itemize}
    \item Binary segmentation. A lab studying cell cultures may use a pixel-wise function $M(x,y)\in\{0,1\}$ to separate “cell” (foreground) from “non-cell” (background). Here, the shape of each cell at the pixel-level is paramount, and each pixel’s label matters.
    \item Object bounding-box detection. Meanwhile, a robotics engineer may prefer bounding boxes $(x_1,y_1,x_2,y_2)$ to quickly locate each cell for automated pipetting, caring less about the precise boundary and more about the global position of each cell in the field.
    \item Landmark or keypoint detection. In some specialized medical-imaging scenarios, the focus might narrow down to a few key coordinates (e.g., lesion centers or joint points), extracting only the minimal geometry needed for diagnosis or manipulation.
\end{itemize}
Although all three of these subdomains ostensibly aim to “find the object,” the internal representation, output format, and error metrics can vary drastically. A method that excels in pixel-level segmentation might be wholly unsuited for bounding-box detection, and vice versa, even though the underlying dataset could be the same set of microscopic images. Such is the interplay between \emph{what we want to see} and \emph{how we choose to see it}.

In our present era, fresh frontiers in visual detection have begun to converge back upon an essential inquiry: \emph{what is an object}? A surgical robot may deem individual fingers as separate “grasps” needing independent control, whereas a self-driving car’s pedestrian detection model lumps them together in a single silhouette labeled “person.” So, too, does the notion of “background” grow fluid: what is background for one task can become foreground for another. This malleability reflects the deeply subjective, goal-oriented nature of perception. This inherent subjectivity in defining objects and background reveals a core challenge for traditional computer vision systems, which often rely on task-specific models tailored to narrow problem definitions. However, recent advancements in foundational models offer a promising shift away from this task-centric paradigm, demonstrating a capacity for broader generalization.

The work by \cite{simeoni2023unsupervised} compellingly illustrates this point. Their study shows how self-supervised Vision Transformers (ViTs), trained for general visual representation, can be effectively applied to classical vision tasks previously addressed with specialized techniques.  For example, in unsupervised saliency detection, methods leveraging self-supervised ViTs achieve significantly improved performance. On the DUTS-TE dataset, IoU scores dramatically increased from 51.1 (with E-BigBiGAN \cite{voynov2021object}, a leading pre-ViT method) to 74.9 (with SEMPART-fine \cite{ravindran2023sempart}, using a self-supervised ViT without task-specific post-processing).  Similarly, for single-object discovery on VOC07, self-supervised ViTs like TokenCut \cite{wang2022self} achieve a CorLoc of 68.8 without dedicated learning, outperforming earlier specialized methods like rOSD \cite{vo2020toward} (54.5 CorLoc). These examples highlight that these newer, broadly trained vision models possess an inherent ability to perform well across a range of classical vision tasks, simply by being appropriately ‘aimed’ at them. The same underlying model can effectively tackle boundary separation, saliency detection, or object localization, showcasing a significant step towards more versatile vision systems.

It is within this mosaic of viewpoints and the emergent capabilities of foundational models that the concept of OWD emerges. As we push toward broader and more adaptive detection paradigms, we draw ever closer to the original ambition of that “summer project” from decades ago: to endow machines with the flexible power to see---and to \emph{understand}---the infinite complexity of the visual world.

\subsection{From Single-Task Pipelines to Large Multimodal Foundations}\label{subsec4_2}
\label{subsec:eras_foundations}

Computer Vision has undergone several paradigm shifts in how it approaches detection and recognition tasks over the past decades. In this subsection, we outline four broad eras of vision methodology: (i)~pre-deep-learning (classical methods), (ii)~deep-learning-based, (iii)~large foundational models, and (iv)~large multimodal models. We highlight key architectures, learning objectives, and their convergences, culminating in today’s push toward massive vision-language models that promise generalized, open-ended perception capabilities.

\subsubsection{Era 1: Pre-Deep-Learning Classical Methods}\label{subsubsec4_2_1}
As introduced in Section 2, early vision tasks such as object detection, edge/saliency detection, and background subtraction were tackled using hand-crafted features and statistical models. Techniques like Scale-Invariant Feature Transform (SIFT) \cite{lowe2004distinctive}, Histogram of Oriented Gradients (HOG) \cite{dalal2005histograms}, or robust background modeling \cite{koller1994towards, stauffer1999adaptive} enabled systems to detect objects in constrained settings. However, these “classical” pipelines were usually limited to specific scenarios:

    Feature Engineering: Manually designed descriptors (e.g., HOG) leveraged gradients and local histograms.
    
    Shallow Classifiers: Models like SVMs \cite{hearst1998support} or Adaboost \cite{freund1999short} operated on these features to classify or detect objects (e.g., the Viola-Jones face detector \cite{viola2001rapid}).

Despite being foundational, these methods lacked scalability and often failed under domain shifts or more complex, “in-the-wild” scenes. Their per-task design also made simultaneous handling of multiple vision challenges cumbersome. This era, however, laid important groundwork by formalizing problem definitions and emphasizing performance metrics (precision, recall, F1, etc.).

\subsubsection{Era 2: Deep Learning and Unified Architectures}\label{subsubsec4_2_2}
The 2010s saw an explosion of Convolutional Neural Network (CNN) architectures that superseded classical pipelines on tasks like image classification, object detection, saliency, and segmentation. Over time, the field converged on a number of design patterns:
\begin{itemize}
    \item CNN Backbones: Networks like AlexNet \cite{krizhevsky2012imagenet}, VGG \cite{simonyan2014very}, ResNet \cite{he2016deep} learned end-to-end feature hierarchies directly from data.
    \item Detection Pipelines: Two-stage detectors (Faster R-CNN \cite{ren2016faster}) generated region proposals before classification, whereas single-stage detectors (YOLO \cite{redmon2016you}, SSD \cite{liu2016ssd}) directly predicted bounding boxes and class confidences in a fully convolutional pass.
    \item Encoder-Decoder Designs: Semantic segmentation or saliency detection models (e.g., U-Net \cite{ronneberger2015u}, FCN \cite{long2015fully}) took advantage of symmetric encoder-decoder architectures for pixel-level predictions.
\end{itemize}
These deep methods started to unify once-disparate tasks by sharing backbones and optimization regimes. For instance, the YOLO framework replaced hand-designed proposal mechanisms with learned anchor priors, while semantic segmentation networks learned feature maps that doubled as saliency predictors in parallel. Yet, a significant caveat remained: most models were \textit{task-specific} and \textit{closed-set}, i.e., they did not generalize well to objects or concepts not seen in training.

It's in the later part of this era, early 2020s, that Open World Object Detection (OWOD) first started to emerge as an independent task. The proposal simple in its ideation - endow object detectors with the capacity to recognize when an object is foreign to them, and then, incrementally learn all the new foreign objects - proved challenging in its formalization. \cite{joseph2021towards} proposed ORE as an evaluation framework and an energy based, unknown-aware solution to serve as a sturdy baseline for the framework. This initial baseline already made use of contrastive learning principles - a method we deem as sitting at the foundation of the 3rd era that furthered open ended detection - by using contrastive clustering in the energy based model.

\subsubsection{Era 3: Large Foundational Models via Contrastive and Self-Distillation Learning}\label{subsubsec4_2_3}
\label{subsec:CLIP_and_friends}
A breakthrough toward more generalizable vision features emerged with large-scale \textit{foundational} models such as CLIP \cite{radford2021learning}, DINO \cite{caron2021emerging, oquab2023dinov2}. These models departed from the single-task approach, instead learning universal representations from massive unlabeled or weakly labeled datasets. Three important methodological paths include \emph{contrastive} learning, \emph{self-distillation} and \emph{masked-autoencoding}:
    
    Contrastive Learning (CLIP, ALIGN):
    Leveraging large-scale image-text pairs, CLIP \cite{radford2021learning} aligns image embeddings $f_\theta(\cdot)$ and text embeddings $g_\phi(\cdot)$ in a shared semantic space. A common objective is the InfoNCE or contrastive loss:
    \begin{equation}
    \label{eq:contrastive_loss}
    \mathcal{L}_{\text{CL}} = -\sum_{(x,y) \in \mathcal{D}} \log
        \frac{\exp\big(\langle f_\theta(x), g_\phi(y)\rangle/\tau\big)}
             {\sum\limits_{y' \in \Omega} \exp\big(\langle f_\theta(x), g_\phi(y')\rangle/\tau\big)},
    \end{equation}
    where $(x,y)$ is an image-text pair, $\Omega$ is a batch or corpus of negative text samples, and $\tau$ is a temperature parameter. By “pulling” correct image-text pairs together and “pushing” mismatched ones apart, models learn \textit{semantic consistency} across modalities, enabling zero-shot classification, open-vocabulary detection, image similarity, and beyond.

\begin{figure}[h]
    \centering
    \includegraphics[width=\linewidth]{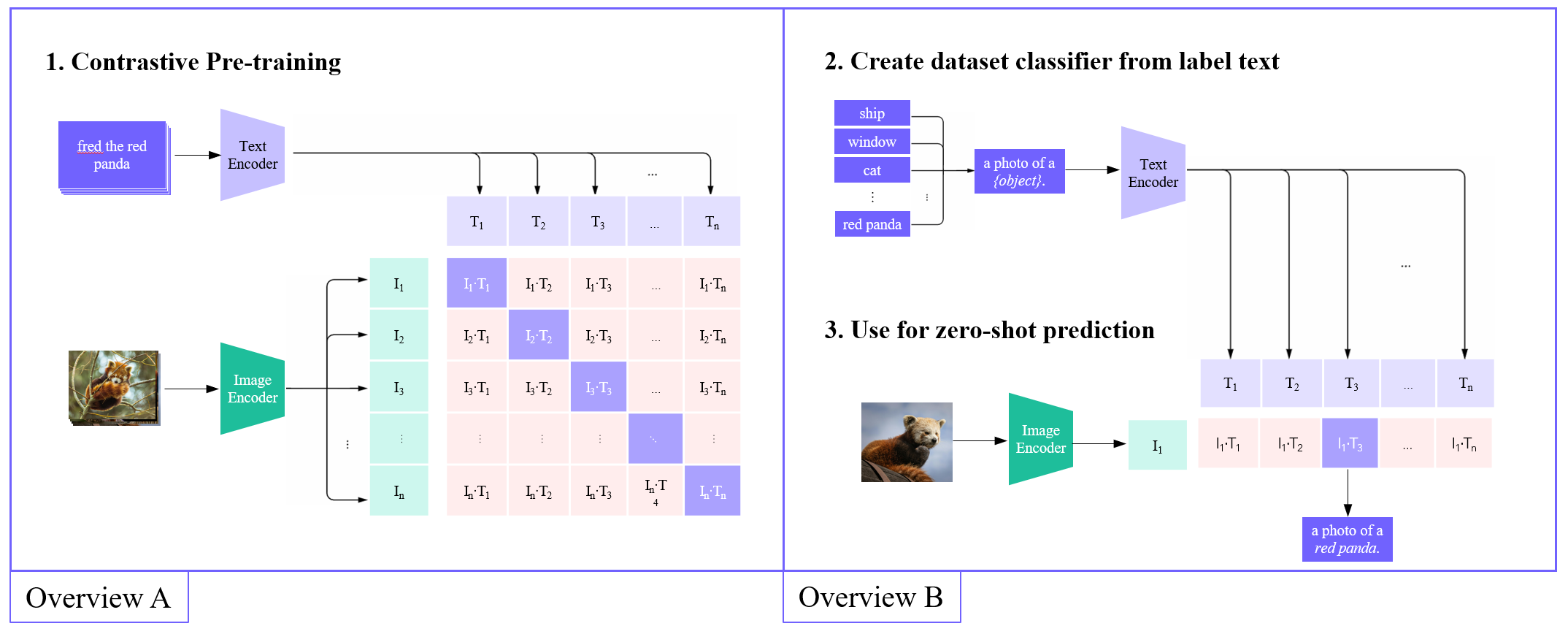}
    \caption{CLIP - A: Pretraining on image-caption sets, B: Create zero-shot classifier using masked text prompting}
    \label{fig:clip}
\end{figure}

    Self-Distillation (DINO, DINOv2):
    Self-distillation \cite{caron2021emerging} trains a \emph{student} network to match the outputs of a \emph{teacher} network, where both networks may share the same architecture (e.g., a Vision Transformer). This process drives feature learning because, without explicit labels, the student must internally discover visual patterns and structures in the unlabeled data to effectively match the teacher. The teacher’s parameters are often updated via an exponential moving average (EMA) of the student’s parameters. A typical loss can be written as:
    \begin{equation}
    \label{eq:sd_loss}
    \mathcal{L}_{\text{SD}} = \sum_{x \in \mathcal{D}} \Big\| f_{\theta}(x) - f_{\theta_{\text{EMA}}}(x)\Big\|^2,
    \end{equation}
    where $f_{\theta}(x)$ is the student output, and $f_{\theta_{\text{EMA}}}(x)$ is the fixed teacher output (updated periodically or via EMA). DINOv2 \cite{oquab2023dinov2} scales this concept to enormous uncurated image corpora, resulting in highly generalizable features that excel in tasks like segmentation, detection, and retrieval—often without further labels.

    Masked Autoencoding (MAE):
    Another key self-supervised approach is \emph{masked autoencoding}, pioneered by BERT \cite{kenton2019bert} in NLP and adapted to vision in MAE \cite{he2022masked}. A random subset of image patches is masked, and the network learns to reconstruct them from the remaining visible patches:
    \begin{equation}
    \label{eq:mae_loss}
    \mathcal{L}_{\text{MAE}} = \sum_{x \in \mathcal{D}} \Big\| M \odot (\widehat{x} - x) \Big\|^2,
    \end{equation}
    where $M$ is a binary mask selecting the missing patches, $x$ is the original image, and $\widehat{x}$ is the reconstructed image. By learning from partially obscured inputs, the model acquires robust \emph{holistic} scene understanding, improving its transfer to diverse downstream tasks.

What unites these approaches is their ability to learn from \emph{unbounded} data: image-text pairs scraped from the web (as in CLIP) or large unlabeled image corpora (as in DINO/MAE). These foundational encoders thus move us closer to open-world detection by capturing diverse, highly transferable representations of “visual reality.” As a practical proof, CLIP-based detectors have been adopted for image search and indexing at scale (e.g., unsplash.com uses captions generated by CLIP), while DINO’s self-distilled features are used for unsupervised saliency and region discovery \cite{simeoni2023unsupervised}.

\noindent
How CLIP works:
In “Overview A” of Fig.~\ref{fig:clip} we see CLIP jointly process large collections of images and their accompanying text. An \emph{image encoder} transforms each input image into a corresponding embedding (denoted $I_1, I_2,\dots,I_N$), while a \emph{text encoder} converts text (e.g., captions, class names) into text embeddings ($T_1, T_2,\dots,T_N$). During contrastive pre-training, CLIP “pulls together” matched image--text pairs by maximizing their embedding dot product (e.g., $I_i \cdot T_i$) and “pushes apart” mismatched pairs by minimizing their similarity (e.g., $I_i \cdot T_j$ for $i \neq j$) - see Contrastive Loss [Equation \ref{eq:contrastive_loss}]. In “Overview B,” the trained text encoder is then used to create a set of \emph{class embeddings} from plain-language labels (e.g., “\texttt{a photo of a plane}, “\texttt{a photo of a dog}”), while the image encoder produces embeddings for new input images. Zero-shot prediction follows naturally: the model compares the image embedding $I_1$ against each candidate text embedding ($T_1, T_2,\dots,T_N$) and picks the label whose text embedding is most similar. This procedure enables CLIP to recognize novel object categories (“dog,” “bird,” etc.) without explicit supervised training on those classes, relying solely on its learned alignment between visual and textual concepts.

\subsubsection{Era 4: Large Multimodal Models and Convergence of Vision \& Language}\label{subsubsec4_2_4}
\begin{figure}[htbp]
    \centering
    \includegraphics[width=\linewidth]{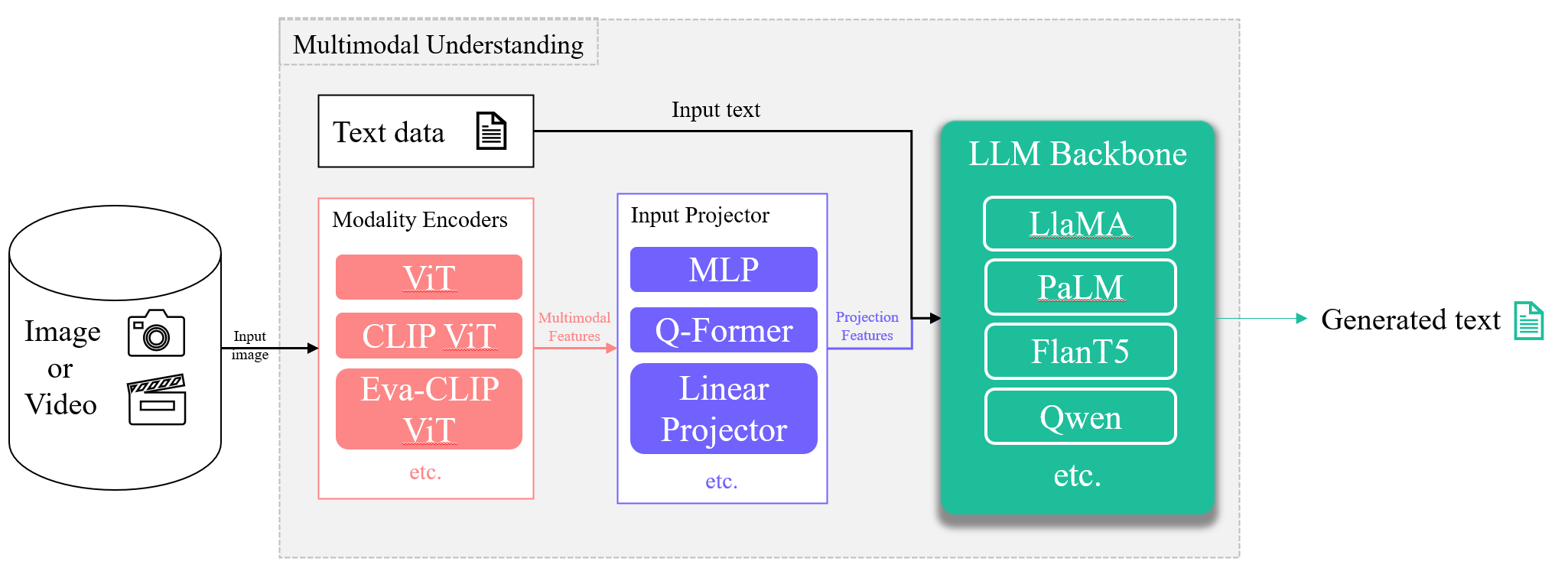}
    \caption{General Large Multimodal Architecture~\cite{zhang2024mm}}
    \label{fig:LMM}
\end{figure}

Having robust vision backbones (CLIP, DINO) naturally sets the stage for \emph{multimodal} foundation models that integrate visual encoders with large language models (LLMs). This final step aims for a single system able to answer open-ended queries, localize arbitrary objects, caption images, or execute complex reasoning about scenes. Fig.~\ref{fig:LMM} displays the general architecture of existing multimodal models, with input and output projectors being the bridge between  multimodality and language.

    Vision-Language Fusion: Models such as BLIP \cite{li2022blip} and Flamingo \cite{alayrac2022flamingo} project image embeddings into partially shared spaces with text embeddings. Concretely, a function $h_\alpha(\cdot)$ may map image features to a set of “vision tokens” $\mathbf{v} = h_\alpha(f_{\theta}(x))$, which are then fed into a language model together with text tokens $\mathbf{t}$. By merging these token sequences, the model can handle tasks from image captioning to object segmentation via textual prompts. High-level “vision tokens” typically come from a pretrained multimodal encoder (e.g., CLIP, EVA-CLIP \cite{sun2023eva}) that preserves broad generalization capabilities.
    
    Instruction Tuning and Dialogue: Systems such as InstructBLIP \cite{dai2023instructblip} or LLaVA \cite{liu2024visual} incorporate instruction tuning, chain-of-thought prompting, and multi-turn dialogue to produce “chat-based” multimodal interactions. Users can sequentially refine queries (e.g., “Identify all peculiar items in the frame,” “Now zoom in on that red object,” etc.)---a capability well-aligned with OWD. Recent research also reveals the growing importance of \emph{visual prompting} to guide a model’s attention, with approaches like Q-Former \cite{zhang2024vision}, FERRET \cite{you2023ferret}, or GPT4RoI \cite{zhang2023gpt4roi} using region-based prompting to emphasize specific parts of an image.
    
    Unified Perception via Visual Tokens: A major challenge has emerged in how to incorporate large numbers of “visual tokens” (particularly for high-resolution images) into a language model’s limited context window. Techniques like \emph{Scaling on Scales (S2)} \cite{shi2024we} attempt to manage this by extracting visual tokens at multiple resolutions, while adaptive token reduction (e.g., PruMerge \cite{shang2024llava}) reduces token redundancy through sampling or clustering. Other lines of work propose nesting visual tokens in hierarchical “Matryoshka” representations \cite{kusupati2022matryoshka,cai2024matryoshka} or stacking them deeply to preserve detail without exploding context length \cite{meng2024deepstack}.
    
\begin{figure}[htbp]
    \centering
    \includegraphics[width=\linewidth]{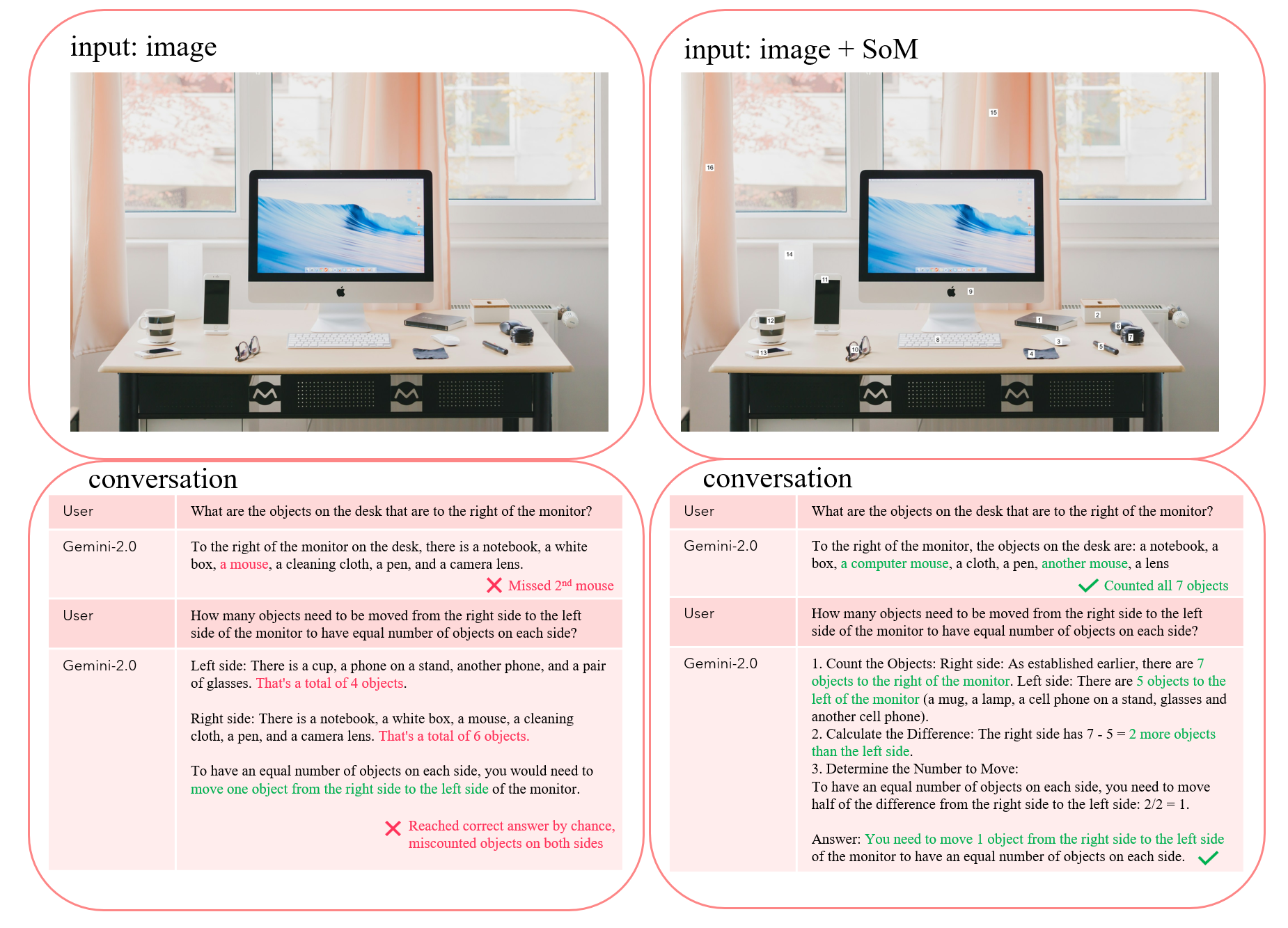}
    \caption{Marks guidance for VLLMs}
    \label{fig:SetOfMarks2}
\end{figure}

    Pixel-Based Prompting: In parallel, several recent methods show that \emph{directly overlaying} bounding boxes, scribbles, or textual annotations onto the input image can be an even more intuitive way to prompt a multimodal model—often called \textit{visual pixel prompting} \cite{wu2022unleashing,yu2024attention}. Systems that employ a set-of-mark method \cite{yang2023set} like ViP-LLaVA \cite{cai2024vip} or SoM-LLaVA \cite{yan2024list} demonstrate significant improvements in region understanding and visual grounding tasks by drawing a user’s prompts directly on the pixel space of the image (e.g., a scribbled arrow, a circled region, a segmentation mask) - in Fig.\ref{fig:SetOfMarks2} we can see how marks in an image contribute to better scene understanding and location grounding.


In addition to these architectural and methodological trends, the community has begun exploring specialized encoders for different visual cues. For instance, VCoder \cite{jain2024vcoder} leverages multiple coders (e.g., depth, segmentation) to deliver more targeted vision inputs to the language model, substantially boosting performance on perception tasks. As these large multimodal models grow, so does their capacity for \emph{open world} reasoning, including tasks like unknown-object detection, spatial reasoning, counting, and more \cite{chen2024spatialvlm}. However, fundamental challenges remain; for example, precise spatial understanding and accurate counting of complex scenes still prove difficult \cite{chandhok2024response}. Furthermore, extending these systems to \emph{video understanding} \cite{madan2024foundation,qian2024streaming} opens up new avenues (and complexities) around temporal and causal reasoning, with potential applications in robotics, user-interface navigation, and beyond \cite{kamboj2024brief}. Ultimately, by fusing powerful language backbones with dynamic vision encoders, \emph{large multimodal models} carry forward the deep-learning revolution in perception, steering us ever closer to flexible open-world detection and truly holistic scene understanding.

\vspace{0.5em}
\noindent
Implications for Open-World Detection.
In each new era, we observe a movement away from specialized, domain-specific heuristics towards \emph{generalizable}, \emph{data-driven} embeddings. Whereas classical pipelines rely on curated features, these large foundational and multimodal models reflect the ongoing convergence in Computer Vision: from pre-specified to open-ended tasks, from single-modal to language-infused representations, and ultimately toward a future where “detecting an unknown object” or “segmenting any region of interest” emerges naturally from an all-encompassing perception engine. By unifying vision and language, we come closer to fulfilling the longstanding dream of a truly open-world perception system.

\section{Future Directions}\label{sec5}
While our survey has aimed to provide a comprehensive picture of the OWD landscape, we acknowledge that multiple subdomains and related topics lie outside the current scope but are nevertheless intertwined with the same fundamental challenges. Our decision to exclude these topics here was motivated by the vastness of the collective field, but we recognize their importance and expect future work to draw many more connections across these complementary threads.

Looking ahead, we envision four primary, interwoven avenues of progress that will be pivotal for pushing OWD toward more powerful, generalizable, and practical outcomes:

\subsection{Advancing Multimodality}\label{subsec5_1}
The remarkable performance gains witnessed in the past few years reflect a growing capacity to process and fuse multiple data modalities. Modern large language models (LLMs) already show early success in pairing image and textual data, and we anticipate much broader integration of video, audio, 3D, and other signals in the near future. While current Vision-Language Models (VLMs) and Large Multimodal Models (LMMs) have expanded the scope of what is “detectable,” there is still considerable work to be done in achieving the fine-grained accuracy required for precise object localization, segmentation, and geometric understanding.

\subsection{Transitioning from 2D to 3D-Integrated Open Detection}\label{subsec5_2}
As industries and robotics applications (e.g., self-driving cars, drone navigation, and smart manufacturing) continue to deploy multi-sensor systems, 3D data captured through LiDAR, depth cameras, or stereo rigs becomes increasingly accessible. We expect imminent advances in open world 3D detection, where foundational models, presently geared mostly toward 2D, will incorporate depth and geometric cues seamlessly. Potential lines of research include:
\begin{itemize}
    \item 3D Foundational Backbones - Emerging 3D versions of the large pre-trained vision models, leveraging sensor-fusion approaches (RGB+depth+LiDAR).
    \item Progressive Self-Supervision - Expanding self-distillation and contrastive objectives to volumetric or point-cloud data, potentially adopting multi-view consistency or 3D-based reconstructions to handle unknown shapes.
    \item Holistic Perception Engines - Systems that unify 2D bounding boxes, 3D object proposals, and textual grounding in a single pipeline, ultimately boosting the capacity to detect unknown objects anywhere in physical space.
\end{itemize}
Over the next few years, we anticipate the most advanced detection models will no longer be purely 2D-based, instead offering full 3D situational awareness and cross-domain adaptability.

\subsection{Scaling Up, Then Distilling Down}\label{subsec5_3}
We also project continued emphasis on scaling: more extensive datasets, larger models, and increased training durations. This scaling often unlocks emergent capabilities, whether in zero-shot transfer, multi-domain generalization, or fine-grained detection of previously unseen items. However, alongside this growth, a parallel stream of research is likely to concentrate on \emph{model distillation}, producing lighter-weight networks that inherit the strengths of their gargantuan ancestors while running efficiently on consumer-grade or embedded hardware.

This dual dynamic of “scale up, then compress down” aligns with what the broader machine learning community has already observed in language modeling. Trends suggest highly distilled open world detectors capable of reaching real-time inference rates above 30~FPS, even on devices with constrained computational budgets. In the near term, applications such as robotics, drones, and autonomous vehicles will catalyze the push for these efficient-yet-powerful OWD models that operate reliably under latency and energy constraints.

\subsection{Robotics Integration}\label{subsec5_4}
As OWD matures, its natural expansion into robotics will accelerate. Systems that parse unknown objects, or reason about dynamic, unstructured environments, will be crucial to safe human-robot collaboration in warehouses, homes, and medical settings. Likewise, innovations in real-time open-ended detection will help drones navigate cluttered airspaces or assist automotive systems in analyzing rare road obstacles. These domains will, in turn, feed back valuable data and pressure for robust, efficient solutions. We expect:

On-Device Intelligence: Greater synergy between on-edge neural accelerators and model architectures that can learn or adapt on the fly. This is something we expect incremental learning to bridge into as methods advance.

Interactive Scene Understanding: Robotic systems that engage in dialogue with humans (via LLM-based interfaces), refining detection targets in real time.

Active Object Discovery: Robots that actively move, probe, or re-position sensors to reduce uncertainty about unknown or partially visible objects.

\medskip
\noindent
Outlook. With the continuing convergence of historically “isolated” tasks (tracking, segmentation, detection, depth, saliency, etc.) into unified frameworks, we see a future where detection is less about bounding boxes or label sets, and more about truly understanding all that is visible. Ultimately, the integration of larger, more holistic models with domain-specific expertise and hardware-specific optimizations will move us closer to a reality in which machines observe, reason about, and adapt to their environments with unprecedented fluidity — progressing, at last, toward a level of visual intelligence once seen as purely speculative. Or, as history would have it, a successful summer project.

\section{Conclusion}\label{sec6}
In this survey, we examined OWD from multiple angles, highlighting foundational subdomains such as saliency detection, foreground/background separation, zero-shot detection, and out-of-distribution recognition. We discussed how each of these technical components contributes to building class-agnostic, generalizable systems that better reflect real-world variability. Furthermore, the shift toward large multimodal foundational models—particularly those bridging vision and language—has demonstrated how high-level semantic understanding and flexible representational capacity jointly enable more effective detection of both known and unforeseen objects. Although these emerging frameworks offer promising directions, many operational and theoretical challenges remain: from improving spatial localization and 3D integration, to managing scaling demands in a manner that also permits efficient distillation for resource-limited environments. We conclude that OWD is likely to become increasingly relevant in applications that require robustness to novelty and dynamic updates, making it essential for the field to continue exploring both the foundational mechanisms of visual perception and the more holistic, multimodal paradigms that are rapidly redefining the scope of computer vision.

\backmatter


\bmhead{Statements and Declarations}

\subsection*{Data Availability}
Data sharing is not applicable to this article as it is a review based on existing literature. No new datasets were generated or analyzed.

\bibliography{owd_bib}


\begin{thebibliography}{215}
\ifx \bisbn   \undefined \def \bisbn  #1{ISBN #1}\fi
\ifx \binits  \undefined \def \binits#1{#1}\fi
\ifx \bauthor  \undefined \def \bauthor#1{#1}\fi
\ifx \batitle  \undefined \def \batitle#1{#1}\fi
\ifx \bjtitle  \undefined \def \bjtitle#1{#1}\fi
\ifx \bvolume  \undefined \def \bvolume#1{\textbf{#1}}\fi
\ifx \byear  \undefined \def \byear#1{#1}\fi
\ifx \bissue  \undefined \def \bissue#1{#1}\fi
\ifx \bfpage  \undefined \def \bfpage#1{#1}\fi
\ifx \blpage  \undefined \def \blpage #1{#1}\fi
\ifx \burl  \undefined \def \burl#1{\textsf{#1}}\fi
\ifx \doiurl  \undefined \def \doiurl#1{\url{https://doi.org/#1}}\fi
\ifx \betal  \undefined \def \betal{\textit{et al.}}\fi
\ifx \binstitute  \undefined \def \binstitute#1{#1}\fi
\ifx \binstitutionaled  \undefined \def \binstitutionaled#1{#1}\fi
\ifx \bctitle  \undefined \def \bctitle#1{#1}\fi
\ifx \beditor  \undefined \def \beditor#1{#1}\fi
\ifx \bpublisher  \undefined \def \bpublisher#1{#1}\fi
\ifx \bbtitle  \undefined \def \bbtitle#1{#1}\fi
\ifx \bedition  \undefined \def \bedition#1{#1}\fi
\ifx \bseriesno  \undefined \def \bseriesno#1{#1}\fi
\ifx \blocation  \undefined \def \blocation#1{#1}\fi
\ifx \bsertitle  \undefined \def \bsertitle#1{#1}\fi
\ifx \bsnm \undefined \def \bsnm#1{#1}\fi
\ifx \bsuffix \undefined \def \bsuffix#1{#1}\fi
\ifx \bparticle \undefined \def \bparticle#1{#1}\fi
\ifx \barticle \undefined \def \barticle#1{#1}\fi
\bibcommenthead
\ifx \bconfdate \undefined \def \bconfdate #1{#1}\fi
\ifx \botherref \undefined \def \botherref #1{#1}\fi
\ifx \url \undefined \def \url#1{\textsf{#1}}\fi
\ifx \bchapter \undefined \def \bchapter#1{#1}\fi
\ifx \bbook \undefined \def \bbook#1{#1}\fi
\ifx \bcomment \undefined \def \bcomment#1{#1}\fi
\ifx \oauthor \undefined \def \oauthor#1{#1}\fi
\ifx \citeauthoryear \undefined \def \citeauthoryear#1{#1}\fi
\ifx \endbibitem  \undefined \def \endbibitem {}\fi
\ifx \bconflocation  \undefined \def \bconflocation#1{#1}\fi
\ifx \arxivurl  \undefined \def \arxivurl#1{\textsf{#1}}\fi
\csname PreBibitemsHook\endcsname

\bibitem[\protect\citeauthoryear{Alayrac et~al.}{2022}]{alayrac2022flamingo}
\begin{barticle}
\bauthor{\bsnm{Alayrac}, \binits{J.-B.}},
\bauthor{\bsnm{Donahue}, \binits{J.}},
\bauthor{\bsnm{Luc}, \binits{P.}},
\bauthor{\bsnm{Miech}, \binits{A.}},
\bauthor{\bsnm{Barr}, \binits{I.}},
\bauthor{\bsnm{Hasson}, \binits{Y.}},
\bauthor{\bsnm{Lenc}, \binits{K.}},
\bauthor{\bsnm{Mensch}, \binits{A.}},
\bauthor{\bsnm{Millican}, \binits{K.}},
\bauthor{\bsnm{Reynolds}, \binits{M.}}, \betal:
\batitle{Flamingo: a visual language model for few-shot learning}.
\bjtitle{Advances in neural information processing systems}
\bvolume{35},
\bfpage{23716}--\blpage{23736}
(\byear{2022})
\end{barticle}
\endbibitem

\bibitem[\protect\citeauthoryear{Akata et~al.}{2015}]{akata2015label}
\begin{barticle}
\bauthor{\bsnm{Akata}, \binits{Z.}},
\bauthor{\bsnm{Perronnin}, \binits{F.}},
\bauthor{\bsnm{Harchaoui}, \binits{Z.}},
\bauthor{\bsnm{Schmid}, \binits{C.}}:
\batitle{Label-embedding for image classification}.
\bjtitle{IEEE transactions on pattern analysis and machine intelligence}
\bvolume{38}(\bissue{7}),
\bfpage{1425}--\blpage{1438}
(\byear{2015})
\end{barticle}
\endbibitem

\bibitem[\protect\citeauthoryear{Bai et~al.}{2023}]{bai2023qwen}
\begin{barticle}
\bauthor{\bsnm{Bai}, \binits{J.}},
\bauthor{\bsnm{Bai}, \binits{S.}},
\bauthor{\bsnm{Yang}, \binits{S.}},
\bauthor{\bsnm{Wang}, \binits{S.}},
\bauthor{\bsnm{Tan}, \binits{S.}},
\bauthor{\bsnm{Wang}, \binits{P.}},
\bauthor{\bsnm{Lin}, \binits{J.}},
\bauthor{\bsnm{Zhou}, \binits{C.}},
\bauthor{\bsnm{Zhou}, \binits{J.}}:
\batitle{Qwen-vl: A versatile vision-language model for understanding, localization, text reading, and beyond}.
\bjtitle{arXiv preprint arXiv:2308.12966}
\bvolume{1}(\bissue{2}),
\bfpage{3}
(\byear{2023})
\end{barticle}
\endbibitem

\bibitem[\protect\citeauthoryear{Bossard et~al.}{2014}]{bossard2014food}
\begin{bchapter}
\bauthor{\bsnm{Bossard}, \binits{L.}},
\bauthor{\bsnm{Guillaumin}, \binits{M.}},
\bauthor{\bsnm{Van~Gool}, \binits{L.}}:
\bctitle{Food-101--mining discriminative components with random forests}.
In: \bbtitle{Computer vision--ECCV 2014: 13th European Conference, Zurich, Switzerland, September 6-12, 2014, Proceedings, Part VI 13},
pp. \bfpage{446}--\blpage{461}
(\byear{2014}).
\bcomment{Springer}
\end{bchapter}
\endbibitem

\bibitem[\protect\citeauthoryear{Bishop}{1994}]{bishop1994novelty}
\begin{barticle}
\bauthor{\bsnm{Bishop}, \binits{C.M.}}:
\batitle{Novelty detection and neural network validation}.
\bjtitle{IEE Proceedings-Vision, Image and Signal processing}
\bvolume{141}(\bissue{4}),
\bfpage{217}--\blpage{222}
(\byear{1994})
\end{barticle}
\endbibitem

\bibitem[\protect\citeauthoryear{Berg et~al.}{2014}]{berg2014birdsnap}
\begin{bchapter}
\bauthor{\bsnm{Berg}, \binits{T.}},
\bauthor{\bsnm{Liu}, \binits{J.}},
\bauthor{\bsnm{Woo~Lee}, \binits{S.}},
\bauthor{\bsnm{Alexander}, \binits{M.L.}},
\bauthor{\bsnm{Jacobs}, \binits{D.W.}},
\bauthor{\bsnm{Belhumeur}, \binits{P.N.}}:
\bctitle{Birdsnap: Large-scale fine-grained visual categorization of birds}.
In: \bbtitle{Proceedings of the IEEE Conference on Computer Vision and Pattern Recognition},
pp. \bfpage{2011}--\blpage{2018}
(\byear{2014})
\end{bchapter}
\endbibitem

\bibitem[\protect\citeauthoryear{Byeon et~al.}{2022}]{kakaobrain2022coyo-700m}
\begin{botherref}
\oauthor{\bsnm{Byeon}, \binits{M.}},
\oauthor{\bsnm{Park}, \binits{B.}},
\oauthor{\bsnm{Kim}, \binits{H.}},
\oauthor{\bsnm{Lee}, \binits{S.}},
\oauthor{\bsnm{Baek}, \binits{W.}},
\oauthor{\bsnm{Kim}, \binits{S.}}:
COYO-700M: Image-Text Pair Dataset.
\url{https://github.com/kakaobrain/coyo-dataset}
(2022)
\end{botherref}
\endbibitem

\bibitem[\protect\citeauthoryear{Bouwmans et~al.}{2018}]{bouwmans2018role}
\begin{barticle}
\bauthor{\bsnm{Bouwmans}, \binits{T.}},
\bauthor{\bsnm{Silva}, \binits{C.}},
\bauthor{\bsnm{Marghes}, \binits{C.}},
\bauthor{\bsnm{Zitouni}, \binits{M.S.}},
\bauthor{\bsnm{Bhaskar}, \binits{H.}},
\bauthor{\bsnm{Frelicot}, \binits{C.}}:
\batitle{On the role and the importance of features for background modeling and foreground detection}.
\bjtitle{Computer Science Review}
\bvolume{28},
\bfpage{26}--\blpage{91}
(\byear{2018})
\end{barticle}
\endbibitem

\bibitem[\protect\citeauthoryear{Beyer et~al.}{2024}]{beyer2024paligemma}
\begin{botherref}
\oauthor{\bsnm{Beyer}, \binits{L.}},
\oauthor{\bsnm{Steiner}, \binits{A.}},
\oauthor{\bsnm{Pinto}, \binits{A.S.}},
\oauthor{\bsnm{Kolesnikov}, \binits{A.}},
\oauthor{\bsnm{Wang}, \binits{X.}},
\oauthor{\bsnm{Salz}, \binits{D.}},
\oauthor{\bsnm{Neumann}, \binits{M.}},
\oauthor{\bsnm{Alabdulmohsin}, \binits{I.}},
\oauthor{\bsnm{Tschannen}, \binits{M.}},
\oauthor{\bsnm{Bugliarello}, \binits{E.}}, et al.:
Paligemma: A versatile 3b vlm for transfer.
arXiv preprint arXiv:2407.07726
(2024)
\end{botherref}
\endbibitem

\bibitem[\protect\citeauthoryear{Bansal et~al.}{2018}]{bansal2018zero}
\begin{bchapter}
\bauthor{\bsnm{Bansal}, \binits{A.}},
\bauthor{\bsnm{Sikka}, \binits{K.}},
\bauthor{\bsnm{Sharma}, \binits{G.}},
\bauthor{\bsnm{Chellappa}, \binits{R.}},
\bauthor{\bsnm{Divakaran}, \binits{A.}}:
\bctitle{Zero-shot object detection}.
In: \bbtitle{Proceedings of the European Conference on Computer Vision (ECCV)},
pp. \bfpage{384}--\blpage{400}
(\byear{2018})
\end{bchapter}
\endbibitem

\bibitem[\protect\citeauthoryear{Bejnordi et~al.}{2017}]{bejnordi2017diagnostic}
\begin{barticle}
\bauthor{\bsnm{Bejnordi}, \binits{B.E.}},
\bauthor{\bsnm{Veta}, \binits{M.}},
\bauthor{\bsnm{Van~Diest}, \binits{P.J.}},
\bauthor{\bsnm{Van~Ginneken}, \binits{B.}},
\bauthor{\bsnm{Karssemeijer}, \binits{N.}},
\bauthor{\bsnm{Litjens}, \binits{G.}},
\bauthor{\bsnm{Van Der~Laak}, \binits{J.A.}},
\bauthor{\bsnm{Hermsen}, \binits{M.}},
\bauthor{\bsnm{Manson}, \binits{Q.F.}},
\bauthor{\bsnm{Balkenhol}, \binits{M.}}, \betal:
\batitle{Diagnostic assessment of deep learning algorithms for detection of lymph node metastases in women with breast cancer}.
\bjtitle{Jama}
\bvolume{318}(\bissue{22}),
\bfpage{2199}--\blpage{2210}
(\byear{2017})
\end{barticle}
\endbibitem

\bibitem[\protect\citeauthoryear{Caesar et~al.}{2020}]{caesar2020nuscenes}
\begin{bchapter}
\bauthor{\bsnm{Caesar}, \binits{H.}},
\bauthor{\bsnm{Bankiti}, \binits{V.}},
\bauthor{\bsnm{Lang}, \binits{A.H.}},
\bauthor{\bsnm{Vora}, \binits{S.}},
\bauthor{\bsnm{Liong}, \binits{V.E.}},
\bauthor{\bsnm{Xu}, \binits{Q.}},
\bauthor{\bsnm{Krishnan}, \binits{A.}},
\bauthor{\bsnm{Pan}, \binits{Y.}},
\bauthor{\bsnm{Baldan}, \binits{G.}},
\bauthor{\bsnm{Beijbom}, \binits{O.}}:
\bctitle{nuscenes: A multimodal dataset for autonomous driving}.
In: \bbtitle{Proceedings of the IEEE/CVF Conference on Computer Vision and Pattern Recognition},
pp. \bfpage{11621}--\blpage{11631}
(\byear{2020})
\end{bchapter}
\endbibitem

\bibitem[\protect\citeauthoryear{Cornia et~al.}{2016}]{cornia2016deep}
\begin{bchapter}
\bauthor{\bsnm{Cornia}, \binits{M.}},
\bauthor{\bsnm{Baraldi}, \binits{L.}},
\bauthor{\bsnm{Serra}, \binits{G.}},
\bauthor{\bsnm{Cucchiara}, \binits{R.}}:
\bctitle{A deep multi-level network for saliency prediction}.
In: \bbtitle{2016 23rd International Conference on Pattern Recognition (ICPR)},
pp. \bfpage{3488}--\blpage{3493}
(\byear{2016}).
\bcomment{IEEE}
\end{bchapter}
\endbibitem

\bibitem[\protect\citeauthoryear{Chen et~al.}{2015}]{chen2015microsoft}
\begin{botherref}
\oauthor{\bsnm{Chen}, \binits{X.}},
\oauthor{\bsnm{Fang}, \binits{H.}},
\oauthor{\bsnm{Lin}, \binits{T.-Y.}},
\oauthor{\bsnm{Vedantam}, \binits{R.}},
\oauthor{\bsnm{Gupta}, \binits{S.}},
\oauthor{\bsnm{Doll{\'a}r}, \binits{P.}},
\oauthor{\bsnm{Zitnick}, \binits{C.L.}}:
Microsoft coco captions: Data collection and evaluation server.
arXiv preprint arXiv:1504.00325
(2015)
\end{botherref}
\endbibitem

\bibitem[\protect\citeauthoryear{Chandhok et~al.}{2024}]{chandhok2024response}
\begin{botherref}
\oauthor{\bsnm{Chandhok}, \binits{S.}},
\oauthor{\bsnm{Fan}, \binits{W.-C.}},
\oauthor{\bsnm{Sigal}, \binits{L.}}:
Response wide shut: Surprising observations in basic vision language model capabilities.
arXiv preprint arXiv:2408.06721
(2024)
\end{botherref}
\endbibitem

\bibitem[\protect\citeauthoryear{Cucchiara et~al.}{2003}]{cucchiara2003detecting}
\begin{barticle}
\bauthor{\bsnm{Cucchiara}, \binits{R.}},
\bauthor{\bsnm{Grana}, \binits{C.}},
\bauthor{\bsnm{Piccardi}, \binits{M.}},
\bauthor{\bsnm{Prati}, \binits{A.}}:
\batitle{Detecting moving objects, ghosts, and shadows in video streams}.
\bjtitle{IEEE transactions on pattern analysis and machine intelligence}
\bvolume{25}(\bissue{10}),
\bfpage{1337}--\blpage{1342}
(\byear{2003})
\end{barticle}
\endbibitem

\bibitem[\protect\citeauthoryear{Cheng et~al.}{2017}]{cheng2017remote}
\begin{barticle}
\bauthor{\bsnm{Cheng}, \binits{G.}},
\bauthor{\bsnm{Han}, \binits{J.}},
\bauthor{\bsnm{Lu}, \binits{X.}}:
\batitle{Remote sensing image scene classification: Benchmark and state of the art}.
\bjtitle{Proceedings of the IEEE}
\bvolume{105}(\bissue{10}),
\bfpage{1865}--\blpage{1883}
(\byear{2017})
\end{barticle}
\endbibitem

\bibitem[\protect\citeauthoryear{Chen et~al.}{2020}]{chen2020simple}
\begin{bchapter}
\bauthor{\bsnm{Chen}, \binits{T.}},
\bauthor{\bsnm{Kornblith}, \binits{S.}},
\bauthor{\bsnm{Norouzi}, \binits{M.}},
\bauthor{\bsnm{Hinton}, \binits{G.}}:
\bctitle{A simple framework for contrastive learning of visual representations}.
In: \bbtitle{International Conference on Machine Learning},
pp. \bfpage{1597}--\blpage{1607}
(\byear{2020}).
\bcomment{PMLR}
\end{bchapter}
\endbibitem

\bibitem[\protect\citeauthoryear{Chen et~al.}{2024}]{chen2024we}
\begin{botherref}
\oauthor{\bsnm{Chen}, \binits{L.}},
\oauthor{\bsnm{Li}, \binits{J.}},
\oauthor{\bsnm{Dong}, \binits{X.}},
\oauthor{\bsnm{Zhang}, \binits{P.}},
\oauthor{\bsnm{Zang}, \binits{Y.}},
\oauthor{\bsnm{Chen}, \binits{Z.}},
\oauthor{\bsnm{Duan}, \binits{H.}},
\oauthor{\bsnm{Wang}, \binits{J.}},
\oauthor{\bsnm{Qiao}, \binits{Y.}},
\oauthor{\bsnm{Lin}, \binits{D.}}, et al.:
Are we on the right way for evaluating large vision-language models?
arXiv preprint arXiv:2403.20330
(2024)
\end{botherref}
\endbibitem

\bibitem[\protect\citeauthoryear{Cai et~al.}{2024}]{cai2024vip}
\begin{bchapter}
\bauthor{\bsnm{Cai}, \binits{M.}},
\bauthor{\bsnm{Liu}, \binits{H.}},
\bauthor{\bsnm{Mustikovela}, \binits{S.K.}},
\bauthor{\bsnm{Meyer}, \binits{G.P.}},
\bauthor{\bsnm{Chai}, \binits{Y.}},
\bauthor{\bsnm{Park}, \binits{D.}},
\bauthor{\bsnm{Lee}, \binits{Y.J.}}:
\bctitle{Vip-llava: Making large multimodal models understand arbitrary visual prompts}.
In: \bbtitle{Proceedings of the IEEE/CVF Conference on Computer Vision and Pattern Recognition},
pp. \bfpage{12914}--\blpage{12923}
(\byear{2024})
\end{bchapter}
\endbibitem

\bibitem[\protect\citeauthoryear{Chen et~al.}{2024}]{chen2024bsdp}
\begin{barticle}
\bauthor{\bsnm{Chen}, \binits{Y.}},
\bauthor{\bsnm{Ma}, \binits{L.}},
\bauthor{\bsnm{Jing}, \binits{L.}},
\bauthor{\bsnm{Yu}, \binits{J.}}:
\batitle{Bsdp: Brain-inspired streaming dual-level perturbations for online open world object detection}.
\bjtitle{Pattern Recognition}
\bvolume{152},
\bfpage{110472}
(\byear{2024})
\end{barticle}
\endbibitem

\bibitem[\protect\citeauthoryear{Cimpoi et~al.}{2014a}]{cimpoi14describing}
\begin{bchapter}
\bauthor{\bsnm{Cimpoi}, \binits{M.}},
\bauthor{\bsnm{Maji}, \binits{S.}},
\bauthor{\bsnm{Kokkinos}, \binits{I.}},
\bauthor{\bsnm{Mohamed}, \binits{S.}},
\bauthor{},
\bauthor{\bsnm{Vedaldi}, \binits{A.}}:
\bctitle{Describing textures in the wild}.
In: \bbtitle{Proceedings of the {IEEE} Conf. on Computer Vision and Pattern Recognition ({CVPR})}
(\byear{2014})
\end{bchapter}
\endbibitem

\bibitem[\protect\citeauthoryear{Cimpoi et~al.}{2014b}]{cimpoi2014describing}
\begin{bchapter}
\bauthor{\bsnm{Cimpoi}, \binits{M.}},
\bauthor{\bsnm{Maji}, \binits{S.}},
\bauthor{\bsnm{Kokkinos}, \binits{I.}},
\bauthor{\bsnm{Mohamed}, \binits{S.}},
\bauthor{\bsnm{Vedaldi}, \binits{A.}}:
\bctitle{Describing textures in the wild}.
In: \bbtitle{Proceedings of the IEEE Conference on Computer Vision and Pattern Recognition},
pp. \bfpage{3606}--\blpage{3613}
(\byear{2014})
\end{bchapter}
\endbibitem

\bibitem[\protect\citeauthoryear{Coates et~al.}{2011}]{coates2011analysis}
\begin{bchapter}
\bauthor{\bsnm{Coates}, \binits{A.}},
\bauthor{\bsnm{Ng}, \binits{A.}},
\bauthor{\bsnm{Lee}, \binits{H.}}:
\bctitle{An analysis of single-layer networks in unsupervised feature learning}.
In: \bbtitle{Proceedings of the Fourteenth International Conference on Artificial Intelligence and Statistics},
pp. \bfpage{215}--\blpage{223}
(\byear{2011}).
\bcomment{JMLR Workshop and Conference Proceedings}
\end{bchapter}
\endbibitem

\bibitem[\protect\citeauthoryear{Cordts et~al.}{2016}]{Cordts2016Cityscapes}
\begin{bchapter}
\bauthor{\bsnm{Cordts}, \binits{M.}},
\bauthor{\bsnm{Omran}, \binits{M.}},
\bauthor{\bsnm{Ramos}, \binits{S.}},
\bauthor{\bsnm{Rehfeld}, \binits{T.}},
\bauthor{\bsnm{Enzweiler}, \binits{M.}},
\bauthor{\bsnm{Benenson}, \binits{R.}},
\bauthor{\bsnm{Franke}, \binits{U.}},
\bauthor{\bsnm{Roth}, \binits{S.}},
\bauthor{\bsnm{Schiele}, \binits{B.}}:
\bctitle{The cityscapes dataset for semantic urban scene understanding}.
In: \bbtitle{Proc. of the IEEE Conference on Computer Vision and Pattern Recognition (CVPR)}
(\byear{2016})
\end{bchapter}
\endbibitem

\bibitem[\protect\citeauthoryear{Changpinyo et~al.}{2021}]{changpinyo2021cc12m}
\begin{bchapter}
\bauthor{\bsnm{Changpinyo}, \binits{S.}},
\bauthor{\bsnm{Sharma}, \binits{P.}},
\bauthor{\bsnm{Ding}, \binits{N.}},
\bauthor{\bsnm{Soricut}, \binits{R.}}:
\bctitle{{Conceptual 12M}: Pushing web-scale image-text pre-training to recognize long-tail visual concepts}.
In: \bbtitle{CVPR}
(\byear{2021})
\end{bchapter}
\endbibitem

\bibitem[\protect\citeauthoryear{Cheng et~al.}{2024}]{cheng2024yolo}
\begin{bchapter}
\bauthor{\bsnm{Cheng}, \binits{T.}},
\bauthor{\bsnm{Song}, \binits{L.}},
\bauthor{\bsnm{Ge}, \binits{Y.}},
\bauthor{\bsnm{Liu}, \binits{W.}},
\bauthor{\bsnm{Wang}, \binits{X.}},
\bauthor{\bsnm{Shan}, \binits{Y.}}:
\bctitle{Yolo-world: Real-time open-vocabulary object detection}.
In: \bbtitle{Proceedings of the IEEE/CVF Conference on Computer Vision and Pattern Recognition},
pp. \bfpage{16901}--\blpage{16911}
(\byear{2024})
\end{bchapter}
\endbibitem

\bibitem[\protect\citeauthoryear{Caron et~al.}{2021}]{caron2021emerging}
\begin{bchapter}
\bauthor{\bsnm{Caron}, \binits{M.}},
\bauthor{\bsnm{Touvron}, \binits{H.}},
\bauthor{\bsnm{Misra}, \binits{I.}},
\bauthor{\bsnm{J{\'e}gou}, \binits{H.}},
\bauthor{\bsnm{Mairal}, \binits{J.}},
\bauthor{\bsnm{Bojanowski}, \binits{P.}},
\bauthor{\bsnm{Joulin}, \binits{A.}}:
\bctitle{Emerging properties in self-supervised vision transformers}.
In: \bbtitle{Proceedings of the IEEE/CVF International Conference on Computer Vision},
pp. \bfpage{9650}--\blpage{9660}
(\byear{2021})
\end{bchapter}
\endbibitem

\bibitem[\protect\citeauthoryear{Chen et~al.}{2024}]{chen2024expanding}
\begin{botherref}
\oauthor{\bsnm{Chen}, \binits{Z.}},
\oauthor{\bsnm{Wang}, \binits{W.}},
\oauthor{\bsnm{Cao}, \binits{Y.}},
\oauthor{\bsnm{Liu}, \binits{Y.}},
\oauthor{\bsnm{Gao}, \binits{Z.}},
\oauthor{\bsnm{Cui}, \binits{E.}},
\oauthor{\bsnm{Zhu}, \binits{J.}},
\oauthor{\bsnm{Ye}, \binits{S.}},
\oauthor{\bsnm{Tian}, \binits{H.}},
\oauthor{\bsnm{Liu}, \binits{Z.}}, et al.:
Expanding performance boundaries of open-source multimodal models with model, data, and test-time scaling.
arXiv preprint arXiv:2412.05271
(2024)
\end{botherref}
\endbibitem

\bibitem[\protect\citeauthoryear{Chen et~al.}{2024}]{chen2024far}
\begin{botherref}
\oauthor{\bsnm{Chen}, \binits{Z.}},
\oauthor{\bsnm{Wang}, \binits{W.}},
\oauthor{\bsnm{Tian}, \binits{H.}},
\oauthor{\bsnm{Ye}, \binits{S.}},
\oauthor{\bsnm{Gao}, \binits{Z.}},
\oauthor{\bsnm{Cui}, \binits{E.}},
\oauthor{\bsnm{Tong}, \binits{W.}},
\oauthor{\bsnm{Hu}, \binits{K.}},
\oauthor{\bsnm{Luo}, \binits{J.}},
\oauthor{\bsnm{Ma}, \binits{Z.}}, et al.:
How far are we to gpt-4v? closing the gap to commercial multimodal models with open-source suites.
arXiv preprint arXiv:2404.16821
(2024)
\end{botherref}
\endbibitem

\bibitem[\protect\citeauthoryear{Chen et~al.}{2024}]{chen2024internvl}
\begin{bchapter}
\bauthor{\bsnm{Chen}, \binits{Z.}},
\bauthor{\bsnm{Wu}, \binits{J.}},
\bauthor{\bsnm{Wang}, \binits{W.}},
\bauthor{\bsnm{Su}, \binits{W.}},
\bauthor{\bsnm{Chen}, \binits{G.}},
\bauthor{\bsnm{Xing}, \binits{S.}},
\bauthor{\bsnm{Zhong}, \binits{M.}},
\bauthor{\bsnm{Zhang}, \binits{Q.}},
\bauthor{\bsnm{Zhu}, \binits{X.}},
\bauthor{\bsnm{Lu}, \binits{L.}}, \betal:
\bctitle{Internvl: Scaling up vision foundation models and aligning for generic visual-linguistic tasks}.
In: \bbtitle{Proceedings of the IEEE/CVF Conference on Computer Vision and Pattern Recognition},
pp. \bfpage{24185}--\blpage{24198}
(\byear{2024})
\end{bchapter}
\endbibitem

\bibitem[\protect\citeauthoryear{Chen et~al.}{2024}]{chen2024spatialvlm}
\begin{bchapter}
\bauthor{\bsnm{Chen}, \binits{B.}},
\bauthor{\bsnm{Xu}, \binits{Z.}},
\bauthor{\bsnm{Kirmani}, \binits{S.}},
\bauthor{\bsnm{Ichter}, \binits{B.}},
\bauthor{\bsnm{Sadigh}, \binits{D.}},
\bauthor{\bsnm{Guibas}, \binits{L.}},
\bauthor{\bsnm{Xia}, \binits{F.}}:
\bctitle{Spatialvlm: Endowing vision-language models with spatial reasoning capabilities}.
In: \bbtitle{Proceedings of the IEEE/CVF Conference on Computer Vision and Pattern Recognition},
pp. \bfpage{14455}--\blpage{14465}
(\byear{2024})
\end{bchapter}
\endbibitem

\bibitem[\protect\citeauthoryear{Cen et~al.}{2021}]{cen2021deep}
\begin{bchapter}
\bauthor{\bsnm{Cen}, \binits{J.}},
\bauthor{\bsnm{Yun}, \binits{P.}},
\bauthor{\bsnm{Cai}, \binits{J.}},
\bauthor{\bsnm{Wang}, \binits{M.Y.}},
\bauthor{\bsnm{Liu}, \binits{M.}}:
\bctitle{Deep metric learning for open world semantic segmentation}.
In: \bbtitle{Proceedings of the IEEE/CVF International Conference on Computer Vision},
pp. \bfpage{15333}--\blpage{15342}
(\byear{2021})
\end{bchapter}
\endbibitem

\bibitem[\protect\citeauthoryear{Cai et~al.}{2024}]{cai2024matryoshka}
\begin{botherref}
\oauthor{\bsnm{Cai}, \binits{M.}},
\oauthor{\bsnm{Yang}, \binits{J.}},
\oauthor{\bsnm{Gao}, \binits{J.}},
\oauthor{\bsnm{Lee}, \binits{Y.J.}}:
Matryoshka multimodal models.
arXiv preprint arXiv:2405.17430
(2024)
\end{botherref}
\endbibitem

\bibitem[\protect\citeauthoryear{Chen et~al.}{2024}]{chen2024florence}
\begin{botherref}
\oauthor{\bsnm{Chen}, \binits{J.}},
\oauthor{\bsnm{Yang}, \binits{J.}},
\oauthor{\bsnm{Wu}, \binits{H.}},
\oauthor{\bsnm{Li}, \binits{D.}},
\oauthor{\bsnm{Gao}, \binits{J.}},
\oauthor{\bsnm{Zhou}, \binits{T.}},
\oauthor{\bsnm{Xiao}, \binits{B.}}:
Florence-vl: Enhancing vision-language models with generative vision encoder and depth-breadth fusion.
arXiv preprint arXiv:2412.04424
(2024)
\end{botherref}
\endbibitem

\bibitem[\protect\citeauthoryear{Diao et~al.}{2024}]{diao2024unveiling}
\begin{botherref}
\oauthor{\bsnm{Diao}, \binits{H.}},
\oauthor{\bsnm{Cui}, \binits{Y.}},
\oauthor{\bsnm{Li}, \binits{X.}},
\oauthor{\bsnm{Wang}, \binits{Y.}},
\oauthor{\bsnm{Lu}, \binits{H.}},
\oauthor{\bsnm{Wang}, \binits{X.}}:
Unveiling encoder-free vision-language models.
arXiv preprint arXiv:2406.11832
(2024)
\end{botherref}
\endbibitem

\bibitem[\protect\citeauthoryear{Dai et~al.}{2017}]{dai2017scannet}
\begin{bchapter}
\bauthor{\bsnm{Dai}, \binits{A.}},
\bauthor{\bsnm{Chang}, \binits{A.X.}},
\bauthor{\bsnm{Savva}, \binits{M.}},
\bauthor{\bsnm{Halber}, \binits{M.}},
\bauthor{\bsnm{Funkhouser}, \binits{T.}},
\bauthor{\bsnm{Nie{\ss}ner}, \binits{M.}}:
\bctitle{Scannet: Richly-annotated 3d reconstructions of indoor scenes}.
In: \bbtitle{Proc. Computer Vision and Pattern Recognition (CVPR), IEEE}
(\byear{2017})
\end{bchapter}
\endbibitem

\bibitem[\protect\citeauthoryear{Deng et~al.}{2009}]{deng2009imagenet}
\begin{bchapter}
\bauthor{\bsnm{Deng}, \binits{J.}},
\bauthor{\bsnm{Dong}, \binits{W.}},
\bauthor{\bsnm{Socher}, \binits{R.}},
\bauthor{\bsnm{Li}, \binits{L.-J.}},
\bauthor{\bsnm{Li}, \binits{K.}},
\bauthor{\bsnm{Fei-Fei}, \binits{L.}}:
\bctitle{Imagenet: A large-scale hierarchical image database}.
In: \bbtitle{2009 IEEE Conference on Computer Vision and Pattern Recognition},
pp. \bfpage{248}--\blpage{255}
(\byear{2009}).
\bcomment{Ieee}
\end{bchapter}
\endbibitem

\bibitem[\protect\citeauthoryear{Dhamija et~al.}{2020}]{dhamija2020overlooked}
\begin{bchapter}
\bauthor{\bsnm{Dhamija}, \binits{A.}},
\bauthor{\bsnm{Gunther}, \binits{M.}},
\bauthor{\bsnm{Ventura}, \binits{J.}},
\bauthor{\bsnm{Boult}, \binits{T.}}:
\bctitle{The overlooked elephant of object detection: Open set}.
In: \bbtitle{Proceedings of the IEEE/CVF Winter Conference on Applications of Computer Vision},
pp. \bfpage{1021}--\blpage{1030}
(\byear{2020})
\end{bchapter}
\endbibitem

\bibitem[\protect\citeauthoryear{Desai et~al.}{2021}]{desai2021redcaps}
\begin{botherref}
\oauthor{\bsnm{Desai}, \binits{K.}},
\oauthor{\bsnm{Kaul}, \binits{G.}},
\oauthor{\bsnm{Aysola}, \binits{Z.}},
\oauthor{\bsnm{Johnson}, \binits{J.}}:
Redcaps: Web-curated image-text data created by the people, for the people.
arXiv preprint arXiv:2111.11431
(2021)
\end{botherref}
\endbibitem

\bibitem[\protect\citeauthoryear{Dai et~al.}{2023}]{dai2023instructblip}
\begin{botherref}
\oauthor{\bsnm{Dai}, \binits{W.}},
\oauthor{\bsnm{Li}, \binits{J.}},
\oauthor{\bsnm{Li}, \binits{D.}},
\oauthor{\bsnm{Tiong}, \binits{A.M.H.}},
\oauthor{\bsnm{Zhao}, \binits{J.}},
\oauthor{\bsnm{Wang}, \binits{W.}},
\oauthor{\bsnm{Li}, \binits{B.}},
\oauthor{\bsnm{Fung}, \binits{P.}},
\oauthor{\bsnm{Hoi}, \binits{S.}}:
Instructblip: Towards general-purpose vision-language models with instruction tuning.
arXiv preprint arXiv:2305.06500
(2023)
\end{botherref}
\endbibitem

\bibitem[\protect\citeauthoryear{Dwyer et~al.}{2024}]{dwyer2024roboflow}
\begin{botherref}
\oauthor{\bsnm{Dwyer}, \binits{B.}},
\oauthor{\bsnm{Nelson}, \binits{J.}},
\oauthor{\bsnm{Hansen}, \binits{T.}}, et al.:
Roboflow (Version 1.0).
[Software]
(2024).
\url{https://roboflow.com}
\end{botherref}
\endbibitem

\bibitem[\protect\citeauthoryear{Dalal and Triggs}{2005}]{dalal2005histograms}
\begin{bchapter}
\bauthor{\bsnm{Dalal}, \binits{N.}},
\bauthor{\bsnm{Triggs}, \binits{B.}}:
\bctitle{Histograms of oriented gradients for human detection}.
In: \bbtitle{2005 IEEE Computer Society Conference on Computer Vision and Pattern Recognition (CVPR'05)},
vol. \bseriesno{1},
pp. \bfpage{886}--\blpage{893}
(\byear{2005}).
\bcomment{Ieee}
\end{bchapter}
\endbibitem

\bibitem[\protect\citeauthoryear{Du et~al.}{2022}]{du2022unknown}
\begin{bchapter}
\bauthor{\bsnm{Du}, \binits{X.}},
\bauthor{\bsnm{Wang}, \binits{X.}},
\bauthor{\bsnm{Gozum}, \binits{G.}},
\bauthor{\bsnm{Li}, \binits{Y.}}:
\bctitle{Unknown-aware object detection: Learning what you don't know from videos in the wild}.
In: \bbtitle{Proceedings of the IEEE/CVF Conference on Computer Vision and Pattern Recognition},
pp. \bfpage{13678}--\blpage{13688}
(\byear{2022})
\end{bchapter}
\endbibitem

\bibitem[\protect\citeauthoryear{Everingham et~al.}{2015}]{Everingham15}
\begin{barticle}
\bauthor{\bsnm{Everingham}, \binits{M.}},
\bauthor{\bsnm{Eslami}, \binits{S.M.A.}},
\bauthor{\bsnm{Van~Gool}, \binits{L.}},
\bauthor{\bsnm{Williams}, \binits{C.K.I.}},
\bauthor{\bsnm{Winn}, \binits{J.}},
\bauthor{\bsnm{Zisserman}, \binits{A.}}:
\batitle{The pascal visual object classes challenge: A retrospective}.
\bjtitle{International Journal of Computer Vision}
\bvolume{111}(\bissue{1}),
\bfpage{98}--\blpage{136}
(\byear{2015})
\end{barticle}
\endbibitem

\bibitem[\protect\citeauthoryear{Elgammal et~al.}{2000}]{elgammal2000non}
\begin{bchapter}
\bauthor{\bsnm{Elgammal}, \binits{A.}},
\bauthor{\bsnm{Harwood}, \binits{D.}},
\bauthor{\bsnm{Davis}, \binits{L.}}:
\bctitle{Non-parametric model for background subtraction}.
In: \bbtitle{Computer Vision—ECCV 2000: 6th European Conference on Computer Vision Dublin, Ireland, June 26--July 1, 2000 Proceedings, Part II 6},
pp. \bfpage{751}--\blpage{767}
(\byear{2000}).
\bcomment{Springer}
\end{bchapter}
\endbibitem

\bibitem[\protect\citeauthoryear{Everingham et~al.}{}]{pascal-voc-2012}
\begin{botherref}
\oauthor{\bsnm{Everingham}, \binits{M.}},
\oauthor{\bsnm{Van~Gool}, \binits{L.}},
\oauthor{\bsnm{Williams}, \binits{C.K.I.}},
\oauthor{\bsnm{Winn}, \binits{J.}},
\oauthor{\bsnm{Zisserman}, \binits{A.}}:
The {PASCAL} {V}isual {O}bject {C}lasses {C}hallenge 2012 {(VOC2012)} {R}esults.
http://www.pascal-network.org/challenges/VOC/voc2012/workshop/index.html
\end{botherref}
\endbibitem

\bibitem[\protect\citeauthoryear{Frome et~al.}{2013}]{frome2013devise}
\begin{botherref}
\oauthor{\bsnm{Frome}, \binits{A.}},
\oauthor{\bsnm{Corrado}, \binits{G.S.}},
\oauthor{\bsnm{Shlens}, \binits{J.}},
\oauthor{\bsnm{Bengio}, \binits{S.}},
\oauthor{\bsnm{Dean}, \binits{J.}},
\oauthor{\bsnm{Ranzato}, \binits{M.}},
\oauthor{\bsnm{Mikolov}, \binits{T.}}:
Devise: A deep visual-semantic embedding model.
Advances in neural information processing systems
\textbf{26}
(2013)
\end{botherref}
\endbibitem

\bibitem[\protect\citeauthoryear{Felzenszwalb et~al.}{2009}]{felzenszwalb2009object}
\begin{barticle}
\bauthor{\bsnm{Felzenszwalb}, \binits{P.F.}},
\bauthor{\bsnm{Girshick}, \binits{R.B.}},
\bauthor{\bsnm{McAllester}, \binits{D.}},
\bauthor{\bsnm{Ramanan}, \binits{D.}}:
\batitle{Object detection with discriminatively trained part-based models}.
\bjtitle{IEEE transactions on pattern analysis and machine intelligence}
\bvolume{32}(\bissue{9}),
\bfpage{1627}--\blpage{1645}
(\byear{2009})
\end{barticle}
\endbibitem

\bibitem[\protect\citeauthoryear{Fort et~al.}{2021}]{fort2021exploring}
\begin{barticle}
\bauthor{\bsnm{Fort}, \binits{S.}},
\bauthor{\bsnm{Ren}, \binits{J.}},
\bauthor{\bsnm{Lakshminarayanan}, \binits{B.}}:
\batitle{Exploring the limits of out-of-distribution detection}.
\bjtitle{Advances in Neural Information Processing Systems}
\bvolume{34},
\bfpage{7068}--\blpage{7081}
(\byear{2021})
\end{barticle}
\endbibitem

\bibitem[\protect\citeauthoryear{Freund et~al.}{1999}]{freund1999short}
\begin{barticle}
\bauthor{\bsnm{Freund}, \binits{Y.}},
\bauthor{\bsnm{Schapire}, \binits{R.}},
\bauthor{\bsnm{Abe}, \binits{N.}}:
\batitle{A short introduction to boosting}.
\bjtitle{Journal-Japanese Society For Artificial Intelligence}
\bvolume{14}(\bissue{771-780}),
\bfpage{1612}
(\byear{1999})
\end{barticle}
\endbibitem

\bibitem[\protect\citeauthoryear{Feng et~al.}{2022}]{feng2022promptdet}
\begin{bchapter}
\bauthor{\bsnm{Feng}, \binits{C.}},
\bauthor{\bsnm{Zhong}, \binits{Y.}},
\bauthor{\bsnm{Jie}, \binits{Z.}},
\bauthor{\bsnm{Chu}, \binits{X.}},
\bauthor{\bsnm{Ren}, \binits{H.}},
\bauthor{\bsnm{Wei}, \binits{X.}},
\bauthor{\bsnm{Xie}, \binits{W.}},
\bauthor{\bsnm{Ma}, \binits{L.}}:
\bctitle{Promptdet: Towards open-vocabulary detection using uncurated images}.
In: \bbtitle{European Conference on Computer Vision},
pp. \bfpage{701}--\blpage{717}
(\byear{2022}).
\bcomment{Springer}
\end{bchapter}
\endbibitem

\bibitem[\protect\citeauthoryear{Girshick et~al.}{2014}]{girshick2014rich}
\begin{bchapter}
\bauthor{\bsnm{Girshick}, \binits{R.}},
\bauthor{\bsnm{Donahue}, \binits{J.}},
\bauthor{\bsnm{Darrell}, \binits{T.}},
\bauthor{\bsnm{Malik}, \binits{J.}}:
\bctitle{Rich feature hierarchies for accurate object detection and semantic segmentation}.
In: \bbtitle{Proceedings of the IEEE Conference on Computer Vision and Pattern Recognition},
pp. \bfpage{580}--\blpage{587}
(\byear{2014})
\end{bchapter}
\endbibitem

\bibitem[\protect\citeauthoryear{Gupta et~al.}{2019}]{gupta2019lvis}
\begin{bchapter}
\bauthor{\bsnm{Gupta}, \binits{A.}},
\bauthor{\bsnm{Dollar}, \binits{P.}},
\bauthor{\bsnm{Girshick}, \binits{R.}}:
\bctitle{Lvis: A dataset for large vocabulary instance segmentation}.
In: \bbtitle{Proceedings of the IEEE/CVF Conference on Computer Vision and Pattern Recognition},
pp. \bfpage{5356}--\blpage{5364}
(\byear{2019})
\end{bchapter}
\endbibitem

\bibitem[\protect\citeauthoryear{Goodfellow et~al.}{2013}]{goodfellow2013challenges}
\begin{bchapter}
\bauthor{\bsnm{Goodfellow}, \binits{I.J.}},
\bauthor{\bsnm{Erhan}, \binits{D.}},
\bauthor{\bsnm{Carrier}, \binits{P.L.}},
\bauthor{\bsnm{Courville}, \binits{A.}},
\bauthor{\bsnm{Mirza}, \binits{M.}},
\bauthor{\bsnm{Hamner}, \binits{B.}},
\bauthor{\bsnm{Cukierski}, \binits{W.}},
\bauthor{\bsnm{Tang}, \binits{Y.}},
\bauthor{\bsnm{Thaler}, \binits{D.}},
\bauthor{\bsnm{Lee}, \binits{D.-H.}}, \betal:
\bctitle{Challenges in representation learning: A report on three machine learning contests}.
In: \bbtitle{Neural Information Processing: 20th International Conference, ICONIP 2013, Daegu, Korea, November 3-7, 2013. Proceedings, Part III 20},
pp. \bfpage{117}--\blpage{124}
(\byear{2013}).
\bcomment{Springer}
\end{bchapter}
\endbibitem

\bibitem[\protect\citeauthoryear{Girshick}{2015}]{girshick2015fast}
\begin{botherref}
\oauthor{\bsnm{Girshick}, \binits{R.}}:
Fast r-cnn.
arXiv preprint arXiv:1504.08083
(2015)
\end{botherref}
\endbibitem

\bibitem[\protect\citeauthoryear{Geiger et~al.}{2012}]{Geiger2012CVPR}
\begin{bchapter}
\bauthor{\bsnm{Geiger}, \binits{A.}},
\bauthor{\bsnm{Lenz}, \binits{P.}},
\bauthor{\bsnm{Urtasun}, \binits{R.}}:
\bctitle{Are we ready for autonomous driving? the kitti vision benchmark suite}.
In: \bbtitle{Conference on Computer Vision and Pattern Recognition (CVPR)}
(\byear{2012})
\end{bchapter}
\endbibitem

\bibitem[\protect\citeauthoryear{Guan et~al.}{2024}]{guan2024hallusionbench}
\begin{bchapter}
\bauthor{\bsnm{Guan}, \binits{T.}},
\bauthor{\bsnm{Liu}, \binits{F.}},
\bauthor{\bsnm{Wu}, \binits{X.}},
\bauthor{\bsnm{Xian}, \binits{R.}},
\bauthor{\bsnm{Li}, \binits{Z.}},
\bauthor{\bsnm{Liu}, \binits{X.}},
\bauthor{\bsnm{Wang}, \binits{X.}},
\bauthor{\bsnm{Chen}, \binits{L.}},
\bauthor{\bsnm{Huang}, \binits{F.}},
\bauthor{\bsnm{Yacoob}, \binits{Y.}}, \betal:
\bctitle{Hallusionbench: an advanced diagnostic suite for entangled language hallucination and visual illusion in large vision-language models}.
In: \bbtitle{Proceedings of the IEEE/CVF Conference on Computer Vision and Pattern Recognition},
pp. \bfpage{14375}--\blpage{14385}
(\byear{2024})
\end{bchapter}
\endbibitem

\bibitem[\protect\citeauthoryear{Goferman et~al.}{2012}]{goferman2012context}
\begin{botherref}
\oauthor{\bsnm{Goferman}, \binits{S.}},
\oauthor{\bsnm{Zelnik-Manor}, \binits{L.}},
\oauthor{\bsnm{Tal}, \binits{A.}}:
Context-aware saliency detection. ieee transactions on pattern analysis and machine intelligence
(2012)
\end{botherref}
\endbibitem

\bibitem[\protect\citeauthoryear{Hiippala et~al.}{2021}]{hiippala2021ai2d}
\begin{barticle}
\bauthor{\bsnm{Hiippala}, \binits{T.}},
\bauthor{\bsnm{Alikhani}, \binits{M.}},
\bauthor{\bsnm{Haverinen}, \binits{J.}},
\bauthor{\bsnm{Kalliokoski}, \binits{T.}},
\bauthor{\bsnm{Logacheva}, \binits{E.}},
\bauthor{\bsnm{Orekhova}, \binits{S.}},
\bauthor{\bsnm{Tuomainen}, \binits{A.}},
\bauthor{\bsnm{Stone}, \binits{M.}},
\bauthor{\bsnm{Bateman}, \binits{J.A.}}:
\batitle{Ai2d-rst: A multimodal corpus of 1000 primary school science diagrams}.
\bjtitle{Language Resources and Evaluation}
\bvolume{55},
\bfpage{661}--\blpage{688}
(\byear{2021})
\end{barticle}
\endbibitem

\bibitem[\protect\citeauthoryear{Helber et~al.}{2019}]{helber2019eurosat}
\begin{barticle}
\bauthor{\bsnm{Helber}, \binits{P.}},
\bauthor{\bsnm{Bischke}, \binits{B.}},
\bauthor{\bsnm{Dengel}, \binits{A.}},
\bauthor{\bsnm{Borth}, \binits{D.}}:
\batitle{Eurosat: A novel dataset and deep learning benchmark for land use and land cover classification}.
\bjtitle{IEEE Journal of Selected Topics in Applied Earth Observations and Remote Sensing}
\bvolume{12}(\bissue{7}),
\bfpage{2217}--\blpage{2226}
(\byear{2019})
\end{barticle}
\endbibitem

\bibitem[\protect\citeauthoryear{He et~al.}{2022}]{he2022masked}
\begin{bchapter}
\bauthor{\bsnm{He}, \binits{K.}},
\bauthor{\bsnm{Chen}, \binits{X.}},
\bauthor{\bsnm{Xie}, \binits{S.}},
\bauthor{\bsnm{Li}, \binits{Y.}},
\bauthor{\bsnm{Doll{\'a}r}, \binits{P.}},
\bauthor{\bsnm{Girshick}, \binits{R.}}:
\bctitle{Masked autoencoders are scalable vision learners}.
In: \bbtitle{Proceedings of the IEEE/CVF Conference on Computer Vision and Pattern Recognition},
pp. \bfpage{16000}--\blpage{16009}
(\byear{2022})
\end{bchapter}
\endbibitem

\bibitem[\protect\citeauthoryear{Hearst et~al.}{1998}]{hearst1998support}
\begin{barticle}
\bauthor{\bsnm{Hearst}, \binits{M.A.}},
\bauthor{\bsnm{Dumais}, \binits{S.T.}},
\bauthor{\bsnm{Osuna}, \binits{E.}},
\bauthor{\bsnm{Platt}, \binits{J.}},
\bauthor{\bsnm{Scholkopf}, \binits{B.}}:
\batitle{Support vector machines}.
\bjtitle{IEEE Intelligent Systems and their applications}
\bvolume{13}(\bissue{4}),
\bfpage{18}--\blpage{28}
(\byear{1998})
\end{barticle}
\endbibitem

\bibitem[\protect\citeauthoryear{Hendrycks and Gimpel}{2016}]{hendrycks2016baseline}
\begin{botherref}
\oauthor{\bsnm{Hendrycks}, \binits{D.}},
\oauthor{\bsnm{Gimpel}, \binits{K.}}:
A baseline for detecting misclassified and out-of-distribution examples in neural networks.
arXiv preprint arXiv:1610.02136
(2016)
\end{botherref}
\endbibitem

\bibitem[\protect\citeauthoryear{Hu et~al.}{2022}]{hu2022scaling}
\begin{bchapter}
\bauthor{\bsnm{Hu}, \binits{X.}},
\bauthor{\bsnm{Gan}, \binits{Z.}},
\bauthor{\bsnm{Wang}, \binits{J.}},
\bauthor{\bsnm{Yang}, \binits{Z.}},
\bauthor{\bsnm{Liu}, \binits{Z.}},
\bauthor{\bsnm{Lu}, \binits{Y.}},
\bauthor{\bsnm{Wang}, \binits{L.}}:
\bctitle{Scaling up vision-language pre-training for image captioning}.
In: \bbtitle{Proceedings of the IEEE/CVF Conference on Computer Vision and Pattern Recognition},
pp. \bfpage{17980}--\blpage{17989}
(\byear{2022})
\end{bchapter}
\endbibitem

\bibitem[\protect\citeauthoryear{Hayat et~al.}{2020}]{hayat2020synthesizing}
\begin{bchapter}
\bauthor{\bsnm{Hayat}, \binits{N.}},
\bauthor{\bsnm{Hayat}, \binits{M.}},
\bauthor{\bsnm{Rahman}, \binits{S.}},
\bauthor{\bsnm{Khan}, \binits{S.}},
\bauthor{\bsnm{Zamir}, \binits{S.W.}},
\bauthor{\bsnm{Khan}, \binits{F.S.}}:
\bctitle{Synthesizing the unseen for zero-shot object detection}.
In: \bbtitle{Proceedings of the Asian Conference on Computer Vision}
(\byear{2020})
\end{bchapter}
\endbibitem

\bibitem[\protect\citeauthoryear{Hudson and Manning}{2019}]{hudson2019gqa}
\begin{bchapter}
\bauthor{\bsnm{Hudson}, \binits{D.A.}},
\bauthor{\bsnm{Manning}, \binits{C.D.}}:
\bctitle{Gqa: A new dataset for real-world visual reasoning and compositional question answering}.
In: \bbtitle{Proceedings of the IEEE/CVF Conference on Computer Vision and Pattern Recognition},
pp. \bfpage{6700}--\blpage{6709}
(\byear{2019})
\end{bchapter}
\endbibitem

\bibitem[\protect\citeauthoryear{Hochreiter}{1997}]{hochreiter1997long}
\begin{botherref}
\oauthor{\bsnm{Hochreiter}, \binits{S.}}:
Long short-term memory.
Neural Computation MIT-Press
(1997)
\end{botherref}
\endbibitem

\bibitem[\protect\citeauthoryear{Hou and Zhang}{2007}]{hou2007saliency}
\begin{bchapter}
\bauthor{\bsnm{Hou}, \binits{X.}},
\bauthor{\bsnm{Zhang}, \binits{L.}}:
\bctitle{Saliency detection: A spectral residual approach}.
In: \bbtitle{2007 IEEE Conference on Computer Vision and Pattern Recognition},
pp. \bfpage{1}--\blpage{8}
(\byear{2007}).
\bcomment{Ieee}
\end{bchapter}
\endbibitem

\bibitem[\protect\citeauthoryear{He et~al.}{2015}]{he2015spatial}
\begin{barticle}
\bauthor{\bsnm{He}, \binits{K.}},
\bauthor{\bsnm{Zhang}, \binits{X.}},
\bauthor{\bsnm{Ren}, \binits{S.}},
\bauthor{\bsnm{Sun}, \binits{J.}}:
\batitle{Spatial pyramid pooling in deep convolutional networks for visual recognition}.
\bjtitle{IEEE transactions on pattern analysis and machine intelligence}
\bvolume{37}(\bissue{9}),
\bfpage{1904}--\blpage{1916}
(\byear{2015})
\end{barticle}
\endbibitem

\bibitem[\protect\citeauthoryear{He et~al.}{2016}]{he2016deep}
\begin{bchapter}
\bauthor{\bsnm{He}, \binits{K.}},
\bauthor{\bsnm{Zhang}, \binits{X.}},
\bauthor{\bsnm{Ren}, \binits{S.}},
\bauthor{\bsnm{Sun}, \binits{J.}}:
\bctitle{Deep residual learning for image recognition}.
In: \bbtitle{Proceedings of the IEEE Conference on Computer Vision and Pattern Recognition},
pp. \bfpage{770}--\blpage{778}
(\byear{2016})
\end{bchapter}
\endbibitem

\bibitem[\protect\citeauthoryear{Itti et~al.}{1998}]{730558}
\begin{barticle}
\bauthor{\bsnm{Itti}, \binits{L.}},
\bauthor{\bsnm{Koch}, \binits{C.}},
\bauthor{\bsnm{Niebur}, \binits{E.}}:
\batitle{A model of saliency-based visual attention for rapid scene analysis}.
\bjtitle{IEEE Transactions on Pattern Analysis and Machine Intelligence}
\bvolume{20}(\bissue{11}),
\bfpage{1254}--\blpage{1259}
(\byear{1998})
\doiurl{10.1109/34.730558}
\end{barticle}
\endbibitem

\bibitem[\protect\citeauthoryear{Johnson et~al.}{2017}]{johnson2017clevr}
\begin{bchapter}
\bauthor{\bsnm{Johnson}, \binits{J.}},
\bauthor{\bsnm{Hariharan}, \binits{B.}},
\bauthor{\bsnm{Van Der~Maaten}, \binits{L.}},
\bauthor{\bsnm{Fei-Fei}, \binits{L.}},
\bauthor{\bsnm{Lawrence~Zitnick}, \binits{C.}},
\bauthor{\bsnm{Girshick}, \binits{R.}}:
\bctitle{Clevr: A diagnostic dataset for compositional language and elementary visual reasoning}.
In: \bbtitle{Proceedings of the IEEE Conference on Computer Vision and Pattern Recognition},
pp. \bfpage{2901}--\blpage{2910}
(\byear{2017})
\end{bchapter}
\endbibitem

\bibitem[\protect\citeauthoryear{Joseph et~al.}{2021}]{joseph2021towards}
\begin{bchapter}
\bauthor{\bsnm{Joseph}, \binits{K.}},
\bauthor{\bsnm{Khan}, \binits{S.}},
\bauthor{\bsnm{Khan}, \binits{F.S.}},
\bauthor{\bsnm{Balasubramanian}, \binits{V.N.}}:
\bctitle{Towards open world object detection}.
In: \bbtitle{Proceedings of the IEEE/CVF Conference on Computer Vision and Pattern Recognition},
pp. \bfpage{5830}--\blpage{5840}
(\byear{2021})
\end{bchapter}
\endbibitem

\bibitem[\protect\citeauthoryear{Jain et~al.}{2024}]{jain2024vcoder}
\begin{bchapter}
\bauthor{\bsnm{Jain}, \binits{J.}},
\bauthor{\bsnm{Yang}, \binits{J.}},
\bauthor{\bsnm{Shi}, \binits{H.}}:
\bctitle{Vcoder: Versatile vision encoders for multimodal large language models}.
In: \bbtitle{Proceedings of the IEEE/CVF Conference on Computer Vision and Pattern Recognition},
pp. \bfpage{27992}--\blpage{28002}
(\byear{2024})
\end{bchapter}
\endbibitem

\bibitem[\protect\citeauthoryear{Jia et~al.}{2021}]{jia2021scaling}
\begin{bchapter}
\bauthor{\bsnm{Jia}, \binits{C.}},
\bauthor{\bsnm{Yang}, \binits{Y.}},
\bauthor{\bsnm{Xia}, \binits{Y.}},
\bauthor{\bsnm{Chen}, \binits{Y.-T.}},
\bauthor{\bsnm{Parekh}, \binits{Z.}},
\bauthor{\bsnm{Pham}, \binits{H.}},
\bauthor{\bsnm{Le}, \binits{Q.}},
\bauthor{\bsnm{Sung}, \binits{Y.-H.}},
\bauthor{\bsnm{Li}, \binits{Z.}},
\bauthor{\bsnm{Duerig}, \binits{T.}}:
\bctitle{Scaling up visual and vision-language representation learning with noisy text supervision}.
In: \bbtitle{International Conference on Machine Learning},
pp. \bfpage{4904}--\blpage{4916}
(\byear{2021}).
\bcomment{PMLR}
\end{bchapter}
\endbibitem

\bibitem[\protect\citeauthoryear{Kusupati et~al.}{2022}]{kusupati2022matryoshka}
\begin{barticle}
\bauthor{\bsnm{Kusupati}, \binits{A.}},
\bauthor{\bsnm{Bhatt}, \binits{G.}},
\bauthor{\bsnm{Rege}, \binits{A.}},
\bauthor{\bsnm{Wallingford}, \binits{M.}},
\bauthor{\bsnm{Sinha}, \binits{A.}},
\bauthor{\bsnm{Ramanujan}, \binits{V.}},
\bauthor{\bsnm{Howard-Snyder}, \binits{W.}},
\bauthor{\bsnm{Chen}, \binits{K.}},
\bauthor{\bsnm{Kakade}, \binits{S.}},
\bauthor{\bsnm{Jain}, \binits{P.}}, \betal:
\batitle{Matryoshka representation learning}.
\bjtitle{Advances in Neural Information Processing Systems}
\bvolume{35},
\bfpage{30233}--\blpage{30249}
(\byear{2022})
\end{barticle}
\endbibitem

\bibitem[\protect\citeauthoryear{Kim et~al.}{2005}]{kim2005real}
\begin{barticle}
\bauthor{\bsnm{Kim}, \binits{K.}},
\bauthor{\bsnm{Chalidabhongse}, \binits{T.H.}},
\bauthor{\bsnm{Harwood}, \binits{D.}},
\bauthor{\bsnm{Davis}, \binits{L.}}:
\batitle{Real-time foreground--background segmentation using codebook model}.
\bjtitle{Real-time imaging}
\bvolume{11}(\bissue{3}),
\bfpage{172}--\blpage{185}
(\byear{2005})
\end{barticle}
\endbibitem

\bibitem[\protect\citeauthoryear{Krasin et~al.}{2016}]{openimages}
\begin{botherref}
\oauthor{\bsnm{Krasin}, \binits{I.}},
\oauthor{\bsnm{Duerig}, \binits{T.}},
\oauthor{\bsnm{Alldrin}, \binits{N.}},
\oauthor{\bsnm{Veit}, \binits{A.}},
\oauthor{\bsnm{Abu-El-Haija}, \binits{S.}},
\oauthor{\bsnm{Belongie}, \binits{S.}},
\oauthor{\bsnm{Cai}, \binits{D.}},
\oauthor{\bsnm{Feng}, \binits{Z.}},
\oauthor{\bsnm{Ferrari}, \binits{V.}},
\oauthor{\bsnm{Gomes}, \binits{V.}},
\oauthor{\bsnm{Gupta}, \binits{A.}},
\oauthor{\bsnm{Narayanan}, \binits{D.}},
\oauthor{\bsnm{Sun}, \binits{C.}},
\oauthor{\bsnm{Chechik}, \binits{G.}},
\oauthor{\bsnm{Murphy}, \binits{K.}}:
Openimages: A public dataset for large-scale multi-label and multi-class image classification.
Dataset available from https://github.com/openimages
(2016)
\end{botherref}
\endbibitem

\bibitem[\protect\citeauthoryear{Kamboj and Driggs-Campbell}{2024}]{kamboj2024brief}
\begin{botherref}
\oauthor{\bsnm{Kamboj}, \binits{A.}},
\oauthor{\bsnm{Driggs-Campbell}, \binits{K.}}:
A brief survey on leveraging large scale vision models for enhanced robot grasping.
arXiv preprint arXiv:2406.11786
(2024)
\end{botherref}
\endbibitem

\bibitem[\protect\citeauthoryear{Kiela et~al.}{2020}]{kiela2020hateful}
\begin{barticle}
\bauthor{\bsnm{Kiela}, \binits{D.}},
\bauthor{\bsnm{Firooz}, \binits{H.}},
\bauthor{\bsnm{Mohan}, \binits{A.}},
\bauthor{\bsnm{Goswami}, \binits{V.}},
\bauthor{\bsnm{Singh}, \binits{A.}},
\bauthor{\bsnm{Ringshia}, \binits{P.}},
\bauthor{\bsnm{Testuggine}, \binits{D.}}:
\batitle{The hateful memes challenge: Detecting hate speech in multimodal memes}.
\bjtitle{Advances in neural information processing systems}
\bvolume{33},
\bfpage{2611}--\blpage{2624}
(\byear{2020})
\end{barticle}
\endbibitem

\bibitem[\protect\citeauthoryear{Krizhevsky et~al.}{2009}]{krizhevsky2009learning}
\begin{botherref}
\oauthor{\bsnm{Krizhevsky}, \binits{A.}},
\oauthor{\bsnm{Hinton}, \binits{G.}}, et al.:
Learning multiple layers of features from tiny images
(2009)
\end{botherref}
\endbibitem

\bibitem[\protect\citeauthoryear{Kebe et~al.}{2021}]{kebe2021a}
\begin{bchapter}
\bauthor{\bsnm{Kebe}, \binits{G.Y.}},
\bauthor{\bsnm{Higgins}, \binits{P.}},
\bauthor{\bsnm{Jenkins}, \binits{P.}},
\bauthor{\bsnm{Darvish}, \binits{K.}},
\bauthor{\bsnm{Sachdeva}, \binits{R.}},
\bauthor{\bsnm{Barron}, \binits{R.}},
\bauthor{\bsnm{Winder}, \binits{J.}},
\bauthor{\bsnm{Engel}, \binits{D.}},
\bauthor{\bsnm{Raff}, \binits{E.}},
\bauthor{\bsnm{Ferraro}, \binits{F.}},
\bauthor{\bsnm{Matuszek}, \binits{C.}}:
\bctitle{A spoken language dataset of descriptions for speech-based grounded language learning}.
In: \bbtitle{Thirty-fifth Conference on Neural Information Processing Systems Datasets and Benchmarks Track (Round 1)}
(\byear{2021}).
\burl{https://openreview.net/forum?id=Yx9jT3fkBaD}
\end{bchapter}
\endbibitem

\bibitem[\protect\citeauthoryear{Kirillov et~al.}{2023}]{kirillov2023segment}
\begin{bchapter}
\bauthor{\bsnm{Kirillov}, \binits{A.}},
\bauthor{\bsnm{Mintun}, \binits{E.}},
\bauthor{\bsnm{Ravi}, \binits{N.}},
\bauthor{\bsnm{Mao}, \binits{H.}},
\bauthor{\bsnm{Rolland}, \binits{C.}},
\bauthor{\bsnm{Gustafson}, \binits{L.}},
\bauthor{\bsnm{Xiao}, \binits{T.}},
\bauthor{\bsnm{Whitehead}, \binits{S.}},
\bauthor{\bsnm{Berg}, \binits{A.C.}},
\bauthor{\bsnm{Lo}, \binits{W.-Y.}}, \betal:
\bctitle{Segment anything}.
In: \bbtitle{Proceedings of the IEEE/CVF International Conference on Computer Vision},
pp. \bfpage{4015}--\blpage{4026}
(\byear{2023})
\end{bchapter}
\endbibitem

\bibitem[\protect\citeauthoryear{Krause et~al.}{2013}]{KrauseStarkDengFei-Fei_3DRR2013}
\begin{bchapter}
\bauthor{\bsnm{Krause}, \binits{J.}},
\bauthor{\bsnm{Stark}, \binits{M.}},
\bauthor{\bsnm{Deng}, \binits{J.}},
\bauthor{\bsnm{Fei-Fei}, \binits{L.}}:
\bctitle{3d object representations for fine-grained categorization}.
In: \bbtitle{4th International IEEE Workshop on 3D Representation and Recognition (3dRR-13)},
\bconflocation{Sydney, Australia}
(\byear{2013})
\end{bchapter}
\endbibitem

\bibitem[\protect\citeauthoryear{Krizhevsky et~al.}{2012}]{krizhevsky2012imagenet}
\begin{botherref}
\oauthor{\bsnm{Krizhevsky}, \binits{A.}},
\oauthor{\bsnm{Sutskever}, \binits{I.}},
\oauthor{\bsnm{Hinton}, \binits{G.E.}}:
Imagenet classification with deep convolutional neural networks.
Advances in neural information processing systems
\textbf{25}
(2012)
\end{botherref}
\endbibitem

\bibitem[\protect\citeauthoryear{Kiros et~al.}{2014}]{kiros2014unifying}
\begin{botherref}
\oauthor{\bsnm{Kiros}, \binits{R.}},
\oauthor{\bsnm{Salakhutdinov}, \binits{R.}},
\oauthor{\bsnm{Zemel}, \binits{R.S.}}:
Unifying visual-semantic embeddings with multimodal neural language models.
arXiv preprint arXiv:1411.2539
(2014)
\end{botherref}
\endbibitem

\bibitem[\protect\citeauthoryear{Kenton and Toutanova}{2019}]{kenton2019bert}
\begin{bchapter}
\bauthor{\bsnm{Kenton}, \binits{J.D.M.-W.C.}},
\bauthor{\bsnm{Toutanova}, \binits{L.K.}}:
\bctitle{Bert: Pre-training of deep bidirectional transformers for language understanding}.
In: \bbtitle{Proceedings of naacL-HLT},
vol. \bseriesno{1}
(\byear{2019}).
\bcomment{Minneapolis, Minnesota}
\end{bchapter}
\endbibitem

\bibitem[\protect\citeauthoryear{Koller et~al.}{1994}]{koller1994towards}
\begin{bchapter}
\bauthor{\bsnm{Koller}, \binits{D.}},
\bauthor{\bsnm{Weber}, \binits{J.}},
\bauthor{\bsnm{Huang}, \binits{T.}},
\bauthor{\bsnm{Malik}, \binits{J.}},
\bauthor{\bsnm{Ogasawara}, \binits{G.}},
\bauthor{\bsnm{Rao}, \binits{B.}},
\bauthor{\bsnm{Russell}, \binits{S.}}:
\bctitle{Towards robust automatic traffic scene analysis in real-time}.
In: \bbtitle{Proceedings of 12th International Conference on Pattern Recognition},
vol. \bseriesno{1},
pp. \bfpage{126}--\blpage{131}
(\byear{1994}).
\bcomment{IEEE}
\end{bchapter}
\endbibitem

\bibitem[\protect\citeauthoryear{Krishna et~al.}{2017}]{krishna2017visual}
\begin{barticle}
\bauthor{\bsnm{Krishna}, \binits{R.}},
\bauthor{\bsnm{Zhu}, \binits{Y.}},
\bauthor{\bsnm{Groth}, \binits{O.}},
\bauthor{\bsnm{Johnson}, \binits{J.}},
\bauthor{\bsnm{Hata}, \binits{K.}},
\bauthor{\bsnm{Kravitz}, \binits{J.}},
\bauthor{\bsnm{Chen}, \binits{S.}},
\bauthor{\bsnm{Kalantidis}, \binits{Y.}},
\bauthor{\bsnm{Li}, \binits{L.-J.}},
\bauthor{\bsnm{Shamma}, \binits{D.A.}}, \betal:
\batitle{Visual genome: Connecting language and vision using crowdsourced dense image annotations}.
\bjtitle{International journal of computer vision}
\bvolume{123},
\bfpage{32}--\blpage{73}
(\byear{2017})
\end{barticle}
\endbibitem

\bibitem[\protect\citeauthoryear{Liu et~al.}{2016}]{liu2016ssd}
\begin{bchapter}
\bauthor{\bsnm{Liu}, \binits{W.}},
\bauthor{\bsnm{Anguelov}, \binits{D.}},
\bauthor{\bsnm{Erhan}, \binits{D.}},
\bauthor{\bsnm{Szegedy}, \binits{C.}},
\bauthor{\bsnm{Reed}, \binits{S.}},
\bauthor{\bsnm{Fu}, \binits{C.-Y.}},
\bauthor{\bsnm{Berg}, \binits{A.C.}}:
\bctitle{Ssd: Single shot multibox detector}.
In: \bbtitle{Computer Vision--ECCV 2016: 14th European Conference, Amsterdam, The Netherlands, October 11--14, 2016, Proceedings, Part I 14},
pp. \bfpage{21}--\blpage{37}
(\byear{2016}).
\bcomment{Springer}
\end{bchapter}
\endbibitem

\bibitem[\protect\citeauthoryear{Li et~al.}{2022}]{li2022caltech}
\begin{botherref}
\oauthor{\bsnm{Li}, \binits{F.}},
\oauthor{\bsnm{Andreeto}, \binits{M.}},
\oauthor{\bsnm{Ranzato}, \binits{M.}},
\oauthor{\bsnm{Perona}, \binits{P.}}:
Caltech 101 (1.0)[Data set]. CaltechDATA.
Tech. Rep., 2022, doi: 10.22002
(2022)
\end{botherref}
\endbibitem

\bibitem[\protect\citeauthoryear{LeCun et~al.}{1998}]{lecun1998gradient}
\begin{barticle}
\bauthor{\bsnm{LeCun}, \binits{Y.}},
\bauthor{\bsnm{Bottou}, \binits{L.}},
\bauthor{\bsnm{Bengio}, \binits{Y.}},
\bauthor{\bsnm{Haffner}, \binits{P.}}:
\batitle{Gradient-based learning applied to document recognition}.
\bjtitle{Proceedings of the IEEE}
\bvolume{86}(\bissue{11}),
\bfpage{2278}--\blpage{2324}
(\byear{1998})
\end{barticle}
\endbibitem

\bibitem[\protect\citeauthoryear{Li et~al.}{2023}]{li2023evaluating}
\begin{botherref}
\oauthor{\bsnm{Li}, \binits{Y.}},
\oauthor{\bsnm{Du}, \binits{Y.}},
\oauthor{\bsnm{Zhou}, \binits{K.}},
\oauthor{\bsnm{Wang}, \binits{J.}},
\oauthor{\bsnm{Zhao}, \binits{W.X.}},
\oauthor{\bsnm{Wen}, \binits{J.-R.}}:
Evaluating object hallucination in large vision-language models.
arXiv preprint arXiv:2305.10355
(2023)
\end{botherref}
\endbibitem

\bibitem[\protect\citeauthoryear{Liu et~al.}{2024}]{liu2024mmbench}
\begin{bchapter}
\bauthor{\bsnm{Liu}, \binits{Y.}},
\bauthor{\bsnm{Duan}, \binits{H.}},
\bauthor{\bsnm{Zhang}, \binits{Y.}},
\bauthor{\bsnm{Li}, \binits{B.}},
\bauthor{\bsnm{Zhang}, \binits{S.}},
\bauthor{\bsnm{Zhao}, \binits{W.}},
\bauthor{\bsnm{Yuan}, \binits{Y.}},
\bauthor{\bsnm{Wang}, \binits{J.}},
\bauthor{\bsnm{He}, \binits{C.}},
\bauthor{\bsnm{Liu}, \binits{Z.}}, \betal:
\bctitle{Mmbench: Is your multi-modal model an all-around player?}
In: \bbtitle{European Conference on Computer Vision},
pp. \bfpage{216}--\blpage{233}
(\byear{2024}).
\bcomment{Springer}
\end{bchapter}
\endbibitem

\bibitem[\protect\citeauthoryear{LeCun}{1998}]{lecun1998mnist}
\begin{botherref}
\oauthor{\bsnm{LeCun}, \binits{Y.}}:
The mnist database of handwritten digits.
http://yann. lecun. com/exdb/mnist/
(1998)
\end{botherref}
\endbibitem

\bibitem[\protect\citeauthoryear{Li et~al.}{2004}]{li2004statistical}
\begin{barticle}
\bauthor{\bsnm{Li}, \binits{L.}},
\bauthor{\bsnm{Huang}, \binits{W.}},
\bauthor{\bsnm{Gu}, \binits{I.Y.-H.}},
\bauthor{\bsnm{Tian}, \binits{Q.}}:
\batitle{Statistical modeling of complex backgrounds for foreground object detection}.
\bjtitle{IEEE Transactions on image processing}
\bvolume{13}(\bissue{11}),
\bfpage{1459}--\blpage{1472}
(\byear{2004})
\end{barticle}
\endbibitem

\bibitem[\protect\citeauthoryear{Liu et~al.}{2020}]{liu2020picanet}
\begin{barticle}
\bauthor{\bsnm{Liu}, \binits{N.}},
\bauthor{\bsnm{Han}, \binits{J.}},
\bauthor{\bsnm{Yang}, \binits{M.-H.}}:
\batitle{Picanet: Pixel-wise contextual attention learning for accurate saliency detection}.
\bjtitle{IEEE Transactions on Image Processing}
\bvolume{29},
\bfpage{6438}--\blpage{6451}
(\byear{2020})
\end{barticle}
\endbibitem

\bibitem[\protect\citeauthoryear{Li et~al.}{2020}]{rs12091435}
\begin{botherref}
\oauthor{\bsnm{Li}, \binits{C.}},
\oauthor{\bsnm{Luo}, \binits{B.}},
\oauthor{\bsnm{Hong}, \binits{H.}},
\oauthor{\bsnm{Su}, \binits{X.}},
\oauthor{\bsnm{Wang}, \binits{Y.}},
\oauthor{\bsnm{Liu}, \binits{J.}},
\oauthor{\bsnm{Wang}, \binits{C.}},
\oauthor{\bsnm{Zhang}, \binits{J.}},
\oauthor{\bsnm{Wei}, \binits{L.}}:
Object detection based on global-local saliency constraint in aerial images.
Remote Sensing
\textbf{12}(9)
(2020)
\doiurl{10.3390/rs12091435}
\end{botherref}
\endbibitem

\bibitem[\protect\citeauthoryear{Li et~al.}{2022}]{li2022elevater}
\begin{botherref}
\oauthor{\bsnm{Li}, \binits{C.}},
\oauthor{\bsnm{Liu}, \binits{H.}},
\oauthor{\bsnm{Li}, \binits{L.H.}},
\oauthor{\bsnm{Zhang}, \binits{P.}},
\oauthor{\bsnm{Aneja}, \binits{J.}},
\oauthor{\bsnm{Yang}, \binits{J.}},
\oauthor{\bsnm{Jin}, \binits{P.}},
\oauthor{\bsnm{Hu}, \binits{H.}},
\oauthor{\bsnm{Liu}, \binits{Z.}},
\oauthor{\bsnm{Lee}, \binits{Y.J.}},
\oauthor{\bsnm{Gao}, \binits{J.}}:
Elevater: A benchmark and toolkit for evaluating language-augmented visual models.
Neural Information Processing Systems
(2022)
\end{botherref}
\endbibitem

\bibitem[\protect\citeauthoryear{Liang et~al.}{2017}]{liang2017enhancing}
\begin{botherref}
\oauthor{\bsnm{Liang}, \binits{S.}},
\oauthor{\bsnm{Li}, \binits{Y.}},
\oauthor{\bsnm{Srikant}, \binits{R.}}:
Enhancing the reliability of out-of-distribution image detection in neural networks.
arXiv preprint arXiv:1706.02690
(2017)
\end{botherref}
\endbibitem

\bibitem[\protect\citeauthoryear{Liu et~al.}{2024}]{liu2024visual}
\begin{botherref}
\oauthor{\bsnm{Liu}, \binits{H.}},
\oauthor{\bsnm{Li}, \binits{C.}},
\oauthor{\bsnm{Wu}, \binits{Q.}},
\oauthor{\bsnm{Lee}, \binits{Y.J.}}:
Visual instruction tuning.
Advances in neural information processing systems
\textbf{36}
(2024)
\end{botherref}
\endbibitem

\bibitem[\protect\citeauthoryear{Li et~al.}{2022}]{li2022blip}
\begin{bchapter}
\bauthor{\bsnm{Li}, \binits{J.}},
\bauthor{\bsnm{Li}, \binits{D.}},
\bauthor{\bsnm{Xiong}, \binits{C.}},
\bauthor{\bsnm{Hoi}, \binits{S.}}:
\bctitle{Blip: Bootstrapping language-image pre-training for unified vision-language understanding and generation}.
In: \bbtitle{International Conference on Machine Learning},
pp. \bfpage{12888}--\blpage{12900}
(\byear{2022}).
\bcomment{PMLR}
\end{bchapter}
\endbibitem

\bibitem[\protect\citeauthoryear{Lin et~al.}{2014}]{lin2014microsoft}
\begin{bchapter}
\bauthor{\bsnm{Lin}, \binits{T.-Y.}},
\bauthor{\bsnm{Maire}, \binits{M.}},
\bauthor{\bsnm{Belongie}, \binits{S.}},
\bauthor{\bsnm{Hays}, \binits{J.}},
\bauthor{\bsnm{Perona}, \binits{P.}},
\bauthor{\bsnm{Ramanan}, \binits{D.}},
\bauthor{\bsnm{Doll{\'a}r}, \binits{P.}},
\bauthor{\bsnm{Zitnick}, \binits{C.L.}}:
\bctitle{Microsoft coco: Common objects in context}.
In: \bbtitle{Computer Vision--ECCV 2014: 13th European Conference, Zurich, Switzerland, September 6-12, 2014, Proceedings, Part V 13},
pp. \bfpage{740}--\blpage{755}
(\byear{2014}).
\bcomment{Springer}
\end{bchapter}
\endbibitem

\bibitem[\protect\citeauthoryear{Lowe}{2004}]{lowe2004distinctive}
\begin{barticle}
\bauthor{\bsnm{Lowe}, \binits{D.G.}}:
\batitle{Distinctive image features from scale-invariant keypoints}.
\bjtitle{International journal of computer vision}
\bvolume{60},
\bfpage{91}--\blpage{110}
(\byear{2004})
\end{barticle}
\endbibitem

\bibitem[\protect\citeauthoryear{Long et~al.}{2015}]{long2015fully}
\begin{bchapter}
\bauthor{\bsnm{Long}, \binits{J.}},
\bauthor{\bsnm{Shelhamer}, \binits{E.}},
\bauthor{\bsnm{Darrell}, \binits{T.}}:
\bctitle{Fully convolutional networks for semantic segmentation}.
In: \bbtitle{Proceedings of the IEEE Conference on Computer Vision and Pattern Recognition},
pp. \bfpage{3431}--\blpage{3440}
(\byear{2015})
\end{bchapter}
\endbibitem

\bibitem[\protect\citeauthoryear{Lo and Velastin}{2001}]{lo2001automatic}
\begin{bchapter}
\bauthor{\bsnm{Lo}, \binits{B.P.L.}},
\bauthor{\bsnm{Velastin}, \binits{S.}}:
\bctitle{Automatic congestion detection system for underground platforms}.
In: \bbtitle{Proceedings of 2001 International Symposium on Intelligent Multimedia, Video and Speech Processing. ISIMP 2001 (IEEE Cat. No. 01EX489)},
pp. \bfpage{158}--\blpage{161}
(\byear{2001}).
\bcomment{IEEE}
\end{bchapter}
\endbibitem

\bibitem[\protect\citeauthoryear{Liu et~al.}{2022}]{liu2022open}
\begin{bchapter}
\bauthor{\bsnm{Liu}, \binits{Q.}},
\bauthor{\bsnm{Wen}, \binits{Y.}},
\bauthor{\bsnm{Han}, \binits{J.}},
\bauthor{\bsnm{Xu}, \binits{C.}},
\bauthor{\bsnm{Xu}, \binits{H.}},
\bauthor{\bsnm{Liang}, \binits{X.}}:
\bctitle{Open-world semantic segmentation via contrasting and clustering vision-language embedding}.
In: \bbtitle{European Conference on Computer Vision},
pp. \bfpage{275}--\blpage{292}
(\byear{2022}).
\bcomment{Springer}
\end{bchapter}
\endbibitem

\bibitem[\protect\citeauthoryear{Liu et~al.}{2020}]{liu2020energy}
\begin{barticle}
\bauthor{\bsnm{Liu}, \binits{W.}},
\bauthor{\bsnm{Wang}, \binits{X.}},
\bauthor{\bsnm{Owens}, \binits{J.}},
\bauthor{\bsnm{Li}, \binits{Y.}}:
\batitle{Energy-based out-of-distribution detection}.
\bjtitle{Advances in neural information processing systems}
\bvolume{33},
\bfpage{21464}--\blpage{21475}
(\byear{2020})
\end{barticle}
\endbibitem

\bibitem[\protect\citeauthoryear{Li et~al.}{2023}]{li2023seed}
\begin{botherref}
\oauthor{\bsnm{Li}, \binits{B.}},
\oauthor{\bsnm{Wang}, \binits{R.}},
\oauthor{\bsnm{Wang}, \binits{G.}},
\oauthor{\bsnm{Ge}, \binits{Y.}},
\oauthor{\bsnm{Ge}, \binits{Y.}},
\oauthor{\bsnm{Shan}, \binits{Y.}}:
Seed-bench: Benchmarking multimodal llms with generative comprehension.
arXiv preprint arXiv:2307.16125
(2023)
\end{botherref}
\endbibitem

\bibitem[\protect\citeauthoryear{Le and Yang}{2015}]{le2015tiny}
\begin{barticle}
\bauthor{\bsnm{Le}, \binits{Y.}},
\bauthor{\bsnm{Yang}, \binits{X.}}:
\batitle{Tiny imagenet visual recognition challenge}.
\bjtitle{CS 231N}
\bvolume{7}(\bissue{7}),
\bfpage{3}
(\byear{2015})
\end{barticle}
\endbibitem

\bibitem[\protect\citeauthoryear{Li et~al.}{2019}]{li2019zero}
\begin{bchapter}
\bauthor{\bsnm{Li}, \binits{Z.}},
\bauthor{\bsnm{Yao}, \binits{L.}},
\bauthor{\bsnm{Zhang}, \binits{X.}},
\bauthor{\bsnm{Wang}, \binits{X.}},
\bauthor{\bsnm{Kanhere}, \binits{S.}},
\bauthor{\bsnm{Zhang}, \binits{H.}}:
\bctitle{Zero-shot object detection with textual descriptions}.
In: \bbtitle{Proceedings of the AAAI Conference on Artificial Intelligence},
vol. \bseriesno{33},
pp. \bfpage{8690}--\blpage{8697}
(\byear{2019})
\end{bchapter}
\endbibitem

\bibitem[\protect\citeauthoryear{Liu et~al.}{2022}]{liu2022generalised}
\begin{botherref}
\oauthor{\bsnm{Liu}, \binits{J.}},
\oauthor{\bsnm{Zhang}, \binits{J.}},
\oauthor{\bsnm{Cui}, \binits{R.}},
\oauthor{\bsnm{Zhang}, \binits{K.}},
\oauthor{\bsnm{Li}, \binits{W.}},
\oauthor{\bsnm{Barnes}, \binits{N.}}:
Generalised co-salient object detection.
arXiv preprint arXiv:2208.09668
(2022)
\end{botherref}
\endbibitem

\bibitem[\protect\citeauthoryear{Liu et~al.}{2025}]{liu2025grounding}
\begin{bchapter}
\bauthor{\bsnm{Liu}, \binits{S.}},
\bauthor{\bsnm{Zeng}, \binits{Z.}},
\bauthor{\bsnm{Ren}, \binits{T.}},
\bauthor{\bsnm{Li}, \binits{F.}},
\bauthor{\bsnm{Zhang}, \binits{H.}},
\bauthor{\bsnm{Yang}, \binits{J.}},
\bauthor{\bsnm{Jiang}, \binits{Q.}},
\bauthor{\bsnm{Li}, \binits{C.}},
\bauthor{\bsnm{Yang}, \binits{J.}},
\bauthor{\bsnm{Su}, \binits{H.}}, \betal:
\bctitle{Grounding dino: Marrying dino with grounded pre-training for open-set object detection}.
In: \bbtitle{European Conference on Computer Vision},
pp. \bfpage{38}--\blpage{55}
(\byear{2025}).
\bcomment{Springer}
\end{bchapter}
\endbibitem

\bibitem[\protect\citeauthoryear{Li et~al.}{2022}]{li2022grounded}
\begin{bchapter}
\bauthor{\bsnm{Li}, \binits{L.H.}},
\bauthor{\bsnm{Zhang}, \binits{P.}},
\bauthor{\bsnm{Zhang}, \binits{H.}},
\bauthor{\bsnm{Yang}, \binits{J.}},
\bauthor{\bsnm{Li}, \binits{C.}},
\bauthor{\bsnm{Zhong}, \binits{Y.}},
\bauthor{\bsnm{Wang}, \binits{L.}},
\bauthor{\bsnm{Yuan}, \binits{L.}},
\bauthor{\bsnm{Zhang}, \binits{L.}},
\bauthor{\bsnm{Hwang}, \binits{J.-N.}}, \betal:
\bctitle{Grounded language-image pre-training}.
In: \bbtitle{Proceedings of the IEEE/CVF Conference on Computer Vision and Pattern Recognition},
pp. \bfpage{10965}--\blpage{10975}
(\byear{2022})
\end{bchapter}
\endbibitem

\bibitem[\protect\citeauthoryear{Mishra et~al.}{2012}]{MishraBMVC12}
\begin{bchapter}
\bauthor{\bsnm{Mishra}, \binits{A.}},
\bauthor{\bsnm{Alahari}, \binits{K.}},
\bauthor{\bsnm{Jawahar}, \binits{C.V.}}:
\bctitle{Scene text recognition using higher order language priors}.
In: \bbtitle{BMVC}
(\byear{2012})
\end{bchapter}
\endbibitem

\bibitem[\protect\citeauthoryear{Marchesotti et~al.}{2009}]{marchesotti2009framework}
\begin{bchapter}
\bauthor{\bsnm{Marchesotti}, \binits{L.}},
\bauthor{\bsnm{Cifarelli}, \binits{C.}},
\bauthor{\bsnm{Csurka}, \binits{G.}}:
\bctitle{A framework for visual saliency detection with applications to image thumbnailing}.
In: \bbtitle{2009 IEEE 12th International Conference on Computer Vision},
pp. \bfpage{2232}--\blpage{2239}
(\byear{2009}).
\bcomment{IEEE}
\end{bchapter}
\endbibitem

\bibitem[\protect\citeauthoryear{Mottaghi et~al.}{2014}]{mottaghi2014role}
\begin{bchapter}
\bauthor{\bsnm{Mottaghi}, \binits{R.}},
\bauthor{\bsnm{Chen}, \binits{X.}},
\bauthor{\bsnm{Liu}, \binits{X.}},
\bauthor{\bsnm{Cho}, \binits{N.-G.}},
\bauthor{\bsnm{Lee}, \binits{S.-W.}},
\bauthor{\bsnm{Fidler}, \binits{S.}},
\bauthor{\bsnm{Urtasun}, \binits{R.}},
\bauthor{\bsnm{Yuille}, \binits{A.}}:
\bctitle{The role of context for object detection and semantic segmentation in the wild}.
In: \bbtitle{Proceedings of the IEEE Conference on Computer Vision and Pattern Recognition},
pp. \bfpage{891}--\blpage{898}
(\byear{2014})
\end{bchapter}
\endbibitem

\bibitem[\protect\citeauthoryear{Martin et~al.}{2001}]{MartinFTM01}
\begin{bchapter}
\bauthor{\bsnm{Martin}, \binits{D.}},
\bauthor{\bsnm{Fowlkes}, \binits{C.}},
\bauthor{\bsnm{Tal}, \binits{D.}},
\bauthor{\bsnm{Malik}, \binits{J.}}:
\bctitle{A database of human segmented natural images and its application to evaluating segmentation algorithms and measuring ecological statistics}.
In: \bbtitle{Proc. 8th Int'l Conf. Computer Vision},
vol. \bseriesno{2},
pp. \bfpage{416}--\blpage{423}
(\byear{2001})
\end{bchapter}
\endbibitem

\bibitem[\protect\citeauthoryear{Mullappilly et~al.}{2024}]{mullappilly2024semi}
\begin{bchapter}
\bauthor{\bsnm{Mullappilly}, \binits{S.S.}},
\bauthor{\bsnm{Gehlot}, \binits{A.S.}},
\bauthor{\bsnm{Anwer}, \binits{R.M.}},
\bauthor{\bsnm{Khan}, \binits{F.S.}},
\bauthor{\bsnm{Cholakkal}, \binits{H.}}:
\bctitle{Semi-supervised open-world object detection}.
In: \bbtitle{Proceedings of the AAAI Conference on Artificial Intelligence},
vol. \bseriesno{38},
pp. \bfpage{4305}--\blpage{4314}
(\byear{2024})
\end{bchapter}
\endbibitem

\bibitem[\protect\citeauthoryear{Minderer et~al.}{2022}]{minderer2022simple}
\begin{bchapter}
\bauthor{\bsnm{Minderer}, \binits{M.}},
\bauthor{\bsnm{Gritsenko}, \binits{A.}},
\bauthor{\bsnm{Stone}, \binits{A.}},
\bauthor{\bsnm{Neumann}, \binits{M.}},
\bauthor{\bsnm{Weissenborn}, \binits{D.}},
\bauthor{\bsnm{Dosovitskiy}, \binits{A.}},
\bauthor{\bsnm{Mahendran}, \binits{A.}},
\bauthor{\bsnm{Arnab}, \binits{A.}},
\bauthor{\bsnm{Dehghani}, \binits{M.}},
\bauthor{\bsnm{Shen}, \binits{Z.}}, \betal:
\bctitle{Simple open-vocabulary object detection}.
In: \bbtitle{European Conference on Computer Vision},
pp. \bfpage{728}--\blpage{755}
(\byear{2022}).
\bcomment{Springer}
\end{bchapter}
\endbibitem

\bibitem[\protect\citeauthoryear{Maji et~al.}{2013}]{maji13fine-grained}
\begin{botherref}
\oauthor{\bsnm{Maji}, \binits{S.}},
\oauthor{\bsnm{Kannala}, \binits{J.}},
\oauthor{\bsnm{Rahtu}, \binits{E.}},
\oauthor{\bsnm{Blaschko}, \binits{M.}},
\oauthor{\bsnm{Vedaldi}, \binits{A.}}:
Fine-grained visual classification of aircraft.
Technical report
(2013)
\end{botherref}
\endbibitem

\bibitem[\protect\citeauthoryear{Madan et~al.}{2024}]{madan2024foundation}
\begin{botherref}
\oauthor{\bsnm{Madan}, \binits{N.}},
\oauthor{\bsnm{M{\o}gelmose}, \binits{A.}},
\oauthor{\bsnm{Modi}, \binits{R.}},
\oauthor{\bsnm{Rawat}, \binits{Y.S.}},
\oauthor{\bsnm{Moeslund}, \binits{T.B.}}:
Foundation models for video understanding: A survey.
arXiv preprint arXiv:2405.03770
(2024)
\end{botherref}
\endbibitem

\bibitem[\protect\citeauthoryear{Ma et~al.}{2023}]{ma2023cat}
\begin{bchapter}
\bauthor{\bsnm{Ma}, \binits{S.}},
\bauthor{\bsnm{Wang}, \binits{Y.}},
\bauthor{\bsnm{Wei}, \binits{Y.}},
\bauthor{\bsnm{Fan}, \binits{J.}},
\bauthor{\bsnm{Li}, \binits{T.H.}},
\bauthor{\bsnm{Liu}, \binits{H.}},
\bauthor{\bsnm{Lv}, \binits{F.}}:
\bctitle{Cat: Localization and identification cascade detection transformer for open-world object detection}.
In: \bbtitle{Proceedings of the IEEE/CVF Conference on Computer Vision and Pattern Recognition},
pp. \bfpage{19681}--\blpage{19690}
(\byear{2023})
\end{bchapter}
\endbibitem

\bibitem[\protect\citeauthoryear{Meng et~al.}{2024}]{meng2024deepstack}
\begin{botherref}
\oauthor{\bsnm{Meng}, \binits{L.}},
\oauthor{\bsnm{Yang}, \binits{J.}},
\oauthor{\bsnm{Tian}, \binits{R.}},
\oauthor{\bsnm{Dai}, \binits{X.}},
\oauthor{\bsnm{Wu}, \binits{Z.}},
\oauthor{\bsnm{Gao}, \binits{J.}},
\oauthor{\bsnm{Jiang}, \binits{Y.-G.}}:
Deepstack: Deeply stacking visual tokens is surprisingly simple and effective for lmms.
arXiv preprint arXiv:2406.04334
(2024)
\end{botherref}
\endbibitem

\bibitem[\protect\citeauthoryear{Netzer et~al.}{2011}]{netzer2011reading}
\begin{bchapter}
\bauthor{\bsnm{Netzer}, \binits{Y.}},
\bauthor{\bsnm{Wang}, \binits{T.}},
\bauthor{\bsnm{Coates}, \binits{A.}},
\bauthor{\bsnm{Bissacco}, \binits{A.}},
\bauthor{\bsnm{Wu}, \binits{B.}},
\bauthor{\bsnm{Ng}, \binits{A.Y.}}, \betal:
\bctitle{Reading digits in natural images with unsupervised feature learning}.
In: \bbtitle{NIPS Workshop on Deep Learning and Unsupervised Feature Learning},
vol. \bseriesno{2011},
p. \bfpage{4}
(\byear{2011}).
\bcomment{Granada}
\end{bchapter}
\endbibitem

\bibitem[\protect\citeauthoryear{Nilsback and Zisserman}{2008}]{Nilsback08}
\begin{bchapter}
\bauthor{\bsnm{Nilsback}, \binits{M.-E.}},
\bauthor{\bsnm{Zisserman}, \binits{A.}}:
\bctitle{Automated flower classification over a large number of classes}.
In: \bbtitle{Indian Conference on Computer Vision, Graphics and Image Processing}
(\byear{2008})
\end{bchapter}
\endbibitem

\bibitem[\protect\citeauthoryear{Oquab et~al.}{2023}]{oquab2023dinov2}
\begin{botherref}
\oauthor{\bsnm{Oquab}, \binits{M.}},
\oauthor{\bsnm{Darcet}, \binits{T.}},
\oauthor{\bsnm{Moutakanni}, \binits{T.}},
\oauthor{\bsnm{Vo}, \binits{H.}},
\oauthor{\bsnm{Szafraniec}, \binits{M.}},
\oauthor{\bsnm{Khalidov}, \binits{V.}},
\oauthor{\bsnm{Fernandez}, \binits{P.}},
\oauthor{\bsnm{Haziza}, \binits{D.}},
\oauthor{\bsnm{Massa}, \binits{F.}},
\oauthor{\bsnm{El-Nouby}, \binits{A.}}, et al.:
Dinov2: Learning robust visual features without supervision.
arXiv preprint arXiv:2304.07193
(2023)
\end{botherref}
\endbibitem

\bibitem[\protect\citeauthoryear{Ordonez et~al.}{2011}]{ordonez2011im2text}
\begin{botherref}
\oauthor{\bsnm{Ordonez}, \binits{V.}},
\oauthor{\bsnm{Kulkarni}, \binits{G.}},
\oauthor{\bsnm{Berg}, \binits{T.}}:
Im2text: Describing images using 1 million captioned photographs.
Advances in neural information processing systems
\textbf{24}
(2011)
\end{botherref}
\endbibitem

\bibitem[\protect\citeauthoryear{Oliver et~al.}{2000}]{oliver2000bayesian}
\begin{barticle}
\bauthor{\bsnm{Oliver}, \binits{N.M.}},
\bauthor{\bsnm{Rosario}, \binits{B.}},
\bauthor{\bsnm{Pentland}, \binits{A.P.}}:
\batitle{A bayesian computer vision system for modeling human interactions}.
\bjtitle{IEEE transactions on pattern analysis and machine intelligence}
\bvolume{22}(\bissue{8}),
\bfpage{831}--\blpage{843}
(\byear{2000})
\end{barticle}
\endbibitem

\bibitem[\protect\citeauthoryear{Otsu}{1979}]{4310076}
\begin{barticle}
\bauthor{\bsnm{Otsu}, \binits{N.}}:
\batitle{A threshold selection method from gray-level histograms}.
\bjtitle{IEEE Transactions on Systems, Man, and Cybernetics}
\bvolume{9}(\bissue{1}),
\bfpage{62}--\blpage{66}
(\byear{1979})
\doiurl{10.1109/TSMC.1979.4310076}
\end{barticle}
\endbibitem

\bibitem[\protect\citeauthoryear{Parkhi et~al.}{2012}]{parkhi2012cats}
\begin{bchapter}
\bauthor{\bsnm{Parkhi}, \binits{O.M.}},
\bauthor{\bsnm{Vedaldi}, \binits{A.}},
\bauthor{\bsnm{Zisserman}, \binits{A.}},
\bauthor{\bsnm{Jawahar}, \binits{C.}}:
\bctitle{Cats and dogs}.
In: \bbtitle{2012 IEEE Conference on Computer Vision and Pattern Recognition},
pp. \bfpage{3498}--\blpage{3505}
(\byear{2012}).
\bcomment{IEEE}
\end{bchapter}
\endbibitem

\bibitem[\protect\citeauthoryear{Plummer et~al.}{2015}]{plummer2015flickr30k}
\begin{bchapter}
\bauthor{\bsnm{Plummer}, \binits{B.A.}},
\bauthor{\bsnm{Wang}, \binits{L.}},
\bauthor{\bsnm{Cervantes}, \binits{C.M.}},
\bauthor{\bsnm{Caicedo}, \binits{J.C.}},
\bauthor{\bsnm{Hockenmaier}, \binits{J.}},
\bauthor{\bsnm{Lazebnik}, \binits{S.}}:
\bctitle{Flickr30k entities: Collecting region-to-phrase correspondences for richer image-to-sentence models}.
In: \bbtitle{Proceedings of the IEEE International Conference on Computer Vision},
pp. \bfpage{2641}--\blpage{2649}
(\byear{2015})
\end{bchapter}
\endbibitem

\bibitem[\protect\citeauthoryear{Pang et~al.}{2020}]{pang2020multi}
\begin{bchapter}
\bauthor{\bsnm{Pang}, \binits{Y.}},
\bauthor{\bsnm{Zhao}, \binits{X.}},
\bauthor{\bsnm{Zhang}, \binits{L.}},
\bauthor{\bsnm{Lu}, \binits{H.}}:
\bctitle{Multi-scale interactive network for salient object detection}.
In: \bbtitle{Proceedings of the IEEE/CVF Conference on Computer Vision and Pattern Recognition},
pp. \bfpage{9413}--\blpage{9422}
(\byear{2020})
\end{bchapter}
\endbibitem

\bibitem[\protect\citeauthoryear{Qian et~al.}{2024}]{qian2024streaming}
\begin{botherref}
\oauthor{\bsnm{Qian}, \binits{R.}},
\oauthor{\bsnm{Dong}, \binits{X.}},
\oauthor{\bsnm{Zhang}, \binits{P.}},
\oauthor{\bsnm{Zang}, \binits{Y.}},
\oauthor{\bsnm{Ding}, \binits{S.}},
\oauthor{\bsnm{Lin}, \binits{D.}},
\oauthor{\bsnm{Wang}, \binits{J.}}:
Streaming long video understanding with large language models.
arXiv preprint arXiv:2405.16009
(2024)
\end{botherref}
\endbibitem

\bibitem[\protect\citeauthoryear{Qin et~al.}{2019}]{qin2019basnet}
\begin{bchapter}
\bauthor{\bsnm{Qin}, \binits{X.}},
\bauthor{\bsnm{Zhang}, \binits{Z.}},
\bauthor{\bsnm{Huang}, \binits{C.}},
\bauthor{\bsnm{Gao}, \binits{C.}},
\bauthor{\bsnm{Dehghan}, \binits{M.}},
\bauthor{\bsnm{Jagersand}, \binits{M.}}:
\bctitle{Basnet: Boundary-aware salient object detection}.
In: \bbtitle{Proceedings of the IEEE/CVF Conference on Computer Vision and Pattern Recognition},
pp. \bfpage{7479}--\blpage{7489}
(\byear{2019})
\end{bchapter}
\endbibitem

\bibitem[\protect\citeauthoryear{Ravindran and Basu}{2023}]{ravindran2023sempart}
\begin{bchapter}
\bauthor{\bsnm{Ravindran}, \binits{S.}},
\bauthor{\bsnm{Basu}, \binits{D.}}:
\bctitle{Sempart: Self-supervised multi-resolution partitioning of image semantics}.
In: \bbtitle{Proceedings of the IEEE/CVF International Conference on Computer Vision},
pp. \bfpage{723}--\blpage{733}
(\byear{2023})
\end{bchapter}
\endbibitem

\bibitem[\protect\citeauthoryear{Redmon}{2016}]{redmon2016you}
\begin{bchapter}
\bauthor{\bsnm{Redmon}, \binits{J.}}:
\bctitle{You only look once: Unified, real-time object detection}.
In: \bbtitle{Proceedings of the IEEE Conference on Computer Vision and Pattern Recognition}
(\byear{2016})
\end{bchapter}
\endbibitem

\bibitem[\protect\citeauthoryear{Ronneberger et~al.}{2015}]{ronneberger2015u}
\begin{bchapter}
\bauthor{\bsnm{Ronneberger}, \binits{O.}},
\bauthor{\bsnm{Fischer}, \binits{P.}},
\bauthor{\bsnm{Brox}, \binits{T.}}:
\bctitle{U-net: Convolutional networks for biomedical image segmentation}.
In: \bbtitle{Medical Image Computing and Computer-assisted intervention--MICCAI 2015: 18th International Conference, Munich, Germany, October 5-9, 2015, Proceedings, Part III 18},
pp. \bfpage{234}--\blpage{241}
(\byear{2015}).
\bcomment{Springer}
\end{bchapter}
\endbibitem

\bibitem[\protect\citeauthoryear{Ravi et~al.}{2024}]{ravi2024sam}
\begin{botherref}
\oauthor{\bsnm{Ravi}, \binits{N.}},
\oauthor{\bsnm{Gabeur}, \binits{V.}},
\oauthor{\bsnm{Hu}, \binits{Y.-T.}},
\oauthor{\bsnm{Hu}, \binits{R.}},
\oauthor{\bsnm{Ryali}, \binits{C.}},
\oauthor{\bsnm{Ma}, \binits{T.}},
\oauthor{\bsnm{Khedr}, \binits{H.}},
\oauthor{\bsnm{R{\"a}dle}, \binits{R.}},
\oauthor{\bsnm{Rolland}, \binits{C.}},
\oauthor{\bsnm{Gustafson}, \binits{L.}}, et al.:
Sam 2: Segment anything in images and videos.
arXiv preprint arXiv:2408.00714
(2024)
\end{botherref}
\endbibitem

\bibitem[\protect\citeauthoryear{Ren et~al.}{2016}]{ren2016faster}
\begin{barticle}
\bauthor{\bsnm{Ren}, \binits{S.}},
\bauthor{\bsnm{He}, \binits{K.}},
\bauthor{\bsnm{Girshick}, \binits{R.}},
\bauthor{\bsnm{Sun}, \binits{J.}}:
\batitle{Faster r-cnn: Towards real-time object detection with region proposal networks}.
\bjtitle{IEEE transactions on pattern analysis and machine intelligence}
\bvolume{39}(\bissue{6}),
\bfpage{1137}--\blpage{1149}
(\byear{2016})
\end{barticle}
\endbibitem

\bibitem[\protect\citeauthoryear{Rumelhart et~al.}{1986}]{rumelhart1986learning}
\begin{barticle}
\bauthor{\bsnm{Rumelhart}, \binits{D.E.}},
\bauthor{\bsnm{Hinton}, \binits{G.E.}},
\bauthor{\bsnm{Williams}, \binits{R.J.}}:
\batitle{Learning internal representations by error propagation, parallel distributed processing, explorations in the microstructure of cognition, ed. de rumelhart and j. mcclelland. vol. 1. 1986}.
\bjtitle{Biometrika}
\bvolume{71}(\bissue{599-607}),
\bfpage{6}
(\byear{1986})
\end{barticle}
\endbibitem

\bibitem[\protect\citeauthoryear{Radford et~al.}{2021}]{radford2021learning}
\begin{bchapter}
\bauthor{\bsnm{Radford}, \binits{A.}},
\bauthor{\bsnm{Kim}, \binits{J.W.}},
\bauthor{\bsnm{Hallacy}, \binits{C.}},
\bauthor{\bsnm{Ramesh}, \binits{A.}},
\bauthor{\bsnm{Goh}, \binits{G.}},
\bauthor{\bsnm{Agarwal}, \binits{S.}},
\bauthor{\bsnm{Sastry}, \binits{G.}},
\bauthor{\bsnm{Askell}, \binits{A.}},
\bauthor{\bsnm{Mishkin}, \binits{P.}},
\bauthor{\bsnm{Clark}, \binits{J.}}, \betal:
\bctitle{Learning transferable visual models from natural language supervision}.
In: \bbtitle{International Conference on Machine Learning},
pp. \bfpage{8748}--\blpage{8763}
(\byear{2021}).
\bcomment{PMLR}
\end{bchapter}
\endbibitem

\bibitem[\protect\citeauthoryear{Rahman et~al.}{2018}]{rahman2018zero}
\begin{bchapter}
\bauthor{\bsnm{Rahman}, \binits{S.}},
\bauthor{\bsnm{Khan}, \binits{S.}},
\bauthor{\bsnm{Porikli}, \binits{F.}}:
\bctitle{Zero-shot object detection: Learning to simultaneously recognize and localize novel concepts}.
In: \bbtitle{Asian Conference on Computer Vision},
pp. \bfpage{547}--\blpage{563}
(\byear{2018}).
\bcomment{Springer}
\end{bchapter}
\endbibitem

\bibitem[\protect\citeauthoryear{Rosin}{1995}]{Rosin1995EdgesSM}
\begin{barticle}
\bauthor{\bsnm{Rosin}, \binits{P.L.}}:
\batitle{Edges: saliency measures and automatic thresholding}.
\bjtitle{Machine Vision and Applications}
\bvolume{9},
\bfpage{139}--\blpage{159}
(\byear{1995})
\end{barticle}
\endbibitem

\bibitem[\protect\citeauthoryear{Schuhmann et~al.}{2022}]{schuhmann2022laion}
\begin{barticle}
\bauthor{\bsnm{Schuhmann}, \binits{C.}},
\bauthor{\bsnm{Beaumont}, \binits{R.}},
\bauthor{\bsnm{Vencu}, \binits{R.}},
\bauthor{\bsnm{Gordon}, \binits{C.}},
\bauthor{\bsnm{Wightman}, \binits{R.}},
\bauthor{\bsnm{Cherti}, \binits{M.}},
\bauthor{\bsnm{Coombes}, \binits{T.}},
\bauthor{\bsnm{Katta}, \binits{A.}},
\bauthor{\bsnm{Mullis}, \binits{C.}},
\bauthor{\bsnm{Wortsman}, \binits{M.}}, \betal:
\batitle{Laion-5b: An open large-scale dataset for training next generation image-text models}.
\bjtitle{Advances in Neural Information Processing Systems}
\bvolume{35},
\bfpage{25278}--\blpage{25294}
(\byear{2022})
\end{barticle}
\endbibitem

\bibitem[\protect\citeauthoryear{Shang et~al.}{2024}]{shang2024llava}
\begin{botherref}
\oauthor{\bsnm{Shang}, \binits{Y.}},
\oauthor{\bsnm{Cai}, \binits{M.}},
\oauthor{\bsnm{Xu}, \binits{B.}},
\oauthor{\bsnm{Lee}, \binits{Y.J.}},
\oauthor{\bsnm{Yan}, \binits{Y.}}:
Llava-prumerge: Adaptive token reduction for efficient large multimodal models.
arXiv preprint arXiv:2403.15388
(2024)
\end{botherref}
\endbibitem

\bibitem[\protect\citeauthoryear{Sharma et~al.}{2018}]{sharma2018conceptual}
\begin{bchapter}
\bauthor{\bsnm{Sharma}, \binits{P.}},
\bauthor{\bsnm{Ding}, \binits{N.}},
\bauthor{\bsnm{Goodman}, \binits{S.}},
\bauthor{\bsnm{Soricut}, \binits{R.}}:
\bctitle{Conceptual captions: A cleaned, hypernymed, image alt-text dataset for automatic image captioning}.
In: \bbtitle{Proceedings of the 56th Annual Meeting of the Association for Computational Linguistics (Volume 1: Long Papers)},
pp. \bfpage{2556}--\blpage{2565}
(\byear{2018})
\end{bchapter}
\endbibitem

\bibitem[\protect\citeauthoryear{Sun et~al.}{2023}]{sun2023eva}
\begin{botherref}
\oauthor{\bsnm{Sun}, \binits{Q.}},
\oauthor{\bsnm{Fang}, \binits{Y.}},
\oauthor{\bsnm{Wu}, \binits{L.}},
\oauthor{\bsnm{Wang}, \binits{X.}},
\oauthor{\bsnm{Cao}, \binits{Y.}}:
Eva-clip: Improved training techniques for clip at scale.
arXiv preprint arXiv:2303.15389
(2023)
\end{botherref}
\endbibitem

\bibitem[\protect\citeauthoryear{Stauffer and Grimson}{1999}]{stauffer1999adaptive}
\begin{bchapter}
\bauthor{\bsnm{Stauffer}, \binits{C.}},
\bauthor{\bsnm{Grimson}, \binits{W.E.L.}}:
\bctitle{Adaptive background mixture models for real-time tracking}.
In: \bbtitle{Proceedings. 1999 IEEE Computer Society Conference on Computer Vision and Pattern Recognition (Cat. No PR00149)},
vol. \bseriesno{2},
pp. \bfpage{246}--\blpage{252}
(\byear{1999}).
\bcomment{IEEE}
\end{bchapter}
\endbibitem

\bibitem[\protect\citeauthoryear{Shaheen et~al.}{2023}]{shaheen2023framework}
\begin{botherref}
\oauthor{\bsnm{Shaheen}, \binits{K.}},
\oauthor{\bsnm{Hanif}, \binits{M.A.}},
\oauthor{\bsnm{Hasan}, \binits{O.}},
\oauthor{\bsnm{Shafique}, \binits{M.}}:
A framework for open world object detection.
Artificial Intelligence Evolution,
154--164
(2023)
\end{botherref}
\endbibitem

\bibitem[\protect\citeauthoryear{Song et~al.}{2015}]{song2015sun}
\begin{bchapter}
\bauthor{\bsnm{Song}, \binits{S.}},
\bauthor{\bsnm{Lichtenberg}, \binits{S.P.}},
\bauthor{\bsnm{Xiao}, \binits{J.}}:
\bctitle{Sun rgb-d: A rgb-d scene understanding benchmark suite}.
In: \bbtitle{Proceedings of the IEEE Conference on Computer Vision and Pattern Recognition},
pp. \bfpage{567}--\blpage{576}
(\byear{2015})
\end{bchapter}
\endbibitem

\bibitem[\protect\citeauthoryear{Shao et~al.}{2019}]{shao2019objects365}
\begin{bchapter}
\bauthor{\bsnm{Shao}, \binits{S.}},
\bauthor{\bsnm{Li}, \binits{Z.}},
\bauthor{\bsnm{Zhang}, \binits{T.}},
\bauthor{\bsnm{Peng}, \binits{C.}},
\bauthor{\bsnm{Yu}, \binits{G.}},
\bauthor{\bsnm{Zhang}, \binits{X.}},
\bauthor{\bsnm{Li}, \binits{J.}},
\bauthor{\bsnm{Sun}, \binits{J.}}:
\bctitle{Objects365: A large-scale, high-quality dataset for object detection}.
In: \bbtitle{Proceedings of the IEEE/CVF International Conference on Computer Vision},
pp. \bfpage{8430}--\blpage{8439}
(\byear{2019})
\end{bchapter}
\endbibitem

\bibitem[\protect\citeauthoryear{Sodano et~al.}{2024}]{sodano2024open}
\begin{bchapter}
\bauthor{\bsnm{Sodano}, \binits{M.}},
\bauthor{\bsnm{Magistri}, \binits{F.}},
\bauthor{\bsnm{Nunes}, \binits{L.}},
\bauthor{\bsnm{Behley}, \binits{J.}},
\bauthor{\bsnm{Stachniss}, \binits{C.}}:
\bctitle{Open-world semantic segmentation including class similarity}.
In: \bbtitle{Proceedings of the IEEE/CVF Conference on Computer Vision and Pattern Recognition},
pp. \bfpage{3184}--\blpage{3194}
(\byear{2024})
\end{bchapter}
\endbibitem

\bibitem[\protect\citeauthoryear{Sun et~al.}{2022}]{sun2022out}
\begin{bchapter}
\bauthor{\bsnm{Sun}, \binits{Y.}},
\bauthor{\bsnm{Ming}, \binits{Y.}},
\bauthor{\bsnm{Zhu}, \binits{X.}},
\bauthor{\bsnm{Li}, \binits{Y.}}:
\bctitle{Out-of-distribution detection with deep nearest neighbors}.
In: \bbtitle{International Conference on Machine Learning},
pp. \bfpage{20827}--\blpage{20840}
(\byear{2022}).
\bcomment{PMLR}
\end{bchapter}
\endbibitem

\bibitem[\protect\citeauthoryear{Sch{\"o}lkopf et~al.}{2001}]{scholkopf2001estimating}
\begin{barticle}
\bauthor{\bsnm{Sch{\"o}lkopf}, \binits{B.}},
\bauthor{\bsnm{Platt}, \binits{J.C.}},
\bauthor{\bsnm{Shawe-Taylor}, \binits{J.}},
\bauthor{\bsnm{Smola}, \binits{A.J.}},
\bauthor{\bsnm{Williamson}, \binits{R.C.}}:
\batitle{Estimating the support of a high-dimensional distribution}.
\bjtitle{Neural computation}
\bvolume{13}(\bissue{7}),
\bfpage{1443}--\blpage{1471}
(\byear{2001})
\end{barticle}
\endbibitem

\bibitem[\protect\citeauthoryear{Steiner et~al.}{2024}]{steiner2024paligemma}
\begin{botherref}
\oauthor{\bsnm{Steiner}, \binits{A.}},
\oauthor{\bsnm{Pinto}, \binits{A.S.}},
\oauthor{\bsnm{Tschannen}, \binits{M.}},
\oauthor{\bsnm{Keysers}, \binits{D.}},
\oauthor{\bsnm{Wang}, \binits{X.}},
\oauthor{\bsnm{Bitton}, \binits{Y.}},
\oauthor{\bsnm{Gritsenko}, \binits{A.}},
\oauthor{\bsnm{Minderer}, \binits{M.}},
\oauthor{\bsnm{Sherbondy}, \binits{A.}},
\oauthor{\bsnm{Long}, \binits{S.}}, et al.:
Paligemma 2: A family of versatile vlms for transfer.
arXiv preprint arXiv:2412.03555
(2024)
\end{botherref}
\endbibitem

\bibitem[\protect\citeauthoryear{Socher et~al.}{2013}]{socher2013recursive}
\begin{bchapter}
\bauthor{\bsnm{Socher}, \binits{R.}},
\bauthor{\bsnm{Perelygin}, \binits{A.}},
\bauthor{\bsnm{Wu}, \binits{J.}},
\bauthor{\bsnm{Chuang}, \binits{J.}},
\bauthor{\bsnm{Manning}, \binits{C.D.}},
\bauthor{\bsnm{Ng}, \binits{A.Y.}},
\bauthor{\bsnm{Potts}, \binits{C.}}:
\bctitle{Recursive deep models for semantic compositionality over a sentiment treebank}.
In: \bbtitle{Proceedings of the 2013 Conference on Empirical Methods in Natural Language Processing},
pp. \bfpage{1631}--\blpage{1642}
(\byear{2013})
\end{bchapter}
\endbibitem

\bibitem[\protect\citeauthoryear{Srinivasan et~al.}{2021}]{srinivasan2021wit}
\begin{bchapter}
\bauthor{\bsnm{Srinivasan}, \binits{K.}},
\bauthor{\bsnm{Raman}, \binits{K.}},
\bauthor{\bsnm{Chen}, \binits{J.}},
\bauthor{\bsnm{Bendersky}, \binits{M.}},
\bauthor{\bsnm{Najork}, \binits{M.}}:
\bctitle{Wit: Wikipedia-based image text dataset for multimodal multilingual machine learning}.
In: \bbtitle{Proceedings of the 44th International ACM SIGIR Conference on Research and Development in Information Retrieval},
pp. \bfpage{2443}--\blpage{2449}
(\byear{2021})
\end{bchapter}
\endbibitem

\bibitem[\protect\citeauthoryear{Sim{\'e}oni et~al.}{2023}]{simeoni2023unsupervised}
\begin{bchapter}
\bauthor{\bsnm{Sim{\'e}oni}, \binits{O.}},
\bauthor{\bsnm{Sekkat}, \binits{C.}},
\bauthor{\bsnm{Puy}, \binits{G.}},
\bauthor{\bsnm{Vobeck{\`y}}, \binits{A.}},
\bauthor{\bsnm{Zablocki}, \binits{{\'E}.}},
\bauthor{\bsnm{P{\'e}rez}, \binits{P.}}:
\bctitle{Unsupervised object localization: Observing the background to discover objects}.
In: \bbtitle{Proceedings of the IEEE/CVF Conference on Computer Vision and Pattern Recognition},
pp. \bfpage{3176}--\blpage{3186}
(\byear{2023})
\end{bchapter}
\endbibitem

\bibitem[\protect\citeauthoryear{Stallkamp et~al.}{2012}]{stallkamp2012man}
\begin{barticle}
\bauthor{\bsnm{Stallkamp}, \binits{J.}},
\bauthor{\bsnm{Schlipsing}, \binits{M.}},
\bauthor{\bsnm{Salmen}, \binits{J.}},
\bauthor{\bsnm{Igel}, \binits{C.}}:
\batitle{Man vs. computer: Benchmarking machine learning algorithms for traffic sign recognition}.
\bjtitle{Neural networks}
\bvolume{32},
\bfpage{323}--\blpage{332}
(\byear{2012})
\end{barticle}
\endbibitem

\bibitem[\protect\citeauthoryear{Sobral and Vacavant}{2014}]{sobral2014comprehensive}
\begin{barticle}
\bauthor{\bsnm{Sobral}, \binits{A.}},
\bauthor{\bsnm{Vacavant}, \binits{A.}}:
\batitle{A comprehensive review of background subtraction algorithms evaluated with synthetic and real videos}.
\bjtitle{Computer Vision and Image Understanding}
\bvolume{122},
\bfpage{4}--\blpage{21}
(\byear{2014})
\end{barticle}
\endbibitem

\bibitem[\protect\citeauthoryear{Schuhmann et~al.}{2021}]{schuhmann2021laion}
\begin{botherref}
\oauthor{\bsnm{Schuhmann}, \binits{C.}},
\oauthor{\bsnm{Vencu}, \binits{R.}},
\oauthor{\bsnm{Beaumont}, \binits{R.}},
\oauthor{\bsnm{Kaczmarczyk}, \binits{R.}},
\oauthor{\bsnm{Mullis}, \binits{C.}},
\oauthor{\bsnm{Katta}, \binits{A.}},
\oauthor{\bsnm{Coombes}, \binits{T.}},
\oauthor{\bsnm{Jitsev}, \binits{J.}},
\oauthor{\bsnm{Komatsuzaki}, \binits{A.}}:
Laion-400m: Open dataset of clip-filtered 400 million image-text pairs.
arXiv preprint arXiv:2111.02114
(2021)
\end{botherref}
\endbibitem

\bibitem[\protect\citeauthoryear{Shi et~al.}{2024}]{shi2024we}
\begin{bchapter}
\bauthor{\bsnm{Shi}, \binits{B.}},
\bauthor{\bsnm{Wu}, \binits{Z.}},
\bauthor{\bsnm{Mao}, \binits{M.}},
\bauthor{\bsnm{Wang}, \binits{X.}},
\bauthor{\bsnm{Darrell}, \binits{T.}}:
\bctitle{When do we not need larger vision models?}
In: \bbtitle{European Conference on Computer Vision},
pp. \bfpage{444}--\blpage{462}
(\byear{2024}).
\bcomment{Springer}
\end{bchapter}
\endbibitem

\bibitem[\protect\citeauthoryear{Shi et~al.}{2015}]{shi2015hierarchical}
\begin{barticle}
\bauthor{\bsnm{Shi}, \binits{J.}},
\bauthor{\bsnm{Yan}, \binits{Q.}},
\bauthor{\bsnm{Xu}, \binits{L.}},
\bauthor{\bsnm{Jia}, \binits{J.}}:
\batitle{Hierarchical image saliency detection on extended cssd}.
\bjtitle{IEEE transactions on pattern analysis and machine intelligence}
\bvolume{38}(\bissue{4}),
\bfpage{717}--\blpage{729}
(\byear{2015})
\end{barticle}
\endbibitem

\bibitem[\protect\citeauthoryear{Simonyan and Zisserman}{2014}]{simonyan2014very}
\begin{botherref}
\oauthor{\bsnm{Simonyan}, \binits{K.}},
\oauthor{\bsnm{Zisserman}, \binits{A.}}:
Very deep convolutional networks for large-scale image recognition.
arXiv preprint arXiv:1409.1556
(2014)
\end{botherref}
\endbibitem

\bibitem[\protect\citeauthoryear{Treisman and Gelade}{1980}]{TREISMAN198097}
\begin{barticle}
\bauthor{\bsnm{Treisman}, \binits{A.M.}},
\bauthor{\bsnm{Gelade}, \binits{G.}}:
\batitle{A feature-integration theory of attention}.
\bjtitle{Cognitive Psychology}
\bvolume{12}(\bissue{1}),
\bfpage{97}--\blpage{136}
(\byear{1980})
\doiurl{10.1016/0010-0285(80)90005-5}
\end{barticle}
\endbibitem

\bibitem[\protect\citeauthoryear{Touvron et~al.}{2023}]{touvron2023llama}
\begin{botherref}
\oauthor{\bsnm{Touvron}, \binits{H.}},
\oauthor{\bsnm{Lavril}, \binits{T.}},
\oauthor{\bsnm{Izacard}, \binits{G.}},
\oauthor{\bsnm{Martinet}, \binits{X.}},
\oauthor{\bsnm{Lachaux}, \binits{M.-A.}},
\oauthor{\bsnm{Lacroix}, \binits{T.}},
\oauthor{\bsnm{Rozi{\`e}re}, \binits{B.}},
\oauthor{\bsnm{Goyal}, \binits{N.}},
\oauthor{\bsnm{Hambro}, \binits{E.}},
\oauthor{\bsnm{Azhar}, \binits{F.}}, et al.:
Llama: Open and efficient foundation language models.
arXiv preprint arXiv:2302.13971
(2023)
\end{botherref}
\endbibitem

\bibitem[\protect\citeauthoryear{Thomee et~al.}{2016}]{thomee2016yfcc100m}
\begin{barticle}
\bauthor{\bsnm{Thomee}, \binits{B.}},
\bauthor{\bsnm{Shamma}, \binits{D.A.}},
\bauthor{\bsnm{Friedland}, \binits{G.}},
\bauthor{\bsnm{Elizalde}, \binits{B.}},
\bauthor{\bsnm{Ni}, \binits{K.}},
\bauthor{\bsnm{Poland}, \binits{D.}},
\bauthor{\bsnm{Borth}, \binits{D.}},
\bauthor{\bsnm{Li}, \binits{L.-J.}}:
\batitle{Yfcc100m: The new data in multimedia research}.
\bjtitle{Communications of the ACM}
\bvolume{59}(\bissue{2}),
\bfpage{64}--\blpage{73}
(\byear{2016})
\end{barticle}
\endbibitem

\bibitem[\protect\citeauthoryear{Vaswani}{2017}]{vaswani2017attention}
\begin{botherref}
\oauthor{\bsnm{Vaswani}, \binits{A.}}:
Attention is all you need.
Advances in Neural Information Processing Systems
(2017)
\end{botherref}
\endbibitem

\bibitem[\protect\citeauthoryear{Vacavant et~al.}{2013}]{vacavant2013benchmark}
\begin{bchapter}
\bauthor{\bsnm{Vacavant}, \binits{A.}},
\bauthor{\bsnm{Chateau}, \binits{T.}},
\bauthor{\bsnm{Wilhelm}, \binits{A.}},
\bauthor{\bsnm{Lequievre}, \binits{L.}}:
\bctitle{A benchmark dataset for outdoor foreground/background extraction}.
In: \bbtitle{Computer Vision-ACCV 2012 Workshops: ACCV 2012 International Workshops, Daejeon, Korea, November 5-6, 2012, Revised Selected Papers, Part I 11},
pp. \bfpage{291}--\blpage{300}
(\byear{2013}).
\bcomment{Springer}
\end{bchapter}
\endbibitem

\bibitem[\protect\citeauthoryear{Viola and Jones}{2001}]{viola2001rapid}
\begin{bchapter}
\bauthor{\bsnm{Viola}, \binits{P.}},
\bauthor{\bsnm{Jones}, \binits{M.}}:
\bctitle{Rapid object detection using a boosted cascade of simple features}.
In: \bbtitle{Proceedings of the 2001 IEEE Computer Society Conference on Computer Vision and Pattern Recognition. CVPR 2001},
vol. \bseriesno{1},
p.
(\byear{2001}).
\bcomment{Ieee}
\end{bchapter}
\endbibitem

\bibitem[\protect\citeauthoryear{Veeling et~al.}{2018}]{veeling2018rotation}
\begin{bchapter}
\bauthor{\bsnm{Veeling}, \binits{B.S.}},
\bauthor{\bsnm{Linmans}, \binits{J.}},
\bauthor{\bsnm{Winkens}, \binits{J.}},
\bauthor{\bsnm{Cohen}, \binits{T.}},
\bauthor{\bsnm{Welling}, \binits{M.}}:
\bctitle{Rotation equivariant cnns for digital pathology}.
In: \bbtitle{Medical Image Computing and Computer Assisted Intervention--MICCAI 2018: 21st International Conference, Granada, Spain, September 16-20, 2018, Proceedings, Part II 11},
pp. \bfpage{210}--\blpage{218}
(\byear{2018}).
\bcomment{Springer}
\end{bchapter}
\endbibitem

\bibitem[\protect\citeauthoryear{Voynov et~al.}{2021}]{voynov2021object}
\begin{bchapter}
\bauthor{\bsnm{Voynov}, \binits{A.}},
\bauthor{\bsnm{Morozov}, \binits{S.}},
\bauthor{\bsnm{Babenko}, \binits{A.}}:
\bctitle{Object segmentation without labels with large-scale generative models}.
In: \bbtitle{International Conference on Machine Learning},
pp. \bfpage{10596}--\blpage{10606}
(\byear{2021}).
\bcomment{PMLR}
\end{bchapter}
\endbibitem

\bibitem[\protect\citeauthoryear{Vo et~al.}{2020}]{vo2020toward}
\begin{bchapter}
\bauthor{\bsnm{Vo}, \binits{H.V.}},
\bauthor{\bsnm{P{\'e}rez}, \binits{P.}},
\bauthor{\bsnm{Ponce}, \binits{J.}}:
\bctitle{Toward unsupervised, multi-object discovery in large-scale image collections}.
In: \bbtitle{Computer Vision--ECCV 2020: 16th European Conference, Glasgow, UK, August 23--28, 2020, Proceedings, Part XXIII 16},
pp. \bfpage{779}--\blpage{795}
(\byear{2020}).
\bcomment{Springer}
\end{bchapter}
\endbibitem

\bibitem[\protect\citeauthoryear{Vinyals et~al.}{2015}]{vinyals2015show}
\begin{bchapter}
\bauthor{\bsnm{Vinyals}, \binits{O.}},
\bauthor{\bsnm{Toshev}, \binits{A.}},
\bauthor{\bsnm{Bengio}, \binits{S.}},
\bauthor{\bsnm{Erhan}, \binits{D.}}:
\bctitle{Show and tell: A neural image caption generator}.
In: \bbtitle{Proceedings of the IEEE Conference on Computer Vision and Pattern Recognition},
pp. \bfpage{3156}--\blpage{3164}
(\byear{2015})
\end{bchapter}
\endbibitem

\bibitem[\protect\citeauthoryear{Wren et~al.}{1997}]{wren1997pfinder}
\begin{barticle}
\bauthor{\bsnm{Wren}, \binits{C.R.}},
\bauthor{\bsnm{Azarbayejani}, \binits{A.}},
\bauthor{\bsnm{Darrell}, \binits{T.}},
\bauthor{\bsnm{Pentland}, \binits{A.P.}}:
\batitle{Pfinder: Real-time tracking of the human body}.
\bjtitle{IEEE Transactions on pattern analysis and machine intelligence}
\bvolume{19}(\bissue{7}),
\bfpage{780}--\blpage{785}
(\byear{1997})
\end{barticle}
\endbibitem

\bibitem[\protect\citeauthoryear{Wang et~al.}{2022}]{wang2022open}
\begin{bchapter}
\bauthor{\bsnm{Wang}, \binits{W.}},
\bauthor{\bsnm{Feiszli}, \binits{M.}},
\bauthor{\bsnm{Wang}, \binits{H.}},
\bauthor{\bsnm{Malik}, \binits{J.}},
\bauthor{\bsnm{Tran}, \binits{D.}}:
\bctitle{Open-world instance segmentation: Exploiting pseudo ground truth from learned pairwise affinity}.
In: \bbtitle{Proceedings of the IEEE/CVF Conference on Computer Vision and Pattern Recognition},
pp. \bfpage{4422}--\blpage{4432}
(\byear{2022})
\end{bchapter}
\endbibitem

\bibitem[\protect\citeauthoryear{Wang et~al.}{2021}]{wang2021unidentified}
\begin{bchapter}
\bauthor{\bsnm{Wang}, \binits{W.}},
\bauthor{\bsnm{Feiszli}, \binits{M.}},
\bauthor{\bsnm{Wang}, \binits{H.}},
\bauthor{\bsnm{Tran}, \binits{D.}}:
\bctitle{Unidentified video objects: A benchmark for dense, open-world segmentation}.
In: \bbtitle{Proceedings of the IEEE/CVF International Conference on Computer Vision},
pp. \bfpage{10776}--\blpage{10785}
(\byear{2021})
\end{bchapter}
\endbibitem

\bibitem[\protect\citeauthoryear{Wu et~al.}{2022}]{wu2022unleashing}
\begin{botherref}
\oauthor{\bsnm{Wu}, \binits{J.}},
\oauthor{\bsnm{Li}, \binits{X.}},
\oauthor{\bsnm{Wei}, \binits{C.}},
\oauthor{\bsnm{Wang}, \binits{H.}},
\oauthor{\bsnm{Yuille}, \binits{A.}},
\oauthor{\bsnm{Zhou}, \binits{Y.}},
\oauthor{\bsnm{Xie}, \binits{C.}}:
Unleashing the power of visual prompting at the pixel level.
arXiv preprint arXiv:2212.10556
(2022)
\end{botherref}
\endbibitem

\bibitem[\protect\citeauthoryear{Wang et~al.}{2022}]{wang2022self}
\begin{bchapter}
\bauthor{\bsnm{Wang}, \binits{Y.}},
\bauthor{\bsnm{Shen}, \binits{X.}},
\bauthor{\bsnm{Hu}, \binits{S.X.}},
\bauthor{\bsnm{Yuan}, \binits{Y.}},
\bauthor{\bsnm{Crowley}, \binits{J.L.}},
\bauthor{\bsnm{Vaufreydaz}, \binits{D.}}:
\bctitle{Self-supervised transformers for unsupervised object discovery using normalized cut}.
In: \bbtitle{Proceedings of the IEEE/CVF Conference on Computer Vision and Pattern Recognition},
pp. \bfpage{14543}--\blpage{14553}
(\byear{2022})
\end{bchapter}
\endbibitem

\bibitem[\protect\citeauthoryear{Wei et~al.}{2022}]{wei2022chain}
\begin{barticle}
\bauthor{\bsnm{Wei}, \binits{J.}},
\bauthor{\bsnm{Wang}, \binits{X.}},
\bauthor{\bsnm{Schuurmans}, \binits{D.}},
\bauthor{\bsnm{Bosma}, \binits{M.}},
\bauthor{\bsnm{Xia}, \binits{F.}},
\bauthor{\bsnm{Chi}, \binits{E.}},
\bauthor{\bsnm{Le}, \binits{Q.V.}},
\bauthor{\bsnm{Zhou}, \binits{D.}}, \betal:
\batitle{Chain-of-thought prompting elicits reasoning in large language models}.
\bjtitle{Advances in neural information processing systems}
\bvolume{35},
\bfpage{24824}--\blpage{24837}
(\byear{2022})
\end{barticle}
\endbibitem

\bibitem[\protect\citeauthoryear{Wang et~al.}{2018}]{wang2018foreground}
\begin{barticle}
\bauthor{\bsnm{Wang}, \binits{Y.}},
\bauthor{\bsnm{Yu}, \binits{Z.}},
\bauthor{\bsnm{Zhu}, \binits{L.}}:
\batitle{Foreground detection with deeply learned multi-scale spatial-temporal features}.
\bjtitle{Sensors}
\bvolume{18}(\bissue{12}),
\bfpage{4269}
(\byear{2018})
\end{barticle}
\endbibitem

\bibitem[\protect\citeauthoryear{Wang et~al.}{2019}]{wang2019salient}
\begin{bchapter}
\bauthor{\bsnm{Wang}, \binits{W.}},
\bauthor{\bsnm{Zhao}, \binits{S.}},
\bauthor{\bsnm{Shen}, \binits{J.}},
\bauthor{\bsnm{Hoi}, \binits{S.C.}},
\bauthor{\bsnm{Borji}, \binits{A.}}:
\bctitle{Salient object detection with pyramid attention and salient edges}.
In: \bbtitle{Proceedings of the IEEE/CVF Conference on Computer Vision and Pattern Recognition},
pp. \bfpage{1448}--\blpage{1457}
(\byear{2019})
\end{bchapter}
\endbibitem

\bibitem[\protect\citeauthoryear{xAI}{2025}]{xai_2025}
\begin{botherref}
\oauthor{\bsnm{xAI}}:
RealworldQA
(2025).
\url{https://huggingface.co/datasets/xai-org/RealworldQA}
\end{botherref}
\endbibitem

\bibitem[\protect\citeauthoryear{Xiao et~al.}{2016}]{xiao2016sun}
\begin{barticle}
\bauthor{\bsnm{Xiao}, \binits{J.}},
\bauthor{\bsnm{Ehinger}, \binits{K.A.}},
\bauthor{\bsnm{Hays}, \binits{J.}},
\bauthor{\bsnm{Torralba}, \binits{A.}},
\bauthor{\bsnm{Oliva}, \binits{A.}}:
\batitle{Sun database: Exploring a large collection of scene categories}.
\bjtitle{International Journal of Computer Vision}
\bvolume{119},
\bfpage{3}--\blpage{22}
(\byear{2016})
\end{barticle}
\endbibitem

\bibitem[\protect\citeauthoryear{Xiao et~al.}{2010}]{xiao2010sun}
\begin{bchapter}
\bauthor{\bsnm{Xiao}, \binits{J.}},
\bauthor{\bsnm{Hays}, \binits{J.}},
\bauthor{\bsnm{Ehinger}, \binits{K.A.}},
\bauthor{\bsnm{Oliva}, \binits{A.}},
\bauthor{\bsnm{Torralba}, \binits{A.}}:
\bctitle{Sun database: Large-scale scene recognition from abbey to zoo}.
In: \bbtitle{2010 IEEE Computer Society Conference on Computer Vision and Pattern Recognition},
pp. \bfpage{3485}--\blpage{3492}
(\byear{2010}).
\bcomment{IEEE}
\end{bchapter}
\endbibitem

\bibitem[\protect\citeauthoryear{Xian et~al.}{2018}]{xian2018zero}
\begin{barticle}
\bauthor{\bsnm{Xian}, \binits{Y.}},
\bauthor{\bsnm{Lampert}, \binits{C.H.}},
\bauthor{\bsnm{Schiele}, \binits{B.}},
\bauthor{\bsnm{Akata}, \binits{Z.}}:
\batitle{Zero-shot learning—a comprehensive evaluation of the good, the bad and the ugly}.
\bjtitle{IEEE transactions on pattern analysis and machine intelligence}
\bvolume{41}(\bissue{9}),
\bfpage{2251}--\blpage{2265}
(\byear{2018})
\end{barticle}
\endbibitem

\bibitem[\protect\citeauthoryear{Yan et~al.}{2022}]{yan2022semantics}
\begin{barticle}
\bauthor{\bsnm{Yan}, \binits{C.}},
\bauthor{\bsnm{Chang}, \binits{X.}},
\bauthor{\bsnm{Luo}, \binits{M.}},
\bauthor{\bsnm{Liu}, \binits{H.}},
\bauthor{\bsnm{Zhang}, \binits{X.}},
\bauthor{\bsnm{Zheng}, \binits{Q.}}:
\batitle{Semantics-guided contrastive network for zero-shot object detection}.
\bjtitle{IEEE transactions on pattern analysis and machine intelligence}
\bvolume{46}(\bissue{3}),
\bfpage{1530}--\blpage{1544}
(\byear{2022})
\end{barticle}
\endbibitem

\bibitem[\protect\citeauthoryear{Yao et~al.}{2021}]{yao2021filip}
\begin{botherref}
\oauthor{\bsnm{Yao}, \binits{L.}},
\oauthor{\bsnm{Huang}, \binits{R.}},
\oauthor{\bsnm{Hou}, \binits{L.}},
\oauthor{\bsnm{Lu}, \binits{G.}},
\oauthor{\bsnm{Niu}, \binits{M.}},
\oauthor{\bsnm{Xu}, \binits{H.}},
\oauthor{\bsnm{Liang}, \binits{X.}},
\oauthor{\bsnm{Li}, \binits{Z.}},
\oauthor{\bsnm{Jiang}, \binits{X.}},
\oauthor{\bsnm{Xu}, \binits{C.}}:
Filip: Fine-grained interactive language-image pre-training.
arXiv preprint arXiv:2111.07783
(2021)
\end{botherref}
\endbibitem

\bibitem[\protect\citeauthoryear{Young et~al.}{2014}]{young2014image}
\begin{barticle}
\bauthor{\bsnm{Young}, \binits{P.}},
\bauthor{\bsnm{Lai}, \binits{A.}},
\bauthor{\bsnm{Hodosh}, \binits{M.}},
\bauthor{\bsnm{Hockenmaier}, \binits{J.}}:
\batitle{From image descriptions to visual denotations: New similarity metrics for semantic inference over event descriptions}.
\bjtitle{Transactions of the Association for Computational Linguistics}
\bvolume{2},
\bfpage{67}--\blpage{78}
(\byear{2014})
\end{barticle}
\endbibitem

\bibitem[\protect\citeauthoryear{Yue et~al.}{2024}]{yue2023mmmu}
\begin{bchapter}
\bauthor{\bsnm{Yue}, \binits{X.}},
\bauthor{\bsnm{Ni}, \binits{Y.}},
\bauthor{\bsnm{Zhang}, \binits{K.}},
\bauthor{\bsnm{Zheng}, \binits{T.}},
\bauthor{\bsnm{Liu}, \binits{R.}},
\bauthor{\bsnm{Zhang}, \binits{G.}},
\bauthor{\bsnm{Stevens}, \binits{S.}},
\bauthor{\bsnm{Jiang}, \binits{D.}},
\bauthor{\bsnm{Ren}, \binits{W.}},
\bauthor{\bsnm{Sun}, \binits{Y.}},
\bauthor{\bsnm{Wei}, \binits{C.}},
\bauthor{\bsnm{Yu}, \binits{B.}},
\bauthor{\bsnm{Yuan}, \binits{R.}},
\bauthor{\bsnm{Sun}, \binits{R.}},
\bauthor{\bsnm{Yin}, \binits{M.}},
\bauthor{\bsnm{Zheng}, \binits{B.}},
\bauthor{\bsnm{Yang}, \binits{Z.}},
\bauthor{\bsnm{Liu}, \binits{Y.}},
\bauthor{\bsnm{Huang}, \binits{W.}},
\bauthor{\bsnm{Sun}, \binits{H.}},
\bauthor{\bsnm{Su}, \binits{Y.}},
\bauthor{\bsnm{Chen}, \binits{W.}}:
\bctitle{Mmmu: A massive multi-discipline multimodal understanding and reasoning benchmark for expert agi}.
In: \bbtitle{Proceedings of CVPR}
(\byear{2024})
\end{bchapter}
\endbibitem

\bibitem[\protect\citeauthoryear{Yang et~al.}{2022}]{yang2022openood}
\begin{barticle}
\bauthor{\bsnm{Yang}, \binits{J.}},
\bauthor{\bsnm{Wang}, \binits{P.}},
\bauthor{\bsnm{Zou}, \binits{D.}},
\bauthor{\bsnm{Zhou}, \binits{Z.}},
\bauthor{\bsnm{Ding}, \binits{K.}},
\bauthor{\bsnm{Peng}, \binits{W.}},
\bauthor{\bsnm{Wang}, \binits{H.}},
\bauthor{\bsnm{Chen}, \binits{G.}},
\bauthor{\bsnm{Li}, \binits{B.}},
\bauthor{\bsnm{Sun}, \binits{Y.}}, \betal:
\batitle{Openood: Benchmarking generalized out-of-distribution detection}.
\bjtitle{Advances in Neural Information Processing Systems}
\bvolume{35},
\bfpage{32598}--\blpage{32611}
(\byear{2022})
\end{barticle}
\endbibitem

\bibitem[\protect\citeauthoryear{Yan et~al.}{2024}]{yan2024list}
\begin{botherref}
\oauthor{\bsnm{Yan}, \binits{A.}},
\oauthor{\bsnm{Yang}, \binits{Z.}},
\oauthor{\bsnm{Wu}, \binits{J.}},
\oauthor{\bsnm{Zhu}, \binits{W.}},
\oauthor{\bsnm{Yang}, \binits{J.}},
\oauthor{\bsnm{Li}, \binits{L.}},
\oauthor{\bsnm{Lin}, \binits{K.}},
\oauthor{\bsnm{Wang}, \binits{J.}},
\oauthor{\bsnm{McAuley}, \binits{J.}},
\oauthor{\bsnm{Gao}, \binits{J.}}, et al.:
List items one by one: A new data source and learning paradigm for multimodal llms.
arXiv preprint arXiv:2404.16375
(2024)
\end{botherref}
\endbibitem

\bibitem[\protect\citeauthoryear{Yu et~al.}{2024}]{yu2024attention}
\begin{bchapter}
\bauthor{\bsnm{Yu}, \binits{R.}},
\bauthor{\bsnm{Yu}, \binits{W.}},
\bauthor{\bsnm{Wang}, \binits{X.}}:
\bctitle{Attention prompting on image for large vision-language models}.
In: \bbtitle{European Conference on Computer Vision},
pp. \bfpage{251}--\blpage{268}
(\byear{2024}).
\bcomment{Springer}
\end{bchapter}
\endbibitem

\bibitem[\protect\citeauthoryear{You et~al.}{2023}]{you2023ferret}
\begin{botherref}
\oauthor{\bsnm{You}, \binits{H.}},
\oauthor{\bsnm{Zhang}, \binits{H.}},
\oauthor{\bsnm{Gan}, \binits{Z.}},
\oauthor{\bsnm{Du}, \binits{X.}},
\oauthor{\bsnm{Zhang}, \binits{B.}},
\oauthor{\bsnm{Wang}, \binits{Z.}},
\oauthor{\bsnm{Cao}, \binits{L.}},
\oauthor{\bsnm{Chang}, \binits{S.-F.}},
\oauthor{\bsnm{Yang}, \binits{Y.}}:
Ferret: Refer and ground anything anywhere at any granularity.
arXiv preprint arXiv:2310.07704
(2023)
\end{botherref}
\endbibitem

\bibitem[\protect\citeauthoryear{Yang et~al.}{2023}]{yang2023set}
\begin{botherref}
\oauthor{\bsnm{Yang}, \binits{J.}},
\oauthor{\bsnm{Zhang}, \binits{H.}},
\oauthor{\bsnm{Li}, \binits{F.}},
\oauthor{\bsnm{Zou}, \binits{X.}},
\oauthor{\bsnm{Li}, \binits{C.}},
\oauthor{\bsnm{Gao}, \binits{J.}}:
Set-of-mark prompting unleashes extraordinary visual grounding in gpt-4v.
arXiv preprint arXiv:2310.11441
(2023)
\end{botherref}
\endbibitem

\bibitem[\protect\citeauthoryear{Yang et~al.}{2024}]{yang2024generalized}
\begin{barticle}
\bauthor{\bsnm{Yang}, \binits{J.}},
\bauthor{\bsnm{Zhou}, \binits{K.}},
\bauthor{\bsnm{Li}, \binits{Y.}},
\bauthor{\bsnm{Liu}, \binits{Z.}}:
\batitle{Generalized out-of-distribution detection: A survey}.
\bjtitle{International Journal of Computer Vision}
\bvolume{132}(\bissue{12}),
\bfpage{5635}--\blpage{5662}
(\byear{2024})
\end{barticle}
\endbibitem

\bibitem[\protect\citeauthoryear{Zhu and Chen}{2024}]{zhu2024survey}
\begin{botherref}
\oauthor{\bsnm{Zhu}, \binits{C.}},
\oauthor{\bsnm{Chen}, \binits{L.}}:
A survey on open-vocabulary detection and segmentation: Past, present, and future.
IEEE Transactions on Pattern Analysis and Machine Intelligence
(2024)
\end{botherref}
\endbibitem

\bibitem[\protect\citeauthoryear{Zou et~al.}{2023}]{zou2023object}
\begin{barticle}
\bauthor{\bsnm{Zou}, \binits{Z.}},
\bauthor{\bsnm{Chen}, \binits{K.}},
\bauthor{\bsnm{Shi}, \binits{Z.}},
\bauthor{\bsnm{Guo}, \binits{Y.}},
\bauthor{\bsnm{Ye}, \binits{J.}}:
\batitle{Object detection in 20 years: A survey}.
\bjtitle{Proceedings of the IEEE}
\bvolume{111}(\bissue{3}),
\bfpage{257}--\blpage{276}
(\byear{2023})
\end{barticle}
\endbibitem

\bibitem[\protect\citeauthoryear{Zou et~al.}{2023}]{zou2023generalized}
\begin{bchapter}
\bauthor{\bsnm{Zou}, \binits{X.}},
\bauthor{\bsnm{Dou}, \binits{Z.-Y.}},
\bauthor{\bsnm{Yang}, \binits{J.}},
\bauthor{\bsnm{Gan}, \binits{Z.}},
\bauthor{\bsnm{Li}, \binits{L.}},
\bauthor{\bsnm{Li}, \binits{C.}},
\bauthor{\bsnm{Dai}, \binits{X.}},
\bauthor{\bsnm{Behl}, \binits{H.}},
\bauthor{\bsnm{Wang}, \binits{J.}},
\bauthor{\bsnm{Yuan}, \binits{L.}}, \betal:
\bctitle{Generalized decoding for pixel, image, and language}.
In: \bbtitle{Proceedings of the IEEE/CVF Conference on Computer Vision and Pattern Recognition},
pp. \bfpage{15116}--\blpage{15127}
(\byear{2023})
\end{bchapter}
\endbibitem

\bibitem[\protect\citeauthoryear{Zhou et~al.}{2021a}]{zhou2021rgb}
\begin{barticle}
\bauthor{\bsnm{Zhou}, \binits{T.}},
\bauthor{\bsnm{Fan}, \binits{D.-P.}},
\bauthor{\bsnm{Cheng}, \binits{M.-M.}},
\bauthor{\bsnm{Shen}, \binits{J.}},
\bauthor{\bsnm{Shao}, \binits{L.}}:
\batitle{Rgb-d salient object detection: A survey}.
\bjtitle{Computational Visual Media}
\bvolume{7},
\bfpage{37}--\blpage{69}
(\byear{2021})
\end{barticle}
\endbibitem

\bibitem[\protect\citeauthoryear{Zhou et~al.}{2021b}]{zhou2021specificity}
\begin{bchapter}
\bauthor{\bsnm{Zhou}, \binits{T.}},
\bauthor{\bsnm{Fu}, \binits{H.}},
\bauthor{\bsnm{Chen}, \binits{G.}},
\bauthor{\bsnm{Zhou}, \binits{Y.}},
\bauthor{\bsnm{Fan}, \binits{D.-P.}},
\bauthor{\bsnm{Shao}, \binits{L.}}:
\bctitle{Specificity-preserving rgb-d saliency detection}.
In: \bbtitle{Proceedings of the IEEE/CVF International Conference on Computer Vision},
pp. \bfpage{4681}--\blpage{4691}
(\byear{2021})
\end{bchapter}
\endbibitem

\bibitem[\protect\citeauthoryear{Zhou et~al.}{2017}]{zhou2017places}
\begin{barticle}
\bauthor{\bsnm{Zhou}, \binits{B.}},
\bauthor{\bsnm{Lapedriza}, \binits{A.}},
\bauthor{\bsnm{Khosla}, \binits{A.}},
\bauthor{\bsnm{Oliva}, \binits{A.}},
\bauthor{\bsnm{Torralba}, \binits{A.}}:
\batitle{Places: A 10 million image database for scene recognition}.
\bjtitle{IEEE transactions on pattern analysis and machine intelligence}
\bvolume{40}(\bissue{6}),
\bfpage{1452}--\blpage{1464}
(\byear{2017})
\end{barticle}
\endbibitem

\bibitem[\protect\citeauthoryear{Zou et~al.}{2023}]{zou2023interfacing}
\begin{botherref}
\oauthor{\bsnm{Zou}, \binits{X.}},
\oauthor{\bsnm{Li}, \binits{L.}},
\oauthor{\bsnm{Wang}, \binits{J.}},
\oauthor{\bsnm{Yang}, \binits{J.}},
\oauthor{\bsnm{Ding}, \binits{M.}},
\oauthor{\bsnm{Yang}, \binits{Z.}},
\oauthor{\bsnm{Li}, \binits{F.}},
\oauthor{\bsnm{Zhang}, \binits{H.}},
\oauthor{\bsnm{Liu}, \binits{S.}},
\oauthor{\bsnm{Aravinthan}, \binits{A.}}, et al.:
Interfacing foundation models' embeddings.
arXiv preprint arXiv:2312.07532
(2023)
\end{botherref}
\endbibitem

\bibitem[\protect\citeauthoryear{Zhu et~al.}{2014}]{zhu2014saliency}
\begin{bchapter}
\bauthor{\bsnm{Zhu}, \binits{W.}},
\bauthor{\bsnm{Liang}, \binits{S.}},
\bauthor{\bsnm{Wei}, \binits{Y.}},
\bauthor{\bsnm{Sun}, \binits{J.}}:
\bctitle{Saliency optimization from robust background detection}.
In: \bbtitle{Proceedings of the IEEE Conference on Computer Vision and Pattern Recognition},
pp. \bfpage{2814}--\blpage{2821}
(\byear{2014})
\end{bchapter}
\endbibitem

\bibitem[\protect\citeauthoryear{Zhao et~al.}{2023}]{zhao2023revisiting}
\begin{botherref}
\oauthor{\bsnm{Zhao}, \binits{X.}},
\oauthor{\bsnm{Ma}, \binits{Y.}},
\oauthor{\bsnm{Wang}, \binits{D.}},
\oauthor{\bsnm{Shen}, \binits{Y.}},
\oauthor{\bsnm{Qiao}, \binits{Y.}},
\oauthor{\bsnm{Liu}, \binits{X.}}:
Revisiting open world object detection.
IEEE Transactions on Circuits and Systems for Video Technology
(2023)
\end{botherref}
\endbibitem

\bibitem[\protect\citeauthoryear{Zhang et~al.}{2023}]{zhang2023gpt4roi}
\begin{botherref}
\oauthor{\bsnm{Zhang}, \binits{S.}},
\oauthor{\bsnm{Sun}, \binits{P.}},
\oauthor{\bsnm{Chen}, \binits{S.}},
\oauthor{\bsnm{Xiao}, \binits{M.}},
\oauthor{\bsnm{Shao}, \binits{W.}},
\oauthor{\bsnm{Zhang}, \binits{W.}},
\oauthor{\bsnm{Liu}, \binits{Y.}},
\oauthor{\bsnm{Chen}, \binits{K.}},
\oauthor{\bsnm{Luo}, \binits{P.}}:
Gpt4roi: Instruction tuning large language model on region-of-interest.
arXiv preprint arXiv:2307.03601
(2023)
\end{botherref}
\endbibitem

\bibitem[\protect\citeauthoryear{Zhu et~al.}{2020}]{zhu2020don}
\begin{bchapter}
\bauthor{\bsnm{Zhu}, \binits{P.}},
\bauthor{\bsnm{Wang}, \binits{H.}},
\bauthor{\bsnm{Saligrama}, \binits{V.}}:
\bctitle{Don't even look once: Synthesizing features for zero-shot detection}.
In: \bbtitle{Proceedings of the IEEE/CVF Conference on Computer Vision and Pattern Recognition},
pp. \bfpage{11693}--\blpage{11702}
(\byear{2020})
\end{bchapter}
\endbibitem

\bibitem[\protect\citeauthoryear{Zhang et~al.}{2024}]{zhang2024mm}
\begin{botherref}
\oauthor{\bsnm{Zhang}, \binits{D.}},
\oauthor{\bsnm{Yu}, \binits{Y.}},
\oauthor{\bsnm{Dong}, \binits{J.}},
\oauthor{\bsnm{Li}, \binits{C.}},
\oauthor{\bsnm{Su}, \binits{D.}},
\oauthor{\bsnm{Chu}, \binits{C.}},
\oauthor{\bsnm{Yu}, \binits{D.}}:
Mm-llms: Recent advances in multimodal large language models.
arXiv preprint arXiv:2401.13601
(2024)
\end{botherref}
\endbibitem

\bibitem[\protect\citeauthoryear{Zou et~al.}{2024}]{zou2024segment}
\begin{botherref}
\oauthor{\bsnm{Zou}, \binits{X.}},
\oauthor{\bsnm{Yang}, \binits{J.}},
\oauthor{\bsnm{Zhang}, \binits{H.}},
\oauthor{\bsnm{Li}, \binits{F.}},
\oauthor{\bsnm{Li}, \binits{L.}},
\oauthor{\bsnm{Wang}, \binits{J.}},
\oauthor{\bsnm{Wang}, \binits{L.}},
\oauthor{\bsnm{Gao}, \binits{J.}},
\oauthor{\bsnm{Lee}, \binits{Y.J.}}:
Segment everything everywhere all at once.
Advances in Neural Information Processing Systems
\textbf{36}
(2024)
\end{botherref}
\endbibitem

\bibitem[\protect\citeauthoryear{Zhou et~al.}{2017}]{zhou2017scene}
\begin{bchapter}
\bauthor{\bsnm{Zhou}, \binits{B.}},
\bauthor{\bsnm{Zhao}, \binits{H.}},
\bauthor{\bsnm{Puig}, \binits{X.}},
\bauthor{\bsnm{Fidler}, \binits{S.}},
\bauthor{\bsnm{Barriuso}, \binits{A.}},
\bauthor{\bsnm{Torralba}, \binits{A.}}:
\bctitle{Scene parsing through ade20k dataset}.
In: \bbtitle{Proceedings of the IEEE Conference on Computer Vision and Pattern Recognition},
pp. \bfpage{633}--\blpage{641}
(\byear{2017})
\end{bchapter}
\endbibitem

\bibitem[\protect\citeauthoryear{Zhou et~al.}{2019}]{zhou2019semantic}
\begin{barticle}
\bauthor{\bsnm{Zhou}, \binits{B.}},
\bauthor{\bsnm{Zhao}, \binits{H.}},
\bauthor{\bsnm{Puig}, \binits{X.}},
\bauthor{\bsnm{Xiao}, \binits{T.}},
\bauthor{\bsnm{Fidler}, \binits{S.}},
\bauthor{\bsnm{Barriuso}, \binits{A.}},
\bauthor{\bsnm{Torralba}, \binits{A.}}:
\batitle{Semantic understanding of scenes through the ade20k dataset}.
\bjtitle{International Journal of Computer Vision}
\bvolume{127},
\bfpage{302}--\blpage{321}
(\byear{2019})
\end{barticle}
\endbibitem

\bibitem[\protect\citeauthoryear{Zhang et~al.}{2024}]{zhang2024vision}
\begin{botherref}
\oauthor{\bsnm{Zhang}, \binits{Q.}},
\oauthor{\bsnm{Zhang}, \binits{J.}},
\oauthor{\bsnm{Xu}, \binits{Y.}},
\oauthor{\bsnm{Tao}, \binits{D.}}:
Vision transformer with quadrangle attention.
IEEE Transactions on Pattern Analysis and Machine Intelligence
(2024)
\end{botherref}
\endbibitem

\bibitem[\protect\citeauthoryear{Zhao et~al.}{2019}]{zhao2019object}
\begin{barticle}
\bauthor{\bsnm{Zhao}, \binits{Z.-Q.}},
\bauthor{\bsnm{Zheng}, \binits{P.}},
\bauthor{\bsnm{Xu}, \binits{S.-t.}},
\bauthor{\bsnm{Wu}, \binits{X.}}:
\batitle{Object detection with deep learning: A review}.
\bjtitle{IEEE transactions on neural networks and learning systems}
\bvolume{30}(\bissue{11}),
\bfpage{3212}--\blpage{3232}
(\byear{2019})
\end{barticle}
\endbibitem

\end{thebibliography}

\end{document}